%% file: llmphy_arxiv_v2.tex
\documentclass[twoside]{article}

\usepackage[preprint]{aistats2026}

\usepackage[utf8]{inputenc} % allow utf-8 input
\usepackage[T1]{fontenc}    % use 8-bit T1 fonts
\usepackage{hyperref}       % hyperlinks
\usepackage{url}            % simple URL typesetting
\usepackage{booktabs}       % professional-quality tables
\usepackage{amsfonts}       % blackboard math symbols
\usepackage{nicefrac}       % compact symbols for 1/2, etc.
\usepackage{microtype}      % microtypography
\usepackage{xcolor}         % colors
\usepackage{subfigure}
\usepackage{graphicx}
\usepackage{amsmath,amssymb,amsthm}
\usepackage{mathtools}
\usepackage{caption}

\usepackage{algorithm}
\usepackage{algorithmic}
\usepackage{overpic}
\usepackage{array}
\usepackage{xcolor}
\usepackage[most]{tcolorbox}
\usepackage{cuted}
\tcbuselibrary{breakable}
\usepackage{listings}
\lstdefinestyle{mypython}{
    language=Python,
    basicstyle=\ttfamily\small,
    keywordstyle=\color{blue!70!black}\bfseries,
    commentstyle=\color{green!40!black}\itshape,
    stringstyle=\color{orange!60!black},
    numberstyle=\tiny\color{gray},
    numbers=left,
    stepnumber=1,
    numbersep=8pt,
    showstringspaces=false,
    breaklines=true,
    breakatwhitespace=true,
    frame=single,
    rulecolor=\color{black!20},
    backgroundcolor=\color{gray!3},
    tabsize=4,
    columns=fullflexible
}

\include{definitions}

\begin{document}

\runningtitle{LLMPhy: Parameter-Identifiable Physical Reasoning}

\twocolumn[

\aistatstitle{LLMPhy: Parameter-Identifiable Physical Reasoning \\Combining Large Language Models and Physics Engines}

\aistatsauthor{ Anoop Cherian \And Radu Corcodel \And  Siddarth Jain \And Diego Romeres }

\aistatsaddress{ Mitsubishi Electric Research Labs (MERL), Cambridge, MA \\
\textcolor{blue}{\href{https://www.merl.com/research/highlights/LLMPhy}{https://merl.com/research/highlights/LLMPhy}}}]

\begin{abstract}
Most learning-based approaches to complex physical reasoning sidestep the crucial problem of parameter identification (e.g., mass, friction) that governs scene dynamics—despite its importance in real-world applications such as collision avoidance and robotic manipulation. In this paper, we present $\llmphy$, a black-box optimization framework that integrates large language models (LLMs) with physics simulators for physical reasoning. The core insight of $\llmphy$ is to bridge the textbook physical knowledge embedded in LLMs with the world models implemented in modern physics engines, enabling the construction of digital twins of input scenes via latent parameter estimation. Specifically, $\llmphy$ decomposes digital twin construction into two subproblems: (i) a continuous problem of estimating physical parameters and (ii) a discrete problem of estimating scene layout. For each subproblem, $\llmphy$ iteratively prompts the LLM to generate computer programs encoding parameter estimates, executes them in the physics engine to reconstruct the scene, and uses the resulting reconstruction error as feedback to refine the LLM’s predictions. As existing physical reasoning benchmarks rarely account for parameter identifiability, we introduce three new datasets designed to evaluate physical reasoning in zero-shot settings. Our results show that $\llmphy$ achieves state-of-the-art performance on our tasks, recovers physical parameters more accurately, and converges more reliably than prior black-box methods.
\end{abstract}

\section{INTRODUCTION}
The emergence of large language models (LLMs) has catalyzed the design of novel approaches that leverage their vast inherent knowledge for solving challenging real-world problems. A few notable such efforts include robotic planning~\cite{huang2022language,kim2024openvla,song2023llm}, complex code generation~\cite{chen2021evaluating,jin2023inferfix,tang2024worldcoder}, solving optimization objectives~\cite{hao2024large,yang2024largelanguagemodelsoptimizers}, conducting mathematical reasoning~\cite{lewkowycz2022solving,polu2022formal,trinh2024solving}, or even making scientific discoveries~\cite{romera2024mathematical}. As we start to incorporate LLMs into  embodied systems that operate in the physical world, it is crucial that they are equipped with precise world models and are adept at complex physical reasoning. There have been attempts at developing LLMs towards this end~\cite{aroca2021prost,bisk2020piqa,tian2023macgyver}, however, it is often seen that many of the leading models struggle with reasoning about physical interactions~\cite{wang2023newton}.
%zellers2018swag

\begin{figure*}[t]
    \centering
    % width=8.5cm,trim={1.5cm 8.5cm 12cm 5cm},clip
    \includegraphics[width=13cm,trim={0.75cm 12.2cm 1.1cm 4.2cm},clip]{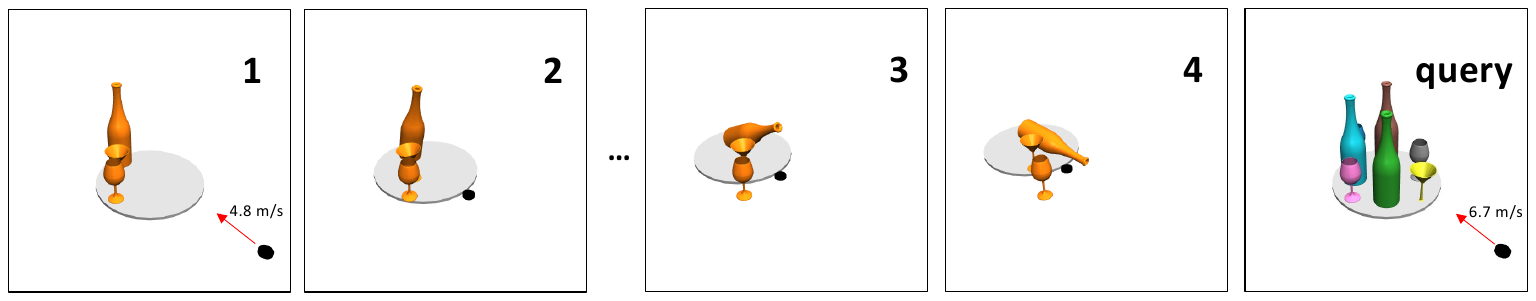}
    \caption{An illustration of our TraySim task. Given a video sequence with a set of objects each having distinct physical parameters (e.g., friction, damping, restitution), and a query image, our task is to predict the objects in the query image which will remain upright after the pusher impact.
    }
    \label{fig:example_scene}
    \vspace*{-0.25cm}
\end{figure*}

While state-of-the-art large language models (LLMs) demonstrate an impressive understanding of the governing physical rules of the world, their inability to dynamically engage with the world is perhaps limiting their abilities in solving complex physical reasoning problems~\cite{harter2020solving, riochet2021intphys, wang2023newton, xue2021phy}. For instance, consider the setup in Figure~\ref{fig:example_scene}, where given a video (with frames marked 1-4), the goal is to predict the stable/upright objects in the query image after the collision. While a sophisticated model might make an educated guess based on ``intuitive physics'', however obtaining the correct solution requires deep reasoning about the underlying physical properties and the non-linear dynamics of the objects involved. Humans, for example, would approach this by constructing a mental model of the scene, reasoning through the underlying physics (e.g., observing that the wine glass slides while the bottle falls, which implies the wine glass likely has less friction), and then using that model to predict the outcome. Can we enable machine learning models to construct such mental models or physical digital twins for accurate physical reasoning? Learning to model these complex dynamics might demand huge datasets and training large models. However, advancements in computer graphics have already led to the development of physics engines that are capable of simulating realistic world models. Thus, can we combine a physics simulator in tandem with the knowledge embedded in pretrained LLMs to build a digital twin of the scene for physical reasoning? 

Guided by this insight, we present $\llmphy$, a black-box optimization framework that leverages the program synthesis capabilities of state-of-the-art LLMs to communicate with a physics simulator for iteratively constructing digital twins and solving physical reasoning tasks. $\llmphy$ operates in two phases: (i) parameter estimation, where the LLM–simulator combination functions as a continuous black-box optimizer to infer the unobservable physical parameters of the scene (e.g., friction, restitution), and (ii) scene understanding, where it acts as a discrete optimizer to reconstruct the observable layout. A key innovation in $\llmphy$ is its implicit analysis-by-synthesis loop: actual physical parameters are recovered indirectly by minimizing a surrogate reconstruction loss through alternating interactions between the LLM and the physics engine. Concretely, the LLM proposes programs encoding updated parameter estimates, the simulator executes them to recreate the scene, and the resulting reconstruction error is fed back to the LLM as prompts for refinement until a convergence criterion on the surrogate loss is met. Since no assumptions are made about the analytic form of the optimization objective, $\llmphy$ operates fully in a black-box setting. 

Numerous datasets have been proposed to study the physical reasoning abilities of machine learning models~\cite{chen2022comphy,tung2023physion++,bakhtin2019phyre,liu2025generative,li2019learning,sanchez2020learning}. While many provide complex realistic simulations, they are not designed for estimating latent physical parameters or to interact with LLMs, making them inadequate for our setup. To address this gap, we introduce three new zero-shot tasks and datasets—TraySim, CLEVRER-LLMPhy, and Real-TraySim—together with their simulator APIs. The first two are synthetic datasets following the template of constructing digital twins by estimating latent physics parameters and then using the twins to solve more complex reasoning tasks (Figure~\ref{fig:example_scene}). The third, Real-TraySim, extends this to a real tabletop robotic manipulation setting that requires explicit digital twin construction, integrating computer vision and physics simulation to predict object trajectories under collision.

Experiments on these datasets show that $\llmphy$ outperforms popular black-box baselines such as Bayesian reasoning~\cite{mockus1974bayesian} and CMA-ES~\cite{hansen2016cma}, achieving 6–12\% higher accuracy on our reasoning tasks. We further observe faster and more stable convergence, suggesting that $\llmphy$ exploits the embedded physics knowledge in LLMs to infer latent parameters more efficiently. Notably, stronger reasoning models (e.g., OpenAI o1) converge more effectively than weaker ones (e.g., GPT-4o, Gemma3-12B). Finally, real-world experiments confirm the effectiveness of $\llmphy$, demonstrating superior generalization and consistent performance gains, highlighting its broad applicability to physical reasoning tasks.
 
We summarize below our main contributions:
\begin{itemize}
    \item \textbf{Task:} We study the novel problem of reasoning over latent physical parameters to predict outcomes in complex, contact-rich scenes.  
    \item \textbf{Method:} We propose $\llmphy$, a zero-shot framework that iteratively estimates physical parameters via LLM-based program synthesis and execution in a physics engine.  
    \item \textbf{Datasets and Experiments:} We introduce three new tasks—TraySim, CLEVRER-LLMPhy, and Real-TraySim. Our experiments on these tasks show that $\llmphy$ achieves state-of-the-art accuracy, underscoring its effectiveness for physical reasoning.  
\end{itemize}

\section{RELATED WORKS}
\noindent\textbf{Scene De-Rendering.} Our work is inspired by the early efforts on physical scene understanding using analysis-by-synthesis with simulation. For example, Galileo~\cite{wu2015galileo} integrated a 3D physics engine into a generative model, using simulations (with an object-based physics representation) to infer latent properties like mass, friction, and shape from videos. The neural scene de-rendering framework~\cite{wu2017neuralA} treats a renderer as the decoder of an autoencoder, forcing the latent code to be physically interpretable.
In~\cite{wu2017learningB}, a visual de-animation framework is proposed that parses video frames to infer physical parameters and then simulating forward, essentially “de-animating” observed motions to explain them. These approaches pioneered the use of simulators-in-the-loop to ensure the inferred scene properties produce realistic outcomes, thereby establishing a foundation to use  inverse graphics/physics for physical reasoning. In contrast, we propose a closed-loop analysis-by-synthesis optimization that leverages LLM-physics engine synergy to configure a world model building on the embedded general knowledge of LLMs for physical reasoning.

\noindent\textbf{LLMs with Physics Engines.} There have been approaches that use LLMs as high-level reasoners and program synthesizers for physical reasoning. For example in Mind’s Eye~\cite{liu2022mind}, an LLM asks a physics engine to run experiments and uses the outcome to answer a question, however does not use the LLM-simulator loop. In SimLM~\cite{memery2023simlm}, an LLM-simulator combination is presented for a simple projectile motion estimation, but assumes access to in-context examples from previous successful runs for LLM guidance. Eureka~\cite{ma2023eureka} proposes an LLM-based evolutionary program synthesis approach, but for designing rewards in reinforcement learning.

Prior works have also used differentiable physics engines for parameter estimation tasks, e.g.,~\cite{kandukuri2020learning}, GradSim~\cite{murthy2020gradsim}. These methods implement physics equations directly, treat physical parameters as learnable variables, and optimize them using gradient backpropagation. A soft differentiable renderer is used to propagate gradients from rendered images/videos back to physical parameters. In contrast, $\llmphy$ is entirely gradient-free. The LLM and the physics simulator are treated as black boxes with no gradient flow across them. The simulator can be any off-the-shelf general purpose physics engine (e.g., MuJoCo, PyBullet), and need not be differentiable. This fundamental difference means the techniques and assumptions in GradSim-style methods are not applicable in our setting.

\noindent\textbf{LLMs as Optimizers:} LLMs have recently been explored for optimization. For example, \cite{yang2024largelanguagemodelsoptimizers} shows that prompting alone can yield good solutions on small-scale problems. The success of any optimizer depends on generating effective descent directions: when gradients are available, efficient methods exist~\cite{sun2019survey}; without them, black-box methods such as Bayesian optimization~\cite{mockus1974bayesian}, Multi-Objective BO~\cite{konakovic2020diversity}, and CMA-ES~\cite{hansen2001completely,terayama2021black} are typically used. Only a few studies have examined LLMs in this setting. \cite{guo2023towards} shows that LLMs can iteratively propose candidate solutions, influenced by their pretrained knowledge, while \cite{nie2024importance} highlights the role of feedback quality in navigating complex objectives. In contrast, our work couples an LLM with a physics engine to enable physical reasoning. Ma et al.~\cite{pmlr-v235-ma24m} propose a bi-level framework in which an LLM generates symbolic expressions of physical systems, while a differentiable simulator instantiates these expressions, estimates parameters via backpropagation, and evaluates their fit to data. In contrast, $\llmphy$ assumes an implicit world model and reasons over physical attributes using a decoupled simulator without requiring explicit symbolic forms or differentiability, enabling broader applicability and complementing the prior approach.

\noindent\textbf{Tasks, Datasets, and Benchmarks.} A closely related work is Physion~\cite{bear2021physion}, which evaluates outcome prediction from passive videos with human comparisons. PHYRE~\cite{bakhtin2019phyre} studies counterfactual planning in 2D puzzles, while CLEVRER~\cite{yi2019clevrer} emphasizes causal and event reasoning from video. However, neither Physion nor PHYRE supervises or evaluates recovery of latent physical parameters (e.g., mass, friction, inertia). ComPhy~\cite{chen2022comphy} introduces hidden properties but addresses them in a supervised video-QA setting rather than through active parameter identification. Learned dynamics approaches such as GNS~\cite{sanchez2020learning} and DPI-Net~\cite{li2019learning} train forward models on large rollout corpora but do not support zero-/few-shot parameter recovery or intervention design. In contrast, our work evaluates latent-parameter–aware reasoning under LLM-guided active interventions, with validation provided via a simulator-agnostic API. Finally, while Physion++~\cite{tung2023physion++} considers latent-property evaluation from video, our pipeline explicitly designs interventions to resolve identifiability in a black-box setting without task-specific supervision.

\section{PROPOSED METHOD}
\begin{figure*}[h!]
    \centering
    \includegraphics[width=14cm,trim={0.3cm 5.5cm 0cm 4.5cm},clip]{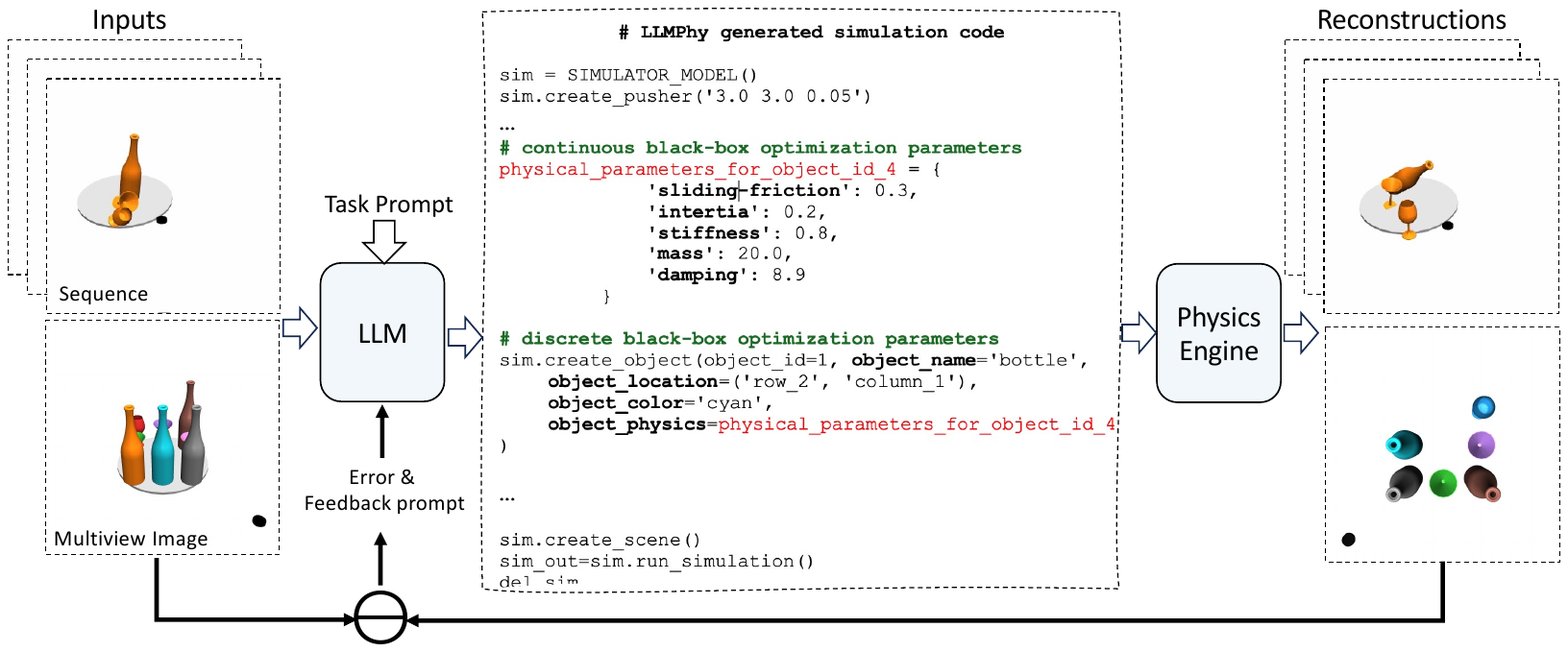}
    \caption{$\llmphy$ architecture. The LLM is prompted with multi-view images and object motion sequences to synthesize Python code characterizing the physics and object layout. The code is executed in the simulator producing scene reconstructions, which are matched to the inputs producing error. In the next iteration, the LLM is prompted to improvise its estimations to reduce the error.}
    \label{fig:arch}
\end{figure*}

In this section, we present $\llmphy$ for physical reasoning through digital twin construction. To illustrate our novel setup, we use the generalized task template from TraySim (Figure~\ref{fig:example_scene}), which helps motivate the modules of our architecture (Figure~\ref{fig:arch}). 

\subsection{Problem Setup and Formulation}\label{sec:ProblemSetup}
As illustrated in Figure~\ref{fig:example_scene}, our goal is to predict the outcomes of highly dynamic physical reasoning setups by constructing digital twins that capture both observable entities and latent physical parameters. Once inferred, these parameters can be reused for reasoning about more complex scenes. This requires addressing two sub-tasks: (i) estimating unobservable physical parameters that govern object dynamics (e.g., mass, friction, restitution), and (ii) estimating observable scene properties (e.g., object placements, appearances, poses). To study these facets, we introduce TraySim, a synthetic environment with complex multi-body dynamics and contact-rich interactions, where ground-truth parameters enable detailed evaluation of parameter estimation.

Each TraySim task instance consists of a multi-view video and a corresponding set of initial images, called \emph{query images} henceforth: the video is used to estimate unobservable parameters, while the query images from each view (captured before any motion) define the task (see Figure~\ref{fig:example_scene}). Multi-view inputs mitigate occlusions and appearance ambiguities, allowing models to focus on physical reasoning. Formally, let $\videov=\tuple{\fr_1, \fr_2, \cdots, \fr_T}$ denote a sequence of $T$ frames from viewpoint $v$ (we omit $v$ when referring to all views jointly). A TraySim scene consists of a circular tray of known mass and radius containing up to $K$ objects, drawn from a set of types $\objectset$ (e.g., bottles, glasses). The $k$-th instance is denoted $\obj_k$, with instances arranged approximately on a square grid.

Let the unobservable physical attributes be denoted $\phy_c$ for all object types $c\in\objectset$, e.g., $\phy_c=\set{\text{sliding-friction, stiffness, damping, inertia}}$ for TraySim, but includes other physics parameters in other datasets we propose subsequently. We assume instances of the same object type have the same values for $\phy_c$. The tray is impacted by a pusher $\pusher$ moving towards the tray center. Each object instance in the setup differs in their visual attributes such as color, shape, or location. Thus, for a scene, there are $|\phy_c||\objectset|$ unknown continuous physical attributes and $|C|K.M$ discrete layout parameters for a color palette of $M$ with no repeats to be estimated for constructing an ideal digital twin, which simplifies to $|C|K^2$ as we use $M=K$. We use $K=9$ and $|C|=3$ in TraySim and each scene has a minimum of 5 object instances, making our optimization landscape non-trivial for reasoning.

With the notation above, we are now ready to formally state our problem. In our setup, we define an input task instance as: $\task=(\set{\gfrv}_{v\in{|\viewset|}}, \pvel, \question, \options, \incontext, \pvideo, \tobjectset)$, where $\gfr$ is the query image of a video sequence $\video$ with views $\viewset$; $\pvel$ is the initial velocity of the pusher $\pusher$; $\question$ is the task description; and $\options$ is the candidate answer set. The goal of the reasoning agent is to predict the correct subset $\answerset \subset \options$. The subset of object types used in $\task$ is denoted by $\tobjectset \subseteq \objectset$. In this work, we assume the question is fixed across all task instances, namely: “Which of the object instances on the tray will remain steady when impacted by the pusher with velocity $\pvel$?” (see Figure~\ref{fig:qual_results}), where $\pvel$ changes across task instances. The prompts and program snippets exposing the simulator APIs to the LLM are denoted by $\incontext$. As show in Figure~\ref{fig:example_scene}, since dynamical parameters cannot be inferred from a single multi-view query image, we also provide a video sequence $\pvideo$ for each task instance. This auxiliary video contains the same set of object types as in the query image $\gfr$, but arranged in a different layout and possibly with fewer objects, and impacted by a pusher with a velocity different from $\pvel$. Each task instance thus has a unique set of attributes. As no training data is provided, we consider the problem in a fully unsupervised, zero-shot setting.

\subsection{Combining LLMs and Physics Engines}
In contrast to prior methods~\cite{wu2017learningB,wu2017neuralA,baradel2019cophy}, our goal is to approach scene de-rendering by leveraging the physics knowledge implicit in state-of-the-art LLMs. Since LLMs do not natively interface with physics engines, yet the outcome of each reasoning task critically depends on inferring both observable and latent physical parameters, two central challenges arise: (i) how to enable an LLM to access a simulator, and (ii) how to harness its embedded physics knowledge.

Building on recent advances in LLM-based program synthesis~\cite{austin2021program,jain2022jigsaw}, our key idea is to frame the reasoning problem such that an LLM can design experimental interventions—programs that configure a physics engine to construct a faithful simulation of the task instance. To this end, we introduce $\llmphy$, which bridges LLMs with physics simulators via dedicated APIs. The LLM generates programs that pass parameter estimates to the simulator, executes them, and receives simulation error feedback for refinement.  Why use programs? There are three key reasons: i) programs provide a compact way to encode varied parameters without redundancy, ii) they enable direct access to physics engine features through simulator APIs, and iii) they associate parameters with explicit variable names, making values interpretable by LLMs (e.g., declaring sliding\_friction = 0.4 gives the LLM a semantic handle to adjust this parameter across iterations using its physics knowledge—such as lowering friction if an object slides instead of falls).

As illustrated in Figure~\ref{fig:arch}, $\llmphy$ processes a task instance $\task$ in two phases: (i) physical parameter estimation, using the auxiliary video $\pvideo$ to infer object-type attributes, and (ii) layout estimation, using the initial multi-view frames ${\gfrv}_{v\in|\viewset|}$ to recover scene layout. Both phases rely on LLM–simulator loops but emphasize different LLM capabilities, as detailed below. Once parameters are estimated, forward simulation reconstructs the scene from the initial frame $\gfr$, and the final poses of object instances are extracted. Upright objects are then matched against the candidate answers $\options$ to produce the task solution $\answerset$.

\noindent\textbf{Phase 1-Inferring Physical Parameters:}
Given a task instance $\task$, $\llmphy$ uses the task video $\pvideo$ to infer the physical attributes $\phy_c$ of the object types $\objectset_\task$ present in $\pvideo$. Let $\traj:\videoset\to\reals{3\times T\times|\objectset_\task|}$ extract 3D trajectories of object centers from $\pvideo\in\videoset$, where $\videoset$ is the set of all videos. In practice, $\tau$ could be implemented with a 3D tracker, though in our experiments we extract trajectories directly from the simulator when generating the dataset.

Let $\llm_1$ denote the Phase 1 LLM, which takes as input a prompt $\incontext_1\in\incontext$ and the object trajectories from $\pvideo$, and outputs a program $\program(\phy)\in\programset$, where $\programset$ is the set of all programs and $\program(\phy)$ denotes the generated program $\pi$ embedding the physics attributes $\phy$ (see Figure~\ref{fig:arch} for an example snippet). A physics simulator $\simu:\programset\to\reals{3\times T\times|\objectset_\task|}$ then executes $\program(\phy)$, producing simulated object trajectories. The goal of Phase 1 is thus to optimize $\program(\phy)$ so that simulated trajectories match those from $\pvideo$, given by:

\begin{equation*}
\phy^*\!\coloneq\!\arg\min_\phy \enorm{\llmphy_1\bigl(\program(\phy)\ |\  \traj(\pvideo), \incontext_1\bigr) - \traj(\pvideo)}^2\!\!,
\label{eq:phase1_eq}
\end{equation*}
where $\llmphy_1\coloneq\simu\circ\llm_1 $ is the composition of the simulator and the LLM through the generated program, for estimating the physical attributes $\phy^*$. 

\noindent\textbf{Phase 2-Simulating Task Examples:}
This phase solves the perception task to recreate the scene given in the multi-view query frames $\gfr$, i.e., finding the location, type, and color of objects. Let $\llm_2$ denote a large vision-and-language model (LVLM) used in Phase 2 that inputs $\gfr$ and a prompt $\incontext_2\in\incontext$ to produce a program $\program(\layout)\in\programset$ where $\Psi :=\set{(\text{type}, \text{location}, \text{color})_k}$ is the set of layout triplets for each object $\obj_k$ that the LVLM infers to be in $\gfr$. The objective for estimating $\Psi$ is then written as:
\begin{equation*}
\Psi^* := \arg\min_\layout \psnr{ \llmphy_2(\program(\layout) | \gfr, \incontext_2),\ \gfr},
\end{equation*}
where $\llmphy_2:=\simu \circ \llm_2$. As we compare $\gfr$ with images produced using $\Psi$, we may use any image similarity metric to compute the loss; we use PSNR.

\noindent\textbf{Forward Simulation and Solution:} Once the physics parameters $\phy^*$ and layout attributes $\layout^*$ are estimated, the query image $\gfrv$ is animated via forward simulation to produce a sequence $\hat{\video}\ |\ \layout^*, \phy^*$. The final state of each object is then obtained—either from the simulator or by parsing the last frame of $\hat{\video}$—to derive the answer set $\answerset$ from the candidates $\options$ (we use the simulator for convenience).

\subsection{Optimizing LLM-Simulator Combination}
In Alg.~\ref{alg:algo}, we detail the steps for optimizing $\llmphy$. As we do not assume any differentiability properties for the  simulator, we cast the objective in a black-box setting, in which the LLM generates parameter estimates through synthesized programs over multiple trials, which are validated using the output from the simulator. The resulting errors are used to refine LLM’s parameter search. In order for the LLM to know the history of its previous parameter choices and the respective errors, we augment the prompt with the program trace from each step. Since LLMs are trained on internet-scale datasets, they may implicitly capture physical principles and commonsense rules. Our insight is to   leverage this prior knowledge through reconstruction-error-guided program synthesis to generate effective descent directions. 

How are the optimization descent directions produced by the LLM for convergence? Our key insight is to tap into the knowledge encoded in 
We note that the optimization variables used in our setup have physical interpretations in the real world. Thus, our idea is to leverage the LLM knowledge about these variables towards seeking better values for them after observing the outcome of the current and past choices. To enable such physical interpretation, we name the variables using their physical names in the generated programs (see Figure~\ref{fig:arch}) that are provided to the LLM in the feedback from the simulator. We see from the program traces (provided in the Appendix~\ref{sec:trace_phase1}) that the LLM is indeed using the physical meanings associated with these variables in efficiently optimizing our objectives.

\begin{algorithm}
\caption{Pseudo-code describing the key steps in optimizing $\llmphy$ for phases 1 and 2.}\label{alg:algo}
    \begin{algorithmic}
        \REQUIRE $\video, \out$ \COMMENT{$\video$ \AC{is the input data, and $\out$ is the desired result, e.g., trajectory, layout, etc.}}
        \STATE $\prompt \gets \textbf{'task prompt'}$ \COMMENT{We assume here a suitable prompt for the LLM.}
        \FOR{$i = \text{1 to max\_steps}$}
            \STATE $\pi \gets \llm(\video, \incontext, \prompt)$ 
            \STATE \COMMENT{ Generated program $\pi$ is assumed to have the optimization variables.}
            \STATE $\hat{\out} \gets \simu(\pi)$ \COMMENT{$\simu$ reproduced result from $\pi$.}
            \STATE $\error \gets \enorm{\out - \hat{\out}}^2$ \COMMENT{or $\PSNR$ in Phase 2}
            \IF{$\error \leq \epsilon$}
                \STATE $\textbf{Return}$ $~\pi$
            \ELSE
                \STATE $\prompt\gets \concat(\prompt, \pi, \textbf{``Error ='' }, \error)$ 
                
            \ENDIF    
        \ENDFOR
    \end{algorithmic}
\end{algorithm}

\section{EXPERIMENTS AND RESULTS}
Our goal in this work is to introduce diagnostic physical-reasoning primitives that admit unambiguous ground truth, while stressing latent parameter inference under partial observability. Motivated by these factors, we present three datasets to evaluate the efficacy of our method, namely TraySim, CLEVRER-LLMPhy, and Real-TraySim. In all these datasets, while their reasoning outcomes are simple (e.g., fall/not fall, collide/not collide), the underlying inference problem is not: with multiple objects and contact configurations, the space of stability/collision outcomes is combinatorial, and correct prediction requires estimating unobservable scene properties rather than relying on video pattern recognition alone.

\noindent\textbf{Simulation Setup:} We use MuJoCo~\cite{todorov2012mujoco} to simulate TraySim tasks and compute rigid-body interactions. The simulator is exposed to the LLM only through Python API calls, where parameters and layouts are passed as program variables. Thus, all simulator details are abstracted from the LLM. The environment is built on a Newtonian physics template with gravity and other initial conditions (see Appendix).

\noindent\textbf{TraySim Dataset:} We generate 100 task sequences (a.k.a problems) using object classes $\objectset={\text{wine glass}, \text{martini glass}, \text{bottle}}$ arranged approximately on a $3\times3$ tray grid. The object classes and the number of object instances in a problem are sampled randomly, with each auxiliary sequence containing at least one instance of every class present in the corresponding task image. Each object instance is uniquely defined by its (color, type, location), and shares physical parameters between the task and its corresponding auxiliary sequence. Physical parameters are randomly sampled per problem task. Specifically, for each problem, we sample a set of physical parameters for each object type; all instances of the same type within that problem share these parameters. For example, a bottle may have different friction coefficients across different problems, but all bottle instances within a given problem share the same friction. Each sequence is simulated for 200 steps with a timestep of 0.01s, and the final frame is used to determine object stability, i.e., if its major axis makes an angle greater than 45 degrees to the direction of the gravity vector. For evaluation, five objects are sampled to form a multiple-choice set, where the correct answer corresponds to the subset of objects that remain upright.

\begin{figure}[t]
    \centering
    \includegraphics[width=8cm,trim={0cm 13.5cm 9cm 3cm},clip]{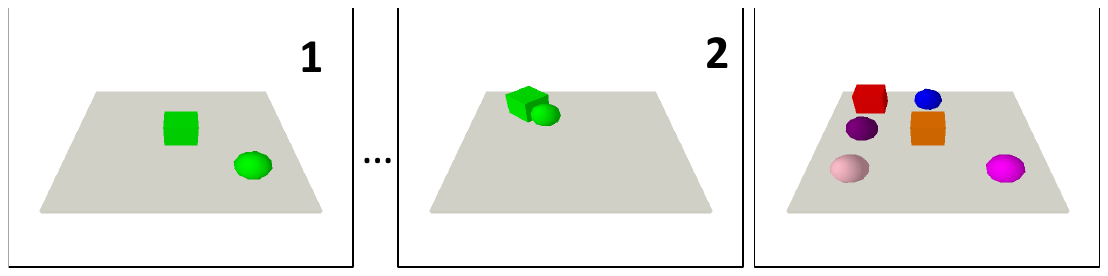}  % Replace with your image file
    \vspace*{-0.3cm}
    \caption{CLEVRER-LLMPhy auxiliary sequence (marked 1-2) and query image.}
    \label{fig:clevrer-samples}
    \vspace*{-0.5cm}
\end{figure}

\noindent\textbf{CLEVRER-LLMPhy Dataset:} Inspired by CLEVRER~\cite{yi2019clevrer}, we developed CLEVRER-LLMPhy for physical reasoning by implementing API interfaces for LLM–simulator interaction. It resembles TraySim but requires multi-step reasoning over simpler geometries. The dataset uses two object types—spheres and cubes—with distinct attributes: mass, sliding friction, rolling friction, and restitution. To demonstrate simulator generalization, CLEVRER-LLMPhy is implemented in PyBullet. As in TraySim, the task is to estimate physics parameters and layout from the auxiliary video and multi-view query frames (see Figure~\ref{fig:clevrer-samples}), however unlike TraySim, in CLEVRER-LLMPhy our goal is not to determine stability, instead is to select objects that are involved in collisions when a given sphere is set in motion. The dataset has 35 problems, each with a 100-frame auxiliary sequence sampled at 1/240s. 

\noindent\textbf{Large Language Model and Phases:} We evaluate a range of models: closed (o1, o1-mini, o3-mini, GPT-4o, GPT-4.1) and open (Gemma3-12B, Phi-4-3.4B). Phase 1 inputs include (i) a prompt describing the task, (ii) program and trajectory examples introducing APIs, and (iii) auxiliary sequence trajectories. Phase 2 uses an example program introducing Phase 2 APIs with multi-view images, enabling the LVLM to learn the mapping between code and image content, and then generate programs for task images.

\noindent\textbf{LLMPhy Feedback Settings:} In Phase 1, we compute trajectory reconstruction error by executing the LLM-synthesized program (with estimated physics parameters) in the simulator to generate object trajectories. Trajectories are sampled every 10 steps, and the L2 norm between input and reconstructed trajectories is measured. We allow up to 30 LLM–simulator iterations and select the parameters from the best reconstruction error. In Phase 2, we use PSNR on the reconstructed initial frame, with at most 5 $\llmphy$ iterations. To reduce LLM calls, we terminate early when the trajectory error drops below 0.1 (Phase 1) or when PSNR exceeds 45 dB (Phase 2). 

\noindent\textbf{Evaluation Metric and Baselines:} We evaluate performance using multiple metrics. Our primary measure is mean intersection-over-union (mIoU), which computes the overlap between $\llmphy$’s Phase 2 predictions and the ground-truth answer set. We also assess instance localization on the tray, which is critical for accurate simulation. Since this is a new task with no prior LLM–physics approaches, we compare against standard continuous black-box optimization baselines: Bayesian optimization~\cite{mockus1974bayesian} and covariance matrix adaptation evolution strategy (CMA-ES)~\cite{hansen2016cma,hansen2001completely}. 

\begin{table}[t!]
\centering
\resizebox{0.5\textwidth}{!}{
\begin{tabular}{c|c|c|c}
\# & Phase1 & Phase2 & Acc (\%) \\
\hline
0  & Human                  & Human                    & 54.2 $\pm$ 3.5 \\
1  & Random                 & GT                       & 50.8 $\pm$ 8.1 \\
2  & Intuitive Physics      & GPT-4o                   & 32.1 $\pm$ 4.5 \\
\hline
3  & Best-of-K              & GT                       & 72.0 $\pm$ 3.5 \\
4  & CMA-ES~\cite{hansen2016cma}                 & GT                       & 75.8 $\pm$ 4.4 \\
5  & LLMPhy (o1-mini)       & GT                       & 77.5 $\pm$ 3.2 \\
6  & LLMPhy (Gemma3)    & GT                       & 77.3 $\pm$ 5.1 \\
7  & LLMPhy (Phi-4)    & GT                       & \textbf{79.8} $\pm$ 3.1 \\
\hline
8  & GT                     & GPT-4o                   & 65.1 $\pm$ 3.0 \\
9  & GT                     & GPT-4.1                  & \textbf{68.3} $\pm$ 2.3 \\
\hline
10 & CMA-ES                 & LLMPhy (GPT-4o)          & 59.7 $\pm$ 3.7 \\
12 & Bayes Opt.~\cite{mockus1974bayesian}             & LLMPhy (GPT-4o)          & 59.6 $\pm$ 2.9 \\
13 & LLMPhy (o1-mini)       & LLMPhy (GPT-4o)          & 62.0 $\pm$ 2.1 \\
14 & LLMPhy (o1-mini)       & LLMPhy (GPT-4.1)         & 66.0 $\pm$ 4.2 \\
15 & LLMPhy (o3-mini)       & LLMPhy (GPT-4.1)         & \textbf{66.3} $\pm$ 3.5\\
\end{tabular}
}
\vspace*{-0.3cm}
\caption{TraySim: Comparisons over methods that can be used in $\llmphy$ phases.}
\label{tab:main_results}
% \end{table}
\vspace{0.3cm}

% \begin{table}[ht]
\resizebox{0.5\textwidth}{!}{
\begin{tabular}{c|l|l|l}
    Phase 2	& (C)olor+(T)ype	& (L)oc+T& C+L+T\\
    \hline
GPT-4o (1 iter)	& 50.0 &	49.3	& 36.8\\
GPT-4o (5 iter)	& 68.7 ± 5.9	&66.3 ± 1.7	&56.0 ± 3.2\\
Gemma3 (5 iter)	& 11.2	&24.6	&5.0\\
GPT-4.1 (5 iter)& \textbf{77.2} ± 5.4	&\textbf{77.1} ± 5.0	&\textbf{66.7} ± 1.6\\
\end{tabular}
}
\vspace*{-0.3cm}
\caption{TraySim: Accuracy (in \%) of inferring the color, type, and location of objects from the Phase 2 images using LVLMs over $\llmphy$ iterations.}
\label{tab:phase2_results}
\vspace{0.3cm}
% \end{table}
% \begin{table}[ht]
\resizebox{0.5\textwidth}{!}{
\begin{tabular}{c|c|c|c}
    Parameter &	LLMPhy/Phase1 &	Bayesian &	CMA-ES\\
    \hline
    Sl-Friction	& \textbf{0.30} ± 0.20	& 0.51 ± 0.29	& 0.35 ± 0.24\\
    Inertia	& \textbf{0.30} ± 0.22	& 0.48 ± 0.29	& 0.32 ± 0.24\\
    Stiffness	& \textbf{0.32} ± 0.24	& 0.49 ± 0.28	& 0.33 ± 0.25\\
    Damping &	2.87 ± 2.25	& 3.37 ± 2.41	& \textbf{2.67} ± 1.78\\
\end{tabular}
}
\vspace*{-0.3cm}
\caption{TraySim: Comparisons of estimated physics parameters using $\llmphy$ against other methods.}
\vspace*{0.3cm}
\label{tab:phy_results}
    \centering
    \resizebox{0.45\textwidth}{!}{
    \begin{tabular}{c|c|c|c}
    Phase 1 & Phase 2 & mIoU (\%) & Precise (\%) \\
    \hline
    Random & GT & 67.0 & 81.0 \\
    BO & GT & 91.7 & 77.8 \\
    CMAES & GT & 85.2 & 77.8 \\
    LLMPhy & GT & \textbf{97.2} & \textbf{88.9} \\
    LLMPhy & LLMPhy & \textbf{94.4} & \textbf{88.9} \\
    \end{tabular}
    }
    \vspace*{-0.3cm}
    \caption{Performance comparison of Phase 1 and Phase 2 methods on CLEVRER-LLMPhy.}
    \label{tab:clevrer-results}
    \vspace*{-0.5cm}
\end{table}

\begin{figure}[ht]
    \centering
    \includegraphics[width=7cm,trim={0cm 5.2cm 6cm 3cm},clip]{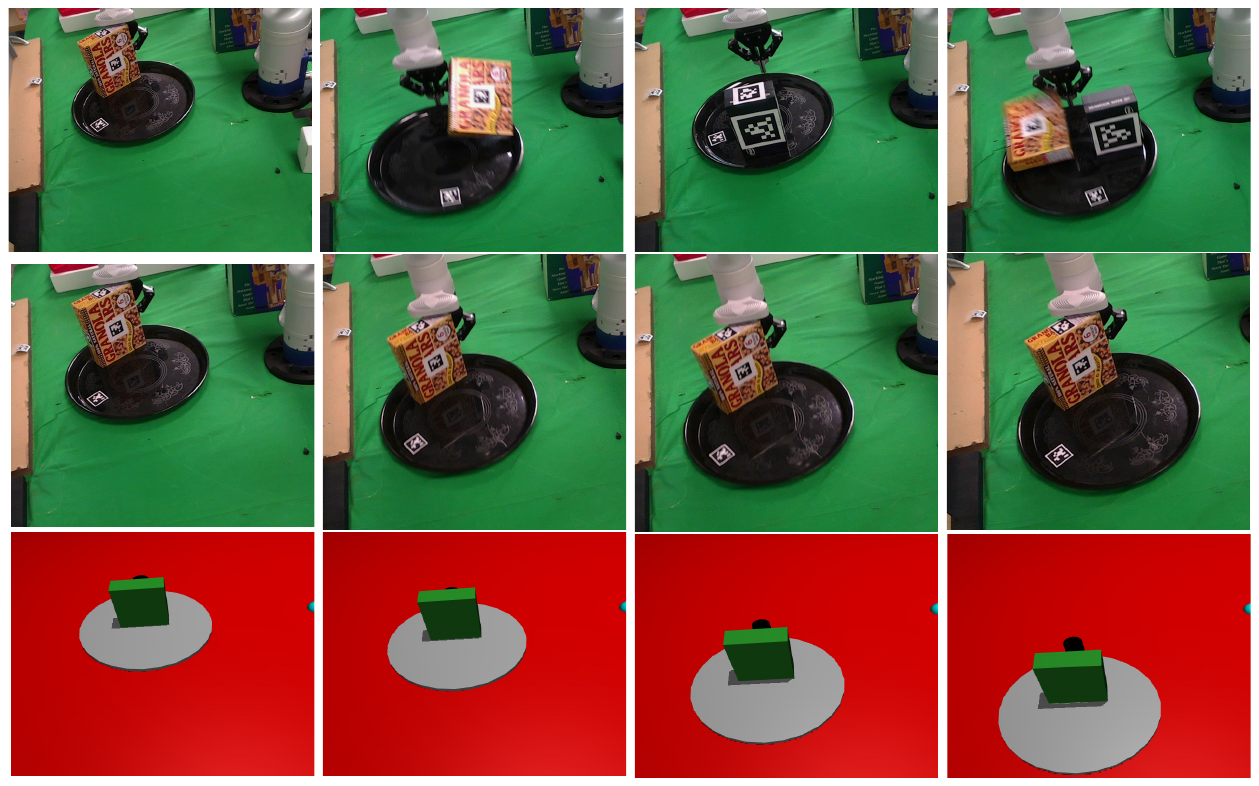}
    \caption{Top: Frame from several of our real-world robotic push sequences. Middle: Frames from a sequence. Bottom: Prediction of box motion in the simulator using LLMPhy estimated physics.}
    \label{fig:real_results}
\end{figure}

\begin{figure*}[ht]
    \centering
    \subfigure[TraySim: Phase 1]{\label{fig:diff_llm}\includegraphics[width=4.5cm,trim={0.5cm 0cm 2cm 1.5cm},clip]{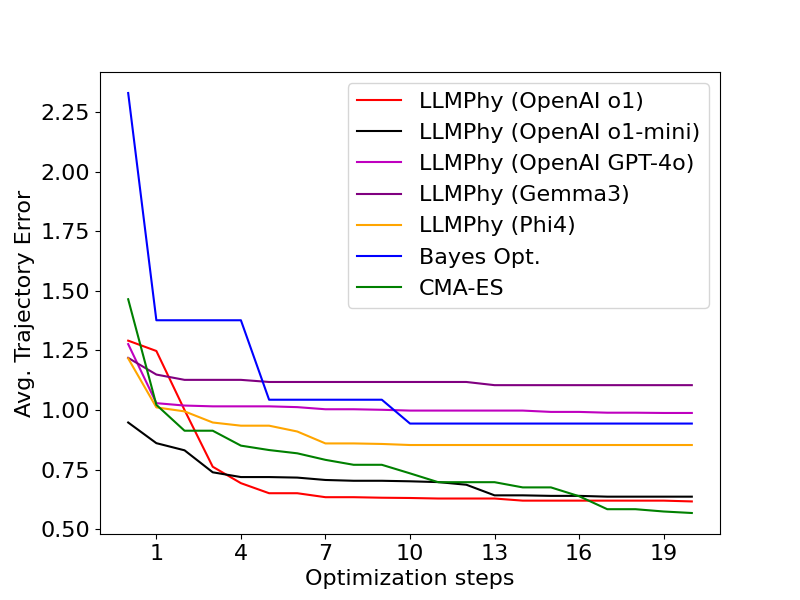}}
    \subfigure[TraySim: Phase 2]{\label{fig:var_p2_appx}\includegraphics[width=4.5cm,trim={0.3cm 0cm 2cm 1.5cm},clip]{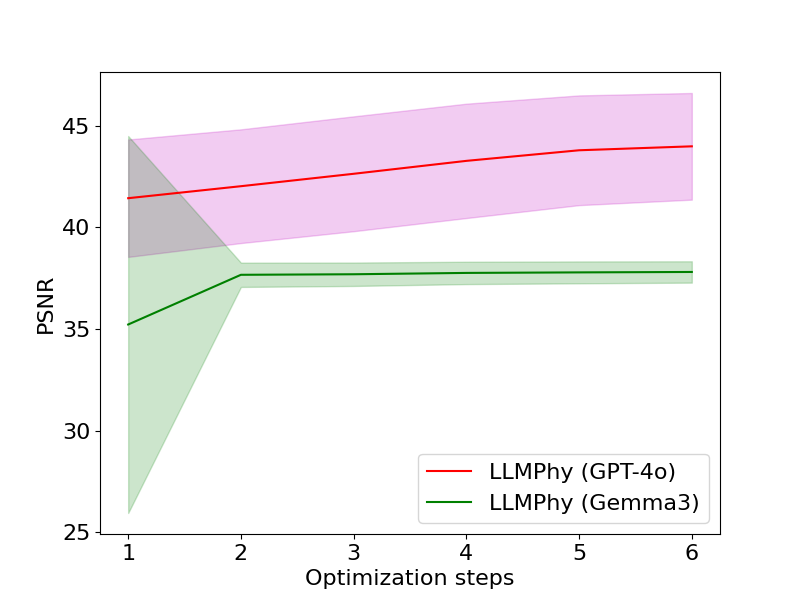}}\\
    \subfigure[CLEVRER: Phase 1]{\label{fig:clevrer_phase1_plot}\includegraphics[width=5.3cm]{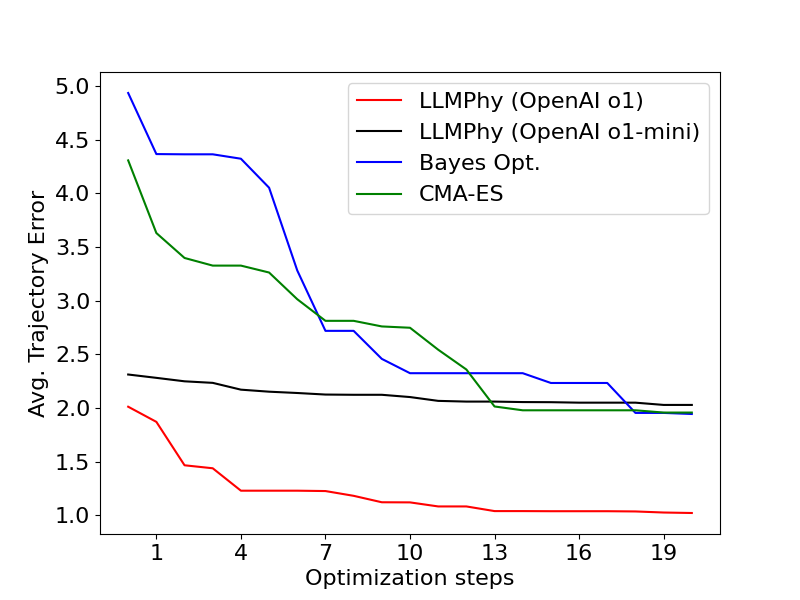}}
    \subfigure[Real-TraySim]{\label{fig:real_analysis}\includegraphics[width=4.5cm]{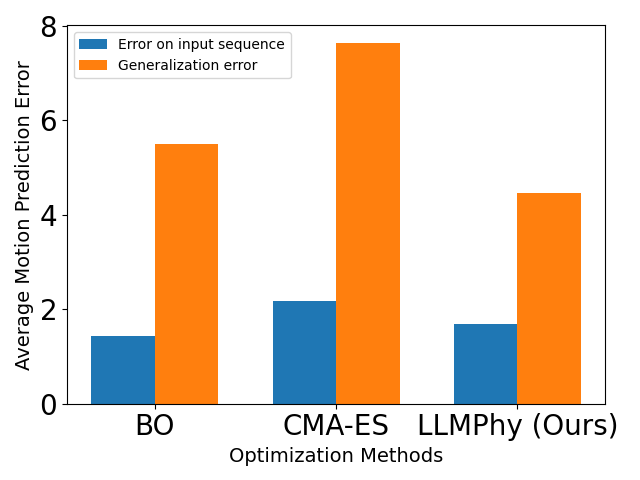}}
    \caption{\AC{(a, c) Convergence comparisons using $\llmphy$ on TraySim and CLEVRER-LLMPhy tasks. We plot the \emph{minimum loss computed thus far} against the number of optimization steps. (b) shows the convergence of $\llmphy$/Phase 2. (d) Motion trajectory prediction error on the real-world input and test sequences. See Appendix for full convergence plots.}}
    \label{fig:error-Phase1}
\end{figure*}

\noindent\textbf{Comparisons to Prior Methods:} In Table~\ref{tab:main_results}, we compare the performance of Phase 1 and Phase 2 of $\llmphy$ against various alternatives and prior black-box optimization methods on the TraySim task. We begin with three baselines: (i) six human subjects (ages 20–50) who were shown the videos and asked to predict the outcomes on the query images, (ii) random sampling of physics parameters while using ground-truth Phase 2 parameters, and (iii) an LVLM directly inferring outcomes from images using intuitive physics. Interestingly, human performance was relatively modest (54.2\%), while random sampling averaged around 50\% with high variance. When the LVLM bypassed the video and attempted to predict outcomes directly from images, performance dropped further to 32\%. Collectively, these baselines confirm that TraySim is a challenging benchmark requiring deep physical reasoning to achieve strong performance.

In experiments 3–7, we evaluate Phase 1 performance of $\llmphy$ in estimating physics parameters against several alternatives. For Best-of-K, each parameter range was discretized into 5 bins, yielding 1,875 combinations followed by sampling 30 times w/o replacement and selecting the one with the lowest simulation error. We also compare to CMA-ES~\cite{hansen2016cma} and Bayesian optimization (BO)~\cite{mockus1974bayesian}, where the simulation error is used as the objective; all models run for the same iteration budget. Unlike $\llmphy$, these methods do not leverage physics knowledge to guide the parameter search.  As these prior methods are computationally cheaper, we also conducted experiments in which both CMA-ES and Bayes-Opt ran for 100 iterations, beyond which we did not see any improvements. With 100 iterations, CMA-ES and Bayes-Opt show an accuracy of 60.7\% and 61.0\% respectively, against 66+\% of LLMPhy. In addition, we test $\llmphy$ with three LLMs: o1-mini, Gemma3, and Phi-4. Across all cases, $\llmphy$ outperforms the prior methods, demonstrating the value of physics-informed reasoning. Results with all three LLMs are consistently stronger, with Phi-4 (trained for mathematical reasoning) surpassing CMA-ES by nearly 4\%. 

In experiments 8 and 9, we isolate Phase 2 performance using GPT-4o and GPT-4.1; the latter (stronger) model yields higher accuracy, suggesting that future advances in LVLM may further improve performance. Table~\ref{tab:phase2_results} provides a detailed comparison with Gemma3, GPT-4o, and GPT-4.1, analyzing the accuracy on the three perceptual parameters. Here, iterative refinement in $\llmphy$ leads to nearly 20\% improvement over 1–5 iterations, clearly demonstrating the benefit of discrete optimization in Phase 2. Finally, experiments 10–15 examine the joint performance of Phase 1 and Phase 2. Compared to Bayesian optimization and CMA-ES, $\llmphy$ improves accuracy by roughly 7\% (59.7\% with CMA-ES vs. 66.3\% with the o3-mini+GPT-4.1 combination). We also observe that more advanced reasoning models (e.g., o3-mini+GPT-4.1) deliver substantial improvements over older model pairs (e.g., o1-mini+GPT-4o), further reinforcing the advantage of stronger LLM–LVLM combinations. 

In Table~\ref{tab:clevrer-results}, we present results on CLEVRER-LLMPhy. As is clear from the table, $\llmphy$ demonstrates clear advantage over other methods, showing the generalizability of our approach to varied simulators and tasks.

\begin{figure}
    \centering
    \includegraphics[width=8cm,trim={1cm 13.5cm 1.5cm 3.1cm},clip]{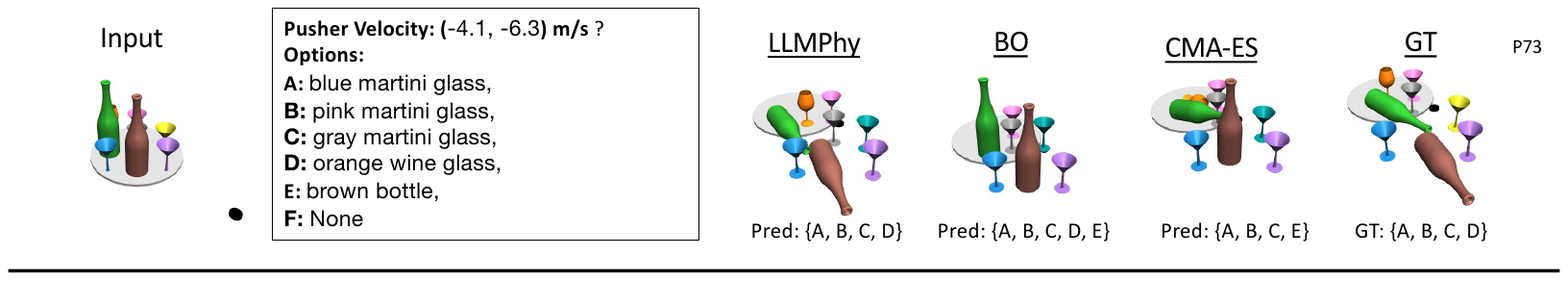}
    \includegraphics[width=8cm,trim={1cm 13.5cm 2cm 3cm},clip]{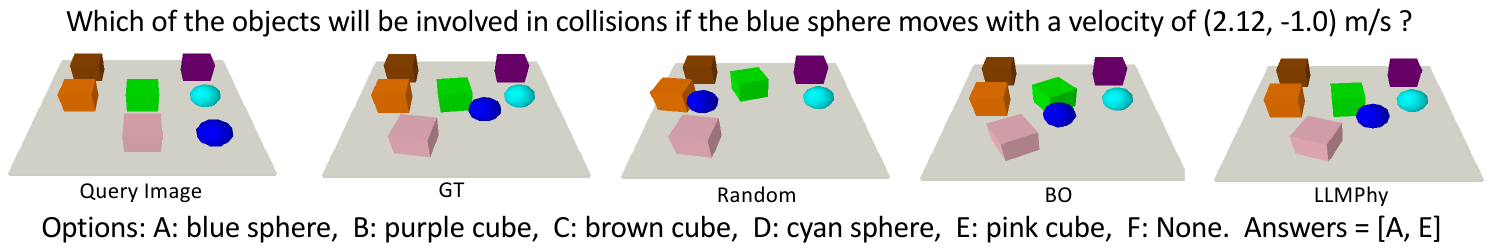}
    \vspace*{-0.3cm}
    \caption{Qualitative comparisons from TraySim (top) and CLEVRER-LLMPhy (bottom) datasets. }
    \label{fig:qual_results}
    \vspace*{-0.5cm}
\end{figure}

\noindent\textbf{Convergence and Correctness:} 
\AC{Figures~\ref{fig:diff_llm} and~\ref{fig:clevrer_phase1_plot} show the convergence of black-box optimization with $\llmphy$ variants compared to prior methods. Strong reasoning LLMs (e.g., o1, o1-mini) converge faster than multimodal models (e.g., GPT-4o), often identifying good parameter values early, while longer CMA-ES runs may eventually catch up on TraySim. Table~\ref{tab:phy_results} reports mean absolute errors between predicted and ground-truth physics parameters, showing that $\llmphy$ consistently achieves lower errors—supporting our claim that LLMs leverage their embedded physical priors to propose plausible parameter values, enabling both faster and better convergence. Figure~\ref{fig:var_p2_appx} further illustrates Phase 2 refinement, where $\llmphy$ iterations improve PSNR between synthesized and task images, with correctness increasing over iterations. Unlike BO and CMA-ES, which are restricted to continuous optimization, $\llmphy$ naturally handles both continuous and discrete spaces through program synthesis.} Finally, Figures~\ref{fig:qual_results} present qualitative results from TraySim and CLEVRER-LLMPhy, compared against other methods.

\noindent\textbf{Real-World Experiments:}
To demonstrate real-world applicability, we conduct experiments with a robotic hand pushing a tray of boxes at varying speeds (0.1–0.4 m/s), where high-speed pushes may cause boxes to fall. Scenes are captured with a single RGB camera. The task is to estimate tray–object physics from one RGB video and transfer the inferred parameters to other sequences of the same objects under different push speeds. Unlike our simulator-based experiments that use ground-truth trajectories, here we approximate object trajectories using computer vision methods (see Appendix). For these experiments, we assume 3D object trajectories in global coordinates can be extracted using standard computer vision  tools, and approximate 3D object CAD models and camera intrinsics/extrinsics are available to construct a digital twin using the simulator. Further the known camera extrinsics are used to scale any depth estimated to world/global coordinates. Object locations (i.e., their center of gravity) are assumed known, and $\llmphy$ infers five attributes per object: mass, friction, damping, inertia, and stiffness. Each sequence lasts 3 seconds. These practical engineering assumptions help isolate evaluation of the physical parameter estimation performance of our method from unrelated scene-specific details. 

Figure~\ref{fig:real_results} (top row) shows sample frames from real sequences, while Figure~\ref{fig:real_analysis} compares trajectory errors. $\llmphy$ performs comparably to BO and CMA-ES on the input sequences but achieves lower generalization error when parameters estimated from one sequence are applied to others with different push speeds. Figure~\ref{fig:real_results} also illustrates a real sequence alongside its digital twin reconstructed by $\llmphy$. See Appendix for details. 

% camera-ready additions.
For the real-world experiments, obtaining true ground-truth values for all physical parameters is inherently challenging. Even if some parameters can be independently estimated (e.g., mass via a scale, friction via tilt tests), the overall system remains highly nonlinear, and many different parameter combinations can result in indistinguishable trajectory outcomes. This ambiguity is precisely why, for Real-TraySim, we focus on predicting object dynamics under varied configurations, rather than evaluating absolute parameter-identification accuracy. To this end, in Table~\ref{tab:real-model-params} we report the five physical parameters inferred by our method and by the baselines. While ground truth is not available for all parameters, the true mass of the box is approximately 18 grams, and we observe that LLMPhy variants estimate this value more accurately than CMA-ES (which underestimates) or Bayesian Optimization (which significantly overestimates). This suggests that the LLM-based approach can leverage implicit physical priors to infer more plausible parameter values, even in the absence of full ground truth.

\begin{table}[h]
\centering
\resizebox{0.5\textwidth}{!}{%
\begin{tabular}{lccccc}
\hline
Model & Mass & Friction & Inertia & Stiffness & Damping \\
\hline
CMA-ES & 0.76 & 0.31 & 0.36 & 0.46 & 19.9 \\
Bayes-Opt & 49.1 & 0.29 & 0.79 & 0.39 & 12 \\
LLMPhy (4o) & 8.7 & 0.32 & 0.8 & 0.98 & 3.4 \\
LLMPhy (o1) & 12.0 & 0.4 & 0.7 & 1.0 & 7.0 \\
\hline
\end{tabular}%
}
\caption{Comparison of the five real physical parameters estimated by $\llmphy$ against other methods on the Real-TraySim dataset.}
\label{tab:real-model-params}
\end{table}

\section{CONCLUSIONS}
We studied the problem of predicting the outcomes of complex physical interactions and introduced $\llmphy$, a novel iterative framework that couples large language models with physics engines. By combining the reasoning and program synthesis capabilities of LLMs with the fidelity of physics-based simulation, $\llmphy$ provides a principled way to infer both latent physical parameters and scene layouts, enabling more accurate digital twin construction. Our experiments on TraySim, CLEVRER-LLMPhy, and real-world datasets highlight the effectiveness of this approach, consistently outperforming classical black-box optimization baselines.

\noindent\textbf{LIMITATIONS:} $\llmphy$ reveals important challenges and opportunities. Its performance is strongly dependent on the fidelity of the underlying simulator, indicating that advances in high-accuracy physics engines are likely to yield direct gains for reasoning frameworks of this nature. Moreover, although LLM-driven optimization demonstrates competitive performance, it incurs substantially higher computational overhead than traditional optimization techniques. In particular, relative to classical black-box methods, $\llmphy$ is currently computationally demanding due to its reliance on large language models, which impose nontrivial hardware requirements. Future work may explore efficiency improvements, integration with more advanced simulators, and extending the framework to broader classes of reasoning tasks involving causality, planning, and real-world robotics. While the LLMs we used in this work are not particularly optimized for black-box optimization, we believe finetuning these models for our setup may improve their speed and effectiveness.

\section{IMPACT STATEMENT}
\label{sec:impact}
Our paper seeks to advance the field of machine learning by addressing scenarios in which state-of-the-art AI models are unable to independently solve complex problems. Specifically, we propose integrating well-designed software modules (such as physics engines) as tools for large language models (LLMs), demonstrating significant performance enhancements. Through our study, we present several  observations below with potential implications for AI: 
\begin{enumerate}
\item We find that LLMs, particularly the latest models such as OpenAI's o1, o3, GPT-4.1 etc. exhibit remarkable capabilities in solving physical reasoning problems when integrated with a physics simulator. 

\item  LLMs possess extensive world knowledge that, when accessed through API calls and utilized via specialized tools, can effectively address complex reasoning tasks.

\item Although we did not explore the full spectrum of LLM-based program synthesis capabilities, our preliminary investigation demonstrates that these models possess a clear understanding of objectives and can effectively integrate their inherent physical knowledge to produce physically realizable solutions. We think understanding such capabilities of AI is important as they can have both positive and negative consequences.
 
\end{enumerate}
These observations are crucial for designing setups like ours for real-world physical reasoning applications and the development of physically embodied systems.

\bibliographystyle{apalike}
\bibliography{biblio}
\clearpage
\appendix
\section*{\underline{Appendix: Table of Contents}}
\begin{enumerate}
    % \item Qualitative videos: See the zip attachment.
    % \item Summary of our contributions \ref{sec:summary}
    \item Simulation Setup: Appendix \ref{sec:simulation_setup}
    \item TraySim Dataset: Appendix \ref{sec:traysim_dataset}
    \item Physics Parameter Sensitivity: Appendix \ref{sec:physics_sensitivity}
    \item Details of LLMPhy Phases: Appendix \ref{sec:phase_details}
    \begin{enumerate}
        \item Phase 1 Prompt and Details: Appendix~\ref{sec:phase1_details}
        \item Phase 2 Prompt and Details: Appendix~\ref{sec:phase2_details}
    \end{enumerate}
    \item Performances to Other LLMs: Appendix~\ref{sec:more_expts}
    \item Ablation Studies: Appendix~\ref{sec:ablations}
    \item LLMPhy Detailed Convergence Analysis: Appendix~\ref{sec:convergence_analysis}
    \item Qualitative Results: Appendix~\ref{sec:qual_results_more}
    \item CLEVRER-LLMPhy Dataset and Experiments: Appendix~\ref{sec:clevrer}
    \item Real-World Experiments and Details: Appendix~\ref{sec:realworld}
    % \item Limitations, Impact Statement: Appendix~\ref{sec:limitations}
    \item LLMPhy Optimization Traces: Appendix~\ref{sec:llm_opt}
\end{enumerate}

\section{TRAYSIM SIMULATION: PHYSICS MODELS}
\label{sec:simulation_setup}
As discussed in the main paper, we are determining the physical characteristics of our simulation using a physics engine. MuJoCo ~\cite{todorov2012mujoco} was used to setup the simulation and compute the rigid body interactions within the scene. The simulation environment is build upon a template of the world $\mathcal{W}$, which contains the initial parametrization of our model of Newtonian physics. This includes the gravity vector $\textbf{g}$, time step, and contact formulation, but also graphical and rendering parameters later invoked by the $\llm$ when executing the synthesized program. 

MuJoCo internally uses a soft contact model to compute for instance complementarity constraints; in our implementation we use a non-linear sigmoid function that allows a very small inter-body penetration and increases the simulation stability during abrupt accelerations. We use elliptic friction cones to replicate natural contacts more closely. We further take advantage of the model architecture of MuJoCo by programmatically inserting arbitrary objects $o_k$ from the classes in $\objectset$ into the scene, as described in the main paper. For each parametric object class in $\objectset$, we generate an arbitrary appearance and physical attributes such as static friction, stiffness, damping, and rotational inertia. An arbitrary number of object instances are created from each class (up to a provided limit on their total number) and placed at randomly chosen positions on a regular grid (scene layout). The graphical renderer is used to record the frame sequences $\video$ corresponding to five orthogonally placed cameras around the world origin, including a top-down camera. The simulated data also contains privileged information such as the pusher-tray contact information (\textit{i.e.} force, location, velocity, and time stamp), and the stability information for each object, $\mathcal{S}_k=\{1| \arccos(\textbf{g},O\textbf{z}_k) < \alpha, 0| otherwise\}$, where $\mathbf{g}$ is the gravity vector, $O\textbf{z}_k$ is the upright direction of object $k$ and $\alpha$ is an arbitrarily chosen allowable tilt. Thus, in our experiments, we use $\alpha=45^{\circ}$. Given that we consider only rigid objects with  uniformly distributed mass, we assume that this a reasonable and conservative threshold.

Other than the physics parametrization of each object class $\mathcal{C}$ and the scene layout $\cup o_k$, the outcome of the simulation for sequence $\textbf{X}$ is given by the initial conditions of the pusher $p$, namely its initial velocity $\dot{p}_s$ and position $\textbf{p}_s$. The usual torque representation is used:
\begin{equation}
    \boldsymbol{\tau} = \textbf{I}_c\boldsymbol{\dot{\omega}}+\boldsymbol{\omega}\times\textbf{I}_c\boldsymbol{\omega},
\end{equation}
which relates the angular acceleration $\mathbf{\alpha}$ and angular velocity $\mathbf{\dot \omega}$  to the object $c\in\objectset$ torque $\mathbf{\tau}$ with moment of inertia $\textbf{I}_c$. The simulator computes in the end the motion of each object based on the contact dynamics model given by:
\begin{equation}
     \mathbf{M}(\mathbf{q}) \ddot{\mathbf{q}} + \mathbf{C}(\mathbf{q},\dot{\mathbf{q}}) = \mathbf{S}_a^T \boldsymbol{\tau} + 
     \mathbf{S}_u^T \boldsymbol{\lambda}_{u} +
     \mathbf{J}^T_{c}(\mathbf{q}) \boldsymbol{\lambda}_{c},
     \label{eq:dynamics}
\end{equation}
where $\mathbf{M}(\mathbf{q}) \in \mathbb{R}^{(n_a+n_u) \times (n_a+n_u)}$ is the mass matrix; $\mathbf{q} \triangleq [\mathbf{q}_a^T, \mathbf{q}_u^T]^T\in\mathbb{R}^{n_a+n_u}$ are generalized coordinates; and  $\mathbf{C}(\mathbf{q},\dot{\mathbf{q}}) \in \mathbb{R}^{n_a+n_u}$ represents the gravitational, centrifugal, and the Coriolis term. The selector matrices $\mathbf{S}_a=[\mathbb{I}_{n_a \times n_a} \ \mathbf{0}_{n_a \times n_u}]$ and $\mathbf{S}_u=[\mathbf{0}_{n_u \times n_a} \ \mathbb{I}_{n_u \times n_u}]$ select  the vector of generalized joint forces $\boldsymbol{\tau} \in \mathbb{R}^{n_a}$ for the \textit{actuated} joints $n_a$, or $\boldsymbol{\lambda_{u}}\in \mathbb{R}^{n_u}$ which are the generalized contact forces of the \textit{unactuated} DOF created by the dynamics model, respectively. $\mathbf{J}_{c}(\mathbf{q}) \in \mathbb{R}^{6n_{c} \times (n_a+n_u)}$ is the Jacobian matrix and $\boldsymbol{\lambda}_{c} \in \mathbb{R}^{6n_{c}}$ are the generalized contact forces at $n_c$ contact points. In our simulated environment, only the pusher object $p$ has actuated joints which sets its initial velocity and heading, while the rest of the joints are either unactuated or created by contacts. The state of the system is represented by $\mathbf{s} \triangleq [\mathbf{q}^T \ \mathbf{\dot{q}}^T]^T $.

\section{TRAYSIM DATASET}
\label{sec:traysim_dataset}
Using the simulation setup described in Sec~\ref{sec:simulation_setup}, we created 100 task sequences using object classes $\objectset=\set{\text{wine glass}, \text{martini glass}, \text{bottle}}$ with object instances from these classes arranged roughly in a $3\times 3$ matrix on the tray. The instance classes and the number of instances are randomly chosen with a minimum of 5 and a maximum of 9. Each task sequence is associated with an  auxiliary sequence for parameter estimation that contains at least one object instance from every class of object appearing in the task images. For example, if a task image (that is, the first image in a task sequence) has 3 bottles, then we will have a bottle in the auxiliary sequence. We assume each instance is defined by a triplet: (color, type, location), where the color is unique across all the instances on the tray so that it can be identified across the multi-view images, especially when some views occlude some of the instances. The physical parameters of the objects are assumed to be the same for both the task sequences and the auxiliary sequences, and instances of the same object classes have the same physical parameters. The physics parameters were randomly sampled for each problem in the dataset. We assume the pusher is placed at the same location in both auxiliary and task data; however this location could be arbitrary and different and will not affect our experiments as such locations will be supplied to the simulator in the respective phases and are not part of inference. 

\noindent\textbf{Ground Truth Physics:} When generating each problem instance in the TraySim dataset, the physical parameters of the object classes are randomly chosen within the following ranges: sliding friction in (0.1, 1], inertia and stiffness in (0, 1), and damping in (0, 10). We assume a fixed and known mass for each object type across problem instances, namely we assume a mass of 20 units for bottle, 10 units for martini glass, and 4 units for the wine glass. The tray used a mass of 0.5 and the pusher with a mass 20. Further, for both the task and the auxiliary sequences we assume the pusher is located at the same initial location in the scene. However, for all the auxiliary sequences, we assume the pusher moves with an initial (x, y) velocity of (-4.8, -4.8) m/s towards the tray, while for the task sequences, this velocity could be arbitrary (but given in the problem question), with each component of velocity in the range of [-7, -3] m/s. We further assume that the pusher impact direction coincides with the center of the circular tray in all problem instances. 

\AC{\noindent\textbf{Optimization Space:} We note that each object class has a unique physics, i.e., each object class has its own friction, stiffness, damping, and inertia, which are different from other object classes. However, instances of the same class share the same physics. Thus, our optimization space for physics estimation when using 3 object instances, each one from a unique class, is thus 12. For the Phase 2 optimization, the LLM has to reason over the object classes for each object instance in the layout image, their positions in the $3\times 3$ grid, and their colors. This is a sufficiently larger optimization space, with 10 instance colors to choose from, 3 object classes, and 9 positions on the grid.} 

\begin{figure*}[t]
    \centering
    \includegraphics[width=13cm,trim={0cm 11.5cm 2cm 3cm},clip]{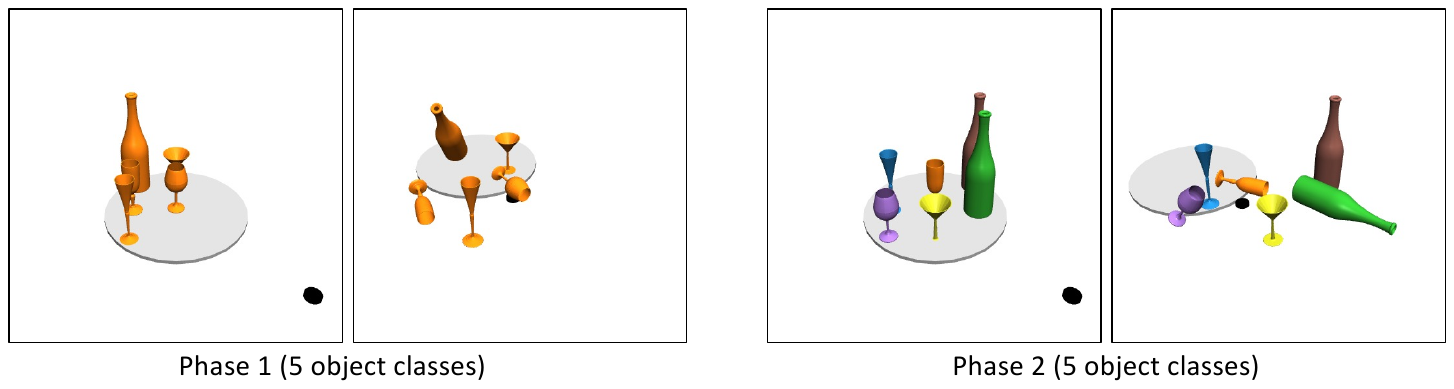}
    \caption{An example illustrating our extended dataset with 5 object classes.}
    \label{fig:qual_add_objs}
\end{figure*}

\AC{\noindent\textbf{Additional Objects:} In addition to the setup above that we use for the experiments in the main paper, we also experiment with additional object classes in this supplementary materials to show the scalability of our approach to more number of parameters to optimize. To this end, we consider two additional object classes, namely: i) \emph{flute\_glass} with a mass of 15.0, and \emph{champagne\_glass}, with again a mass of 15.0. The physics parameters for these classes are sampled from the same range described above. Even when we use these additional classes, the layout uses the same $3\times 3$ matrix for phase 2, however their Phase 1 evaluation has now $5\times 4=20$ variables to optimize instead of 12. We created 10 sequences with these additional objects, as our goal is to ablate on the scalability of our approach, than running on a full evaluation as against the results reported in the main paper. Figure~\ref{fig:qual_add_objs} shows sample frames from our extended setup.}

\noindent\textbf{Simulation and QA Task:} Each sequence was rendered using the simulator for 200 time steps, each step has a duration of 0.01s. We used the last video frame from the task sequence to check the stability of each instance. Specifically, if the major axis of an object instance in the last frame of a task sequence makes an angle of more than 45 degrees with the ground plane, then we deem that instance as \emph{stable}. \AC{We randomly select five object instances and create a multiple choice candidate answer set for the question-answering task, where the ground truth answer is the subset of the candidates that are deemed upright in the last frame. Our QA question is ``Which of the object instances on the tray will remain upright when the tray is impacted by a pusher with a velocity of ($x$, $y$) m/s from the location ($loc_x$, $loc_y$) in a direction coinciding with the center of the tray``. 

As the correct estimation of the physical parameters and spatial layout of our problem setup completely characterizes all the unknown variables, making our forward simulation deterministic (except for the slight stochasticity in the simulator ODE solvers)  simulating more complex interactions post-phase 2 is straightforward. Thus, without any loss of generality, we assume $(loc_x, loc_y)$ are fixed in all cases, although as it is a part of the question and is simulated (and not inferred) any other location of the tray or the pusher will not be an issue when inferring using $\llmphy$. From an evaluation perspective, keeping the pusher too close to the tray may result in all object instances toppling down, while placing it far with smaller velocity may result in the pusher halting before colliding with the tray. Our choice of the pusher velocity was empirically selected such that in most cases the outcome of the impact is mixed and cannot be guessed from the setup.}

\section{PHYSICS PARAMETER SENSITIVITY}
\label{sec:physics_sensitivity}
\begin{figure*}[ht]
    \centering
    \includegraphics[width=14cm,trim={0cm 6cm 1.5cm 3cm},clip]{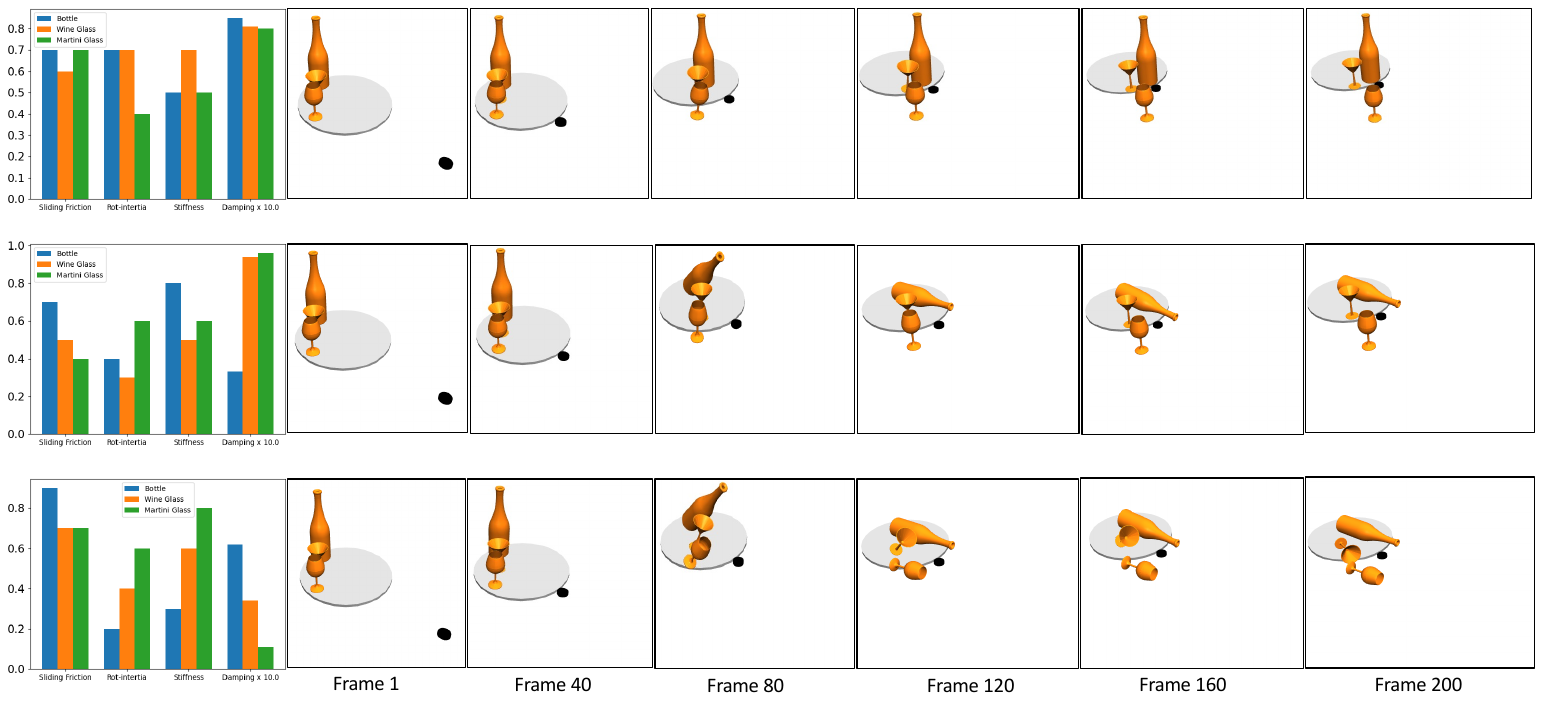}
    \caption{Illustration of the changes in the physical parameters (left histogram, sliding friction, rotation inertia, stiffness, and damping, respectively), and the result of the impact on three objects placed at the same location on the tray (Frame 1) and being impacted by the same force from the pusher. The examples are from the Phase 1 of our dataset. As is clear in the last frame (Frame 200) that changes in the the physical parameters results in entirely different outcomes, substantiating that the correct estimation of these parameters is important.}
    \label{fig:qual_results_phy_params}
\end{figure*}

\AC{A natural question one may ask about the TraySim dataset is \emph{"how sensitive are the physics parameters to influence the outcome?} In Figure~\ref{fig:qual_results_phy_params}, we show three Phase 1 sequences consisting of the same objects and their layout, however varying the physics attributes as shown in the histogram plots. The pusher velocity is fixed for all the sequences. As can be seen from the figure, varying the parameters result in entirely different stability for the objects after the impact, substantiating that the correct inference of these parameters is important to produce the correct the outcome.}

\section{DETAILS OF PHASES}
\label{sec:phase_details}
In this section, we detail the inputs and expected outputs provided in each phase of $\llmphy$.

\subsection{Phase 1 Prompt and Details} 
\label{sec:phase1_details}
\AC{In this phase, we provide as input to the LLM four items: i) a prompt describing the problem setup, the qualitative parameters of the objects (such as mass, height, size of tray, etc.) and the task description, ii) an in-context example consisting of sample trajectories of the object instances from its example auxiliary sequence, iii) a program example that, for the given example auxiliary sequence trajectories, shows their physical parameters and the output structure, and iv)  auxiliary task sequence trajectories (from the sequence for which the physical parameters have to be estimated) and a prompt describing what the LLM should do. The in-context example is meant to guide the LLM to understand the setup, the program structure we expect the LLM to produce, and  our specific APIs that need to be called from the synthesized program. Figure~\ref{fig:prompt-phase-1-preamble} shows the prompt preamble we use in Phase 1. Figure~\ref{fig:prompt-phase-1-example} show sample code, API interfaces, program structure and format of input object centroid trajectories used in Phase 1. Figure~\ref{fig:prompt-phase-1-problem-trajectories} provides problem specific trajectories that the model should generate code for. 
}

When iterating over the LLM predictions, we augment the above prompt with the history of all the estimations of the physical parameters that the LLM produced in the previous iterations (extracted from the then generated code) and the $\ell_2$ norm between the generated and ground truth object trajectories for each object instance in the auxiliary sequence, with an additional prompt to the LLM as follows: ``\textit{We ran your code in our simulator using the physical parameters you provided below... The error in the prediction of the trajectories using these physical parameters is given below. Can you refine your code to make the trajectories look more similar to the ones given in ...? Your written code should strictly follow the same code structure as provided in ...}''. See our optimizaton logs in Appendix~\ref{sec:llm_opt} for examples. While, we may use computer vision methods for estimating the trajectory of motion of the objects in this Phase, i.e., $\tau$ function in the main paper (as done in our real-world experiments provided subsequently), in this current experimental setup, we directly use the trajectories from the simulator for optimization for two reasons: i) we assume the Phase 1 allows complete access to the objects and the setup for parameter estimation, and ii) the focus of this phase is to estimate the physics parameters assuming everything else is known, while the perception task is dealt with in Phase 2. In a real-world setup, we may use AprilTags for producing the object trajectories. This simulation trajectories for Phase 1 will also be provided as part of our TraySim dataset, while also providing the multiview Phase 1 videos for anyone to use vision foundation models for solving the perception problem.

\subsection{Phase 2 Prompt and Details}
\label{sec:phase2_details}
\AC{The goal of the LVLM in Phase 2 is to predict the object instance triplet from the multi-view task images. Towards this end, the LVLM generates code that incorporates these triplets, so that when this code is executed, the simulator will reproduce the scene layout. Similar to Phase 1, we provide to the LVLM an in-context example for guiding its code generation, where this in-context example contains multi-view images and the respective program, with the goal that the LVLM learns the relation between parts of the code and the respective multi-view images, and use this knowledge to write code to synthesize the layout of the provided task images. See Figure~\ref{fig:prompt-phase-2-preamble} for the prompt preamble in Phase 2, followed Figure~\ref{fig:prompt-phase-2-problem} showing the program structure and problem specific instances that the VLM should generate code for. When iterating over the optimization steps, we compute an error feedback to the LVLM to improve its previously generated code, where the feedback consists of the following items: i) the program that the LVLM synthesized in the previous optimization step, ii) the PSNR between the task image and the simulated  image (top-down views), and iii) the color of the object instances in error\footnote{This is done by inputting a difference image (between the task and synthesized images) to another vision-and-language LVLM that is prompted to identify the triplets that are in error}. Using this feedback, the Phase 2 LVLM is prompted to fix the code associated with the triplets in error. Our feedback prompt in Phase 2 thus looks like in the following example: ``\textit{The chat history below shows a previous attempt of GPT-4o in generating Python code to reproduce the task images .... For each attempt, we ran the GPT-4o generated code in our simulator and found mistakes. Below we provide the code GPT produced, as well as the PSNR of the generated image against the given top-down image. Can you refine your code to reproduce the task images correctly? You should not change any part of the code corresponding to correctly inferred objects. $\langle$ code ...$\rangle$. Colors of the objects in the code above that are misplaced: colors = \{'orange', 'purple', 'cyan'\}. PSNR for the generated image against given top-down image = 39.2
Please check the locations of these objects in task\_image\_top\_view\_1.png and fix the code accordingly.}''. } % We show a full prompt for the Phase 2 LLM in Sec.~\ref{sec:phase2_example}.

\section{PERFORMANCES OF OTHER LLMS: o1, Gemma-3, Phi-4}
\label{sec:more_expts}
\AC{In Tables~\ref{tab:other_llm_expts} and ~\ref{tab:expts_o1_appx}, we compare the performances to other LLM choices in Phase 1 of $\llmphy$. Specifically, in Table~\ref{tab:other_llm_expts}, we compare performances when using various open models, including Gemma3~\cite{team2025gemma} and Phi-4~\cite{abdin2024phi}. We used the 12B parameter Gemma3 VL model, and 3.8B parameter Phi-4-mini-instruct text-LLM for our experiments. We also considered other LVLMs including Qwen-VL and intern-VL, however they failed to generate code that we could run in our simulator and thus could not use them in our experiments. We find from Table~\ref{tab:other_llm_expts} that various LLMs in Phase 1 of $\llmphy$ demonstrates strong results against baselines such as CMA-ES. We ran all the LLM experiments for 20 iterations. We used A100 GPUs for running the open-models. It was found that open-model such as Gemma3 and Phi-4 took significantly longer for each iteration, nearly 5-10 minutes per iteration while the time taken by OpenAI models were within 1 minute on average. 

\begin{table}[ht]
\centering
    \begin{tabular}{c|c|c|c}\\
    Expt \# & Phase 1 & Phase 2 & mIoU (\%)\\
    \hline
     1& CMA-ES   & GT & 75.8 \\
     2& $\llmphy$ (GPT-4o)  & GT & 77.5\\
     3&$\llmphy$ (Gemma3)   & GT & 77.3\\
     4& $\llmphy$ (Phi-4) & GT &  79.8 \\
     % 5 & $\llmphy$ (Gemma3) & $\llmphy$ (Gemma3) & 29.0
    \end{tabular}
    \caption{\AC{Performance on TraySim-100 sequences (extended results to Table 1 in the main paper) against other LLMs used in $\llmphy$ Phase 1 and Phase 2. }}
    \label{tab:other_llm_expts}
\end{table}

\begin{table}[ht]
    \centering
    \resizebox{0.5\textwidth}{!}{
    \begin{tabular}{c|c|c|c}\\
    Expt \# & Phase 1 & Phase 2 & Avg. IoU (\%)\\
    \hline
    1 &BO  & $\llmphy$ (GPT-4o) & 49.6\\
    2 & CMA-ES & $\llmphy$ (GPT-4o)  & 53.0\\
    3 & $\llmphy$ (GPT-4o) & $\llmphy$ (GPT-4o) & 53.0\\
    4 &$\llmphy$ (o1-mini) & $\llmphy$ (GPT-4o)& 55.3\\
    5 & $\llmphy$ (o1) & $\llmphy$ (GPT-4o) & \textbf{57.0}\\
    \end{tabular}
    }
    \caption{\AC{Performance analysis (on a small subset of 10 examples) of $\llmphy$ Phase 1 and Phase 2 combinations against various alternatives using various LLMs within $\llmphy$.}}
    \label{tab:expts_o1_appx}
\end{table}
In Table~\ref{tab:expts_o1_appx}, we report the experiments that use various OpenAI models for Phase 1 on a smaller subset of ten problems from the TraySim dataset using 5 object categories (instead of 3 as in Table~\ref{tab:other_llm_expts}. We report only the performance on this subset for all methods. We find that the o1 variant of the models demonstrates better performances against CMA-ES and substantially better than BO. }

\section{ABLATION STUDIES}
\label{sec:ablations}
\AC{In this section, we analyze various aspects of $\llmphy$ performance and is reported in Table~\ref{tab:expts_ablations_appx}. In addition to Avg. IoU performance as done in the main paper, we also report the 'precise IoU' that counts the number of times the predicted answer (i.e., the set of stable object instances listed in the answer options) match precisely with the ground truth. }

\noindent\textit{1. How will $\llmphy$ scale to more number of object types?} To answer this question, we used the extended TraySim dataset described in Sec.~\ref{sec:traysim_dataset} with additional data using five object types, including a flute\_glass and  champagne\_glass. These two items are assumed to have the same mass of 15.0. We created 10 examples with this setup (using five object types) for our ablation study and re-ran all methods on this dataset. We report the results in Table~\ref{tab:expts_ablations_appx} below. In Expt 1-3 of Table~\ref{tab:expts_ablations_appx}, we compare the performance of $\llmphy$ to BO and CMA-ES. We see that LLM performs the best. We also repeated the experiment in Expt 4-6 using the ground truth (GT) Phase 2 layout, thus specifically evaluating on $\llmphy$ Phase 1 physics estimation. Again we see the clear benefit in using $\llmphy$ on both Avg. IoU and Precise IoU, underlining that using more objects and complicating the setup does not affect the performance of our model. We note that all methods in this comparison used the same settings, that is the number of optimization iterations was set to 30, and we used o1-mini for $\llmphy$.

\noindent\textit{2. Robustness of $\llmphy$ Performances?} \AC{A natural question is how well do $\llmphy$ perform when the 3D trajectories are not accurate? We provide experiments testing the robustness of our scheme in a controlled setting by artificially perturbing the trajectories, for example, by injecting noise to the feedback provided to the LLM/LVLM at each iteration. We attempted this route by adding a noise equal to 25\% of the smallest prediction error for each of the object instance trajectories in Phase 1. Specifically, we compute $\ell_2$ error between the predicted and the provided object trajectory for each object class in Phase 1 of $\llmphy$ (let's call it $\set{e_k}_{k=1}^5$), computed the minimum of these errors say $e_m$, and replaced as $\hat{e}_k:=e_k + e_m.\zeta/4.0$ for $k=1,2,\cdots, 5$  and $\zeta\sim\mathcal{N}(0,1)$. This will make the LLM essentially uncertain about its physical parameter predictions, while the error (which is sufficiently high given the usual range of the error is between 0.5-4) simulates any underlying errors from a real physical system or simulation errors when using another physics engine. Our results in Expt. 7-8 in Table~\ref{tab:expts_ablations_appx} show that $\llmphy$ is not very much impacted by the noise. While there is a drop of about 5\% in accuracy (72.5\% to 67.2\%) when using GT, it is still higher than for example, when using CMA-ES on this additional dataset. }

\noindent\textit{3. Advantage of using Optimization Trace?} \AC{One of the differences from prior work such as~\cite{ma2023eureka} is that $\llmphy$ uses the optimization trace against only the last feedback as in an evolutionary search procedure. In Table~\ref{tab:expts_ablations_appx} Expt 9-10, we compare the performance when not using the full optimization trace. We see a drop of 5\% (i.e., 56.4\% Avg. IoU to 51.1\%) showing that the optimization trace is useful. While using the optimization trace may demand longer context windows, we believe it also helps the LLM to avoid reconsidering previously generated parameter values and thus aids better convergence in our setting, especially for black-box optimization approaches, unless there is a provision to include a summary of the optimization trajectory to the LLM in another manner.}

\begin{table*}[ht]
    \centering
    \begin{tabular}{c|c|c|c|c}\\
    Expt \# & Phase 1 & Phase 2 & Avg. IoU (\%) & Precise IoU(\%)\\
    \hline
    1 &BO  & $\llmphy$ & 51.2 & 0.0\\
    2 & CMA-ES & $\llmphy$  & 39.5 & 0.0\\
    3 & $\llmphy$ & $\llmphy$ & \textbf{56.4} &  11.0\\
    \hline
    4 & BO & GT & 71.0 & 11.0\\
    5 & CMA-ES & GT & 63.2 & 22.0\\
    6 & $\llmphy$ & GT & \textbf{72.}5 & \textbf{33.0}\\
    \hline
    7 & $\llmphy$ + noise & $\llmphy$ & 52.1 & 22.0\\
    8 & $\llmphy$ + noise & GT & 67.2 & 22.0\\
    \hline
    9 & $\llmphy$ (last-only) & $\llmphy$ & 51.1 & 11.0\\
    10 & $\llmphy$ (last-only) & GT & 70.5 & 33.0\\  
    \end{tabular}
    \caption{\AC{Performance comparison of $\llmphy$ against alternatives on various scene conditions and when using more number of objects on the simulated tray. In the experiments that show $\llmphy$+noise, we perturb the object trajectories with 25\% noise so that $\llmphy$ receives a noisy feedback. In the experiments $\llmphy$ (last-only), we feedback to $\llmphy$ only error and the physics parameters from the last iteration, without the full optimization trace. } }
    \label{tab:expts_ablations_appx}
\end{table*}

\begin{figure*}[ht]
    \centering
    \subfigure[Phase 1]{\label{fig:diff_llm_suppl}\includegraphics[width=4.5cm,trim={0.5cm 0cm 2cm 1.5cm},clip]{figs/mean_phase1_plot_with_gemma3_phi4.png}}
    \subfigure[Phase 1 (w/ variance)]{\label{fig:var_p1_appx}\includegraphics[width=4.5cm,trim={0.3cm 0cm 2cm 1.5cm},clip]{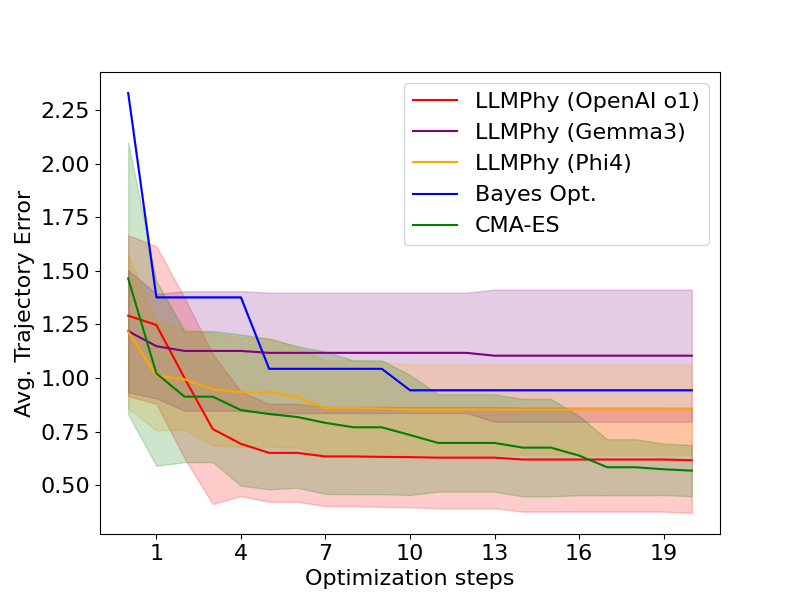}}
    \subfigure[Phase 2 (w/ variance)]{\label{fig:var_p2_appx_suppl}\includegraphics[width=4.5cm,trim={0.3cm 0cm 2cm 1.5cm},clip]{figs/variance_phase2_plot_with_gemma3.png}}
    \subfigure[]{\label{fig:iteration_hist}\includegraphics[width=6.5cm,trim={0.3cm 0cm 2cm 1.5cm},clip]{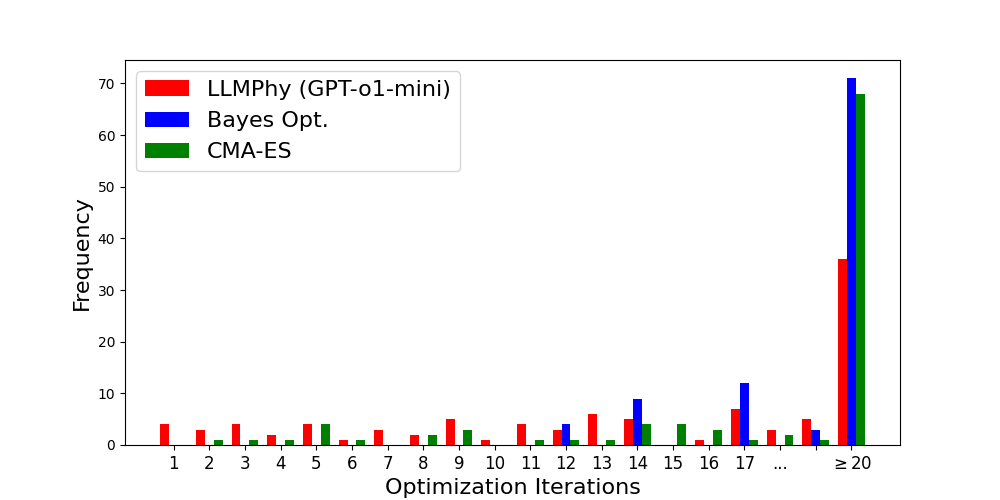}}
    \subfigure[]{\label{fig:llmphy_opt_trace}\includegraphics[width=6.5cm,trim={0.3cm 0cm 2cm 1.5cm},clip]{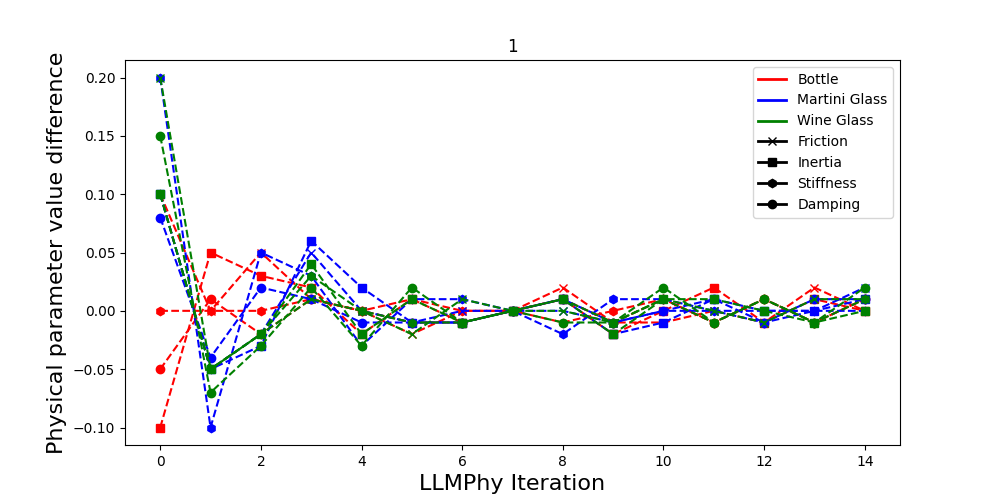}}
    \caption{\AC{(a) shows comparison of convergence when using various state-of-the-art LLMs in $\llmphy$ against Bayesian optimization and CMA-ES. We plot the minimum loss computed thus far in the optimization process against the number of optimization steps.  (b) plots show the convergence of $\llmphy$ and the error variance for Phase 1. (c) plots the convergence in Phase 2. We also compare the convergence using OpenAI o1 model as the LLM used in $\llmphy$. (d) Histogram of the best optimization iteration when using $\llmphy$ against other methods. (e) shows the differences between subsequent values for the various physical parameters in a typical iteration of $\llmphy$ from its value in the previous iteration. }}
    \label{fig:error-Phase1-appx}
\end{figure*}

\begin{figure*}[ht]
    \centering    
    \includegraphics[width=14cm,trim={0cm 3.1cm 0cm 7.5cm},clip]{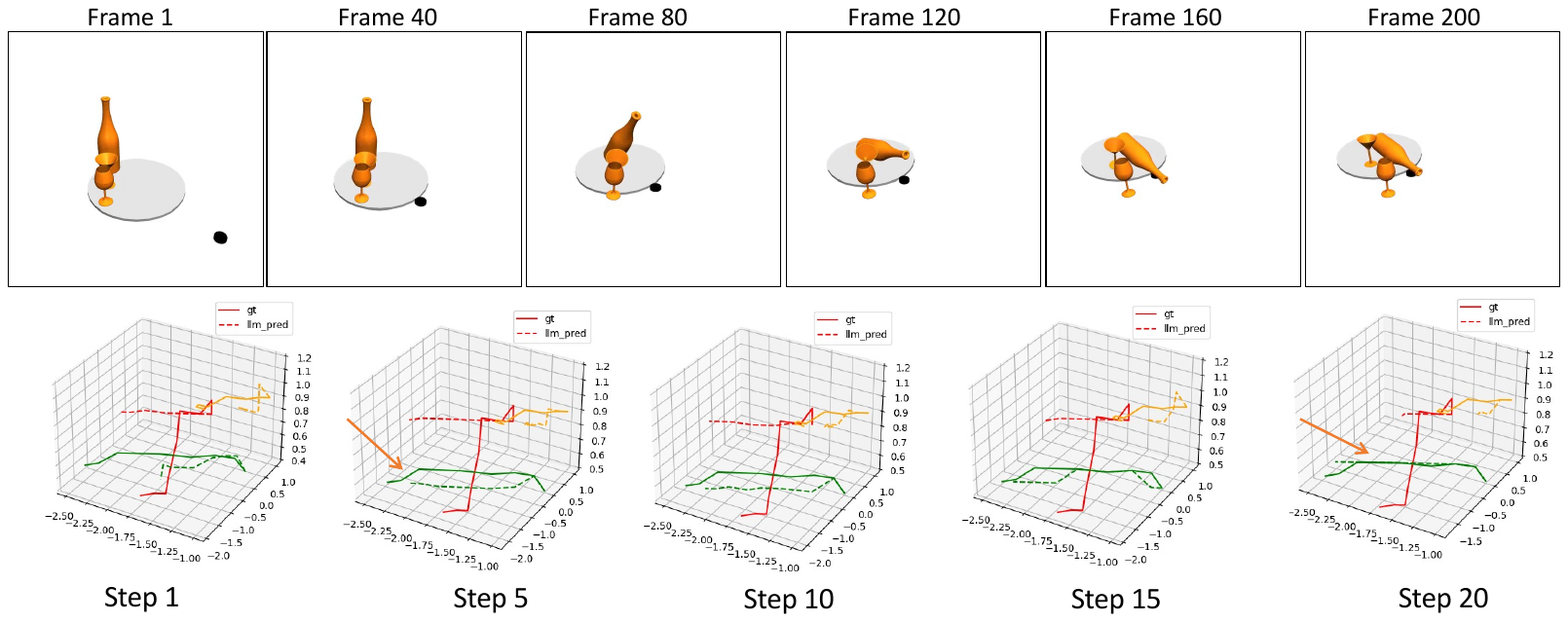}
    \caption{\AC{We show an example Phase 1 sequence (top). Below, we plot the motion trajectories for each of the objects in the frames and the predicted trajectories by $\llmphy$ from the optimization steps. The trajectory plots (below) show the ground truth trajectory (gt) and the predicted trajectory (llm\_pred), and as the iterations continue, we can see improvements in the alignment of the predicted and the ground truth object trajectories (as pointed out by the arrows). }}
    \label{fig:traj_alignment}
\end{figure*}

\begin{figure*}[ht]
    \centering    
    \includegraphics[width=14cm,trim={1cm 6.4cm 1cm 3.5cm},clip]{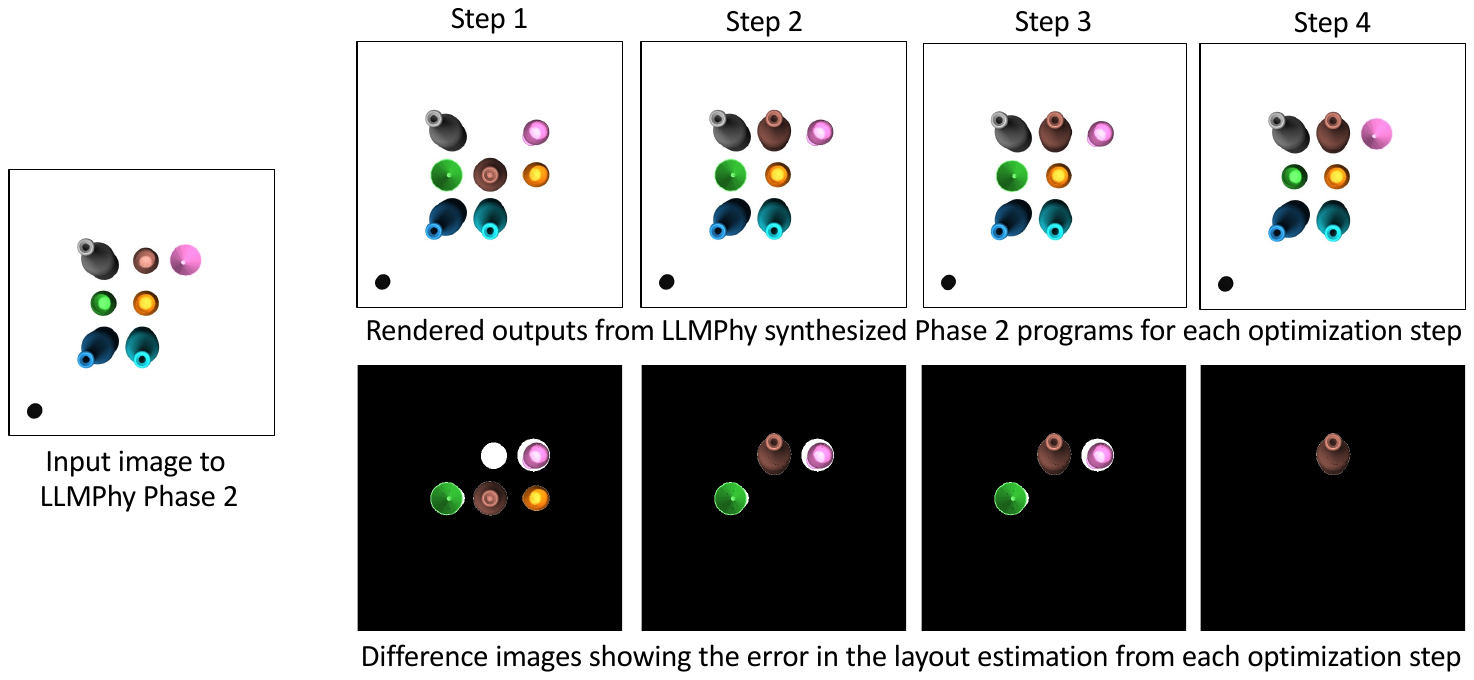}
    \caption{\AC{We show qualitative results from $\llmphy$ Phase 2 iterations. The input Phase 2 image is shown on the left. The top row shows the images produced by the simulator using the layout prediction code generated by $\llmphy$ for each Phase 2 optimization step. Below, we show the difference image between the predicted and the input Phase 2 images, clearly showing the errors. In Phase 2, the feedback to $\llmphy$ is produced using PSNR computed on the predicted and the ground truth images, as well as asking LLM (using the difference image) which of the objects are in error, and asking the LLM to fix the layout of these objects in the next iteration. As can be seen, the errors in the LLM layout prediction improves over iterations.}}
    \label{fig:Phase2_steps}
\end{figure*}

\section{LLMPhy CONVERGENCE ANALYSIS}
\label{sec:convergence_analysis}
\AC{In Figure~\ref{fig:diff_llm_suppl}, we plot the mean convergence (over a subset of the dataset) when using o1, GPT-4o, o1-mini, Bayesian Optimization, and CMA-ES. We see that the o1 model, that is explicitly trained for solving scientific reasoning, appears to be beneficial in our task. Interestingly, we see that o1's initial convergence is fast, however with longer iterations CMA-ES appears to outperform in minimizing the trajectory error. That being said, the plots in Figure 3 and Table 1 of the main paper points out that having lower trajectory error does not necessarily imply the physical parameters are estimated correctly (as they are implicitly found and are non-linear with regards to the trajectories), and having knowledge of physics in optimization leads to superior results. }

\AC{Further to this, in Figure~\ref{fig:iteration_hist}, we plot the histogram of best Phase 1 iterations between the various algorithms. Recall that the optimization methods we use are not based on gradients, instead are sampled discrete points, and the optimization approach is to select the next best sample towards minimizing the error. The plot shows that $\llmphy$ results in its best sample selections happen early on in its iterations than other methods.}

\AC{In Figure~\ref{fig:llmphy_opt_trace}, we plot the optimization parameter trace for one sample sequence, where we plot the differences between the values of the physics parameters produced by the LLM at an iteration against the values from the previous iteration. The plot shows the relative magnitude of changes the LLM makes to the parameters towards adjusting for the object trajectory error. We plot these adjustments for all the three objects and all the four parameters together in one plot so as to see the overall trend that the LLM makes. We also see that the LLM makes large adjustments in the first few iterations and that reduces in magnitude subsequently. For this particular example, $\llmphy$ converged in 15 iterations. }

\AC{In Figure~\ref{fig:var_p1_appx}, we plot the convergence of $\llmphy$-Phase 1, alongside plotting the variance in the trajectory error from the estimated physical parameters when used in the simulations. We found that a powerful LLM such as OpenAI o1-mini LLM or o1 demonstrates compelling convergence, with the lower bound of variance below that of other models. Our experiments suggest that better LLMs may lead to even stronger results.}

\AC{In Figure~\ref{fig:var_p2_appx_suppl}, we plot the convergence of $\llmphy$  Phase 2 iterations improving the PSNR between the synthesized (using the program) and the provide task images. As is clear, their correctness of the program improves over iterations. We would like to emphasize that BO and CMA-ES are continuous optimization methods and thus cannot optimize over the discrete space of Phase 2 layout. This is an important benefit of using $\llmphy$ for optimization that can operate on both continuous and discrete state spaces. }

\begin{figure*}[ht]
    \centering
    \includegraphics[width=\linewidth,trim={0.1cm 4.5cm 1.2cm 3.3cm},clip]{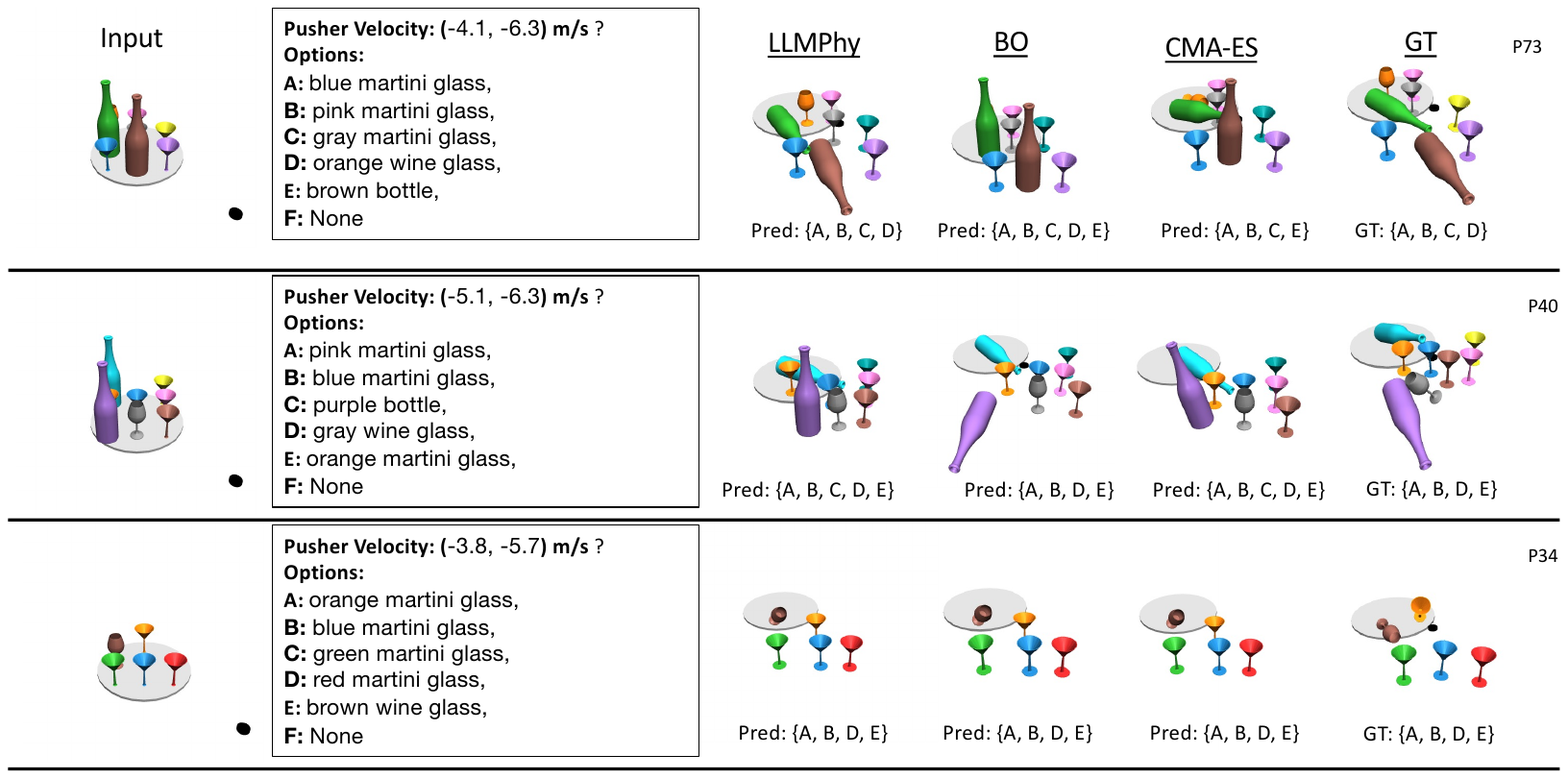}
    \includegraphics[width=\linewidth,trim={0.1cm 4cm 1.1cm 3.3cm},clip]{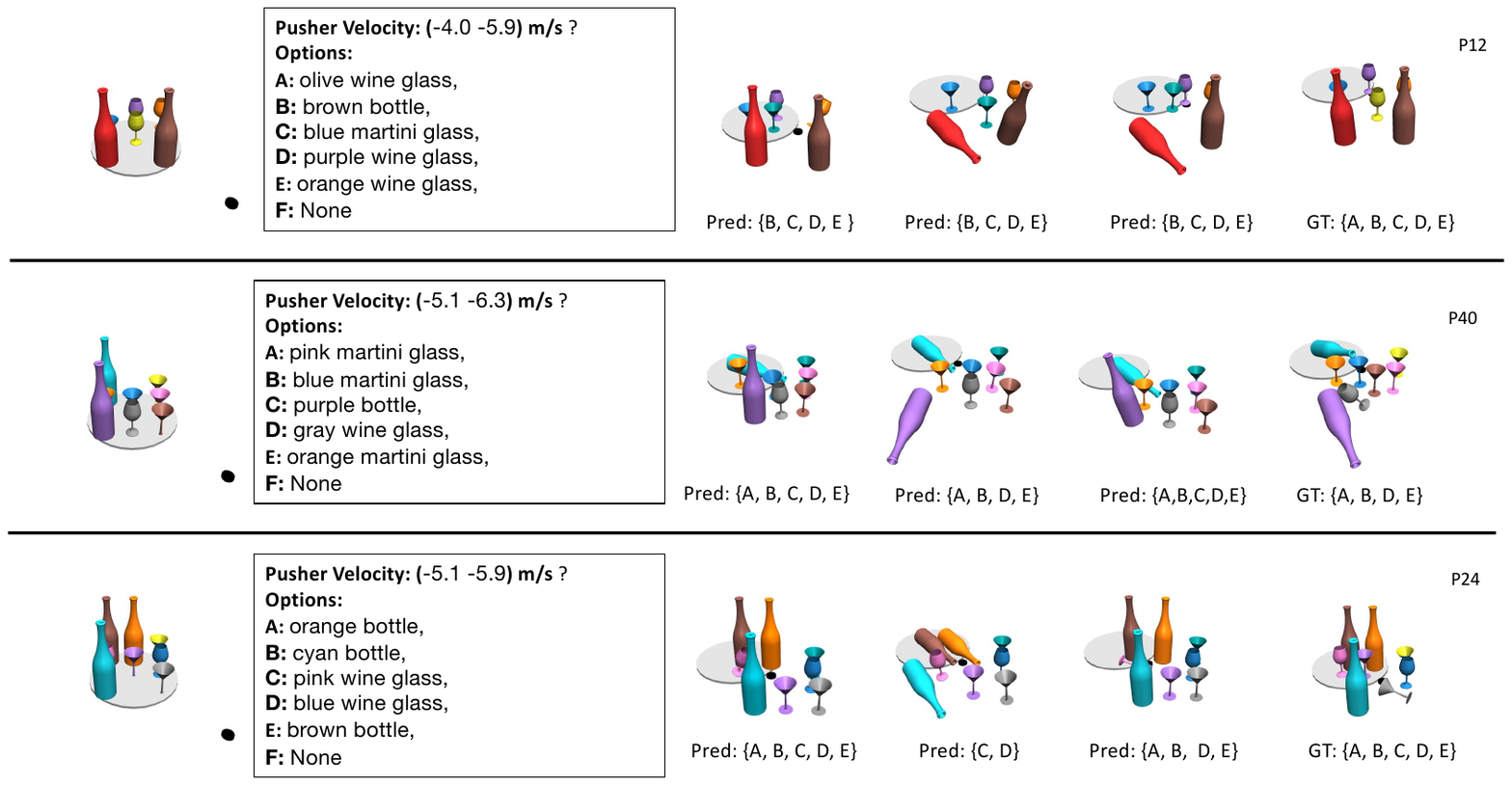}
    \caption{\AC{Qualitative comparisons between $\llmphy$, Bayesian optimization, and CMA-ES on the TraySim Dataset. $\llmphy$ uses o1-mini in Phase 1, but using the ground truth layout in Phase 2.}}
    \label{fig:qual_results_all}
\end{figure*}

\section{QUALITATIVE RESULTS}
\label{sec:qual_results_more}
\AC{In Figure~\ref{fig:qual_results_all}, we show several qualitative results from our TraySim dataset and comparisons of $\llmphy$ predictions to those of BO and CMA-ES. In general, we find that when the velocity of the pusher is lower, and the sliding friction is high, objects tend to stay stable if they are heavier (e.g., a bottle), albeit other physics parameters also playing into the outcome. In Figure~\ref{fig:traj_alignment}, we show example iterations from Phase 1 that explicitly show how the adjustment of the physical parameters by $\llmphy$ is causing the predicted object trajectories to align with the ground truth. In Figure~\ref{fig:Phase2_steps}, we show qualitative outputs from the optimization steps in Phase 2, demonstrating how the error feedback to the LLM corrects its previous mistakes to improve the layout estimation. }

\section{CLEVRER-LLMPhy EXPERIMENTS}
\label{sec:clevrer}
Recall that the TraySim dataset used the MuJoCo physics engine for the simulations and considered the estimation of four physics parameters, namely: sliding-friction, damping, inertia, and stiffness of the objects. While the object mass was considered, it was restricted to one of the three values in {10.0, 15.0, 20.0} for the three object categories we used. In order to validate the generalizability of $\llmphy$ to other settings, we consider a new dataset in the spirit of CLEVRER~\cite{yi2019clevrer}, however adapted to our TraySim setting and implemented with appropriate APIs to invoke a simulator using code generated by the LLM. 

Through CLEVRER-LLMPhy, we seek to explore three aspects of $\llmphy$: i) how does it work when used with a different physics engine? To this end, we used PyBullet~\cite{coumans2021pybullet} as the physics engine,  ii) consider other physics parameters, to which we consider {sliding-friction, restitution, rolling-friction, mass}, and iii) consider other geometries, motions, and collision physics, to which we use spheres and cubes for the objects. Recall that the \emph{coefficient of restitution} captures the amount of energy that is transferred when bodies collide, while the \emph{coefficient of rolling-friction} plays a key role in how much an object (e.g., a cube) will topple or how much is the resistance for a sphere to roll. Further, unlike TraySim, we consider \emph{mass} to be estimated in CLEVRER-LLMPhy, and can any value between [0.1, 20].

\subsection{CLEVRER-LLMPhy}
As described in the main paper, the CLEVRER-LLMPhy task follows the same setup as in TraySim with two phases: Phase 1 for physics parameter estimation, and Phase 2 for collision reasoning. 

\begin{figure*}
    \centering
    \subfigure[Ex:1-step-1]{\includegraphics[width=0.24\linewidth]{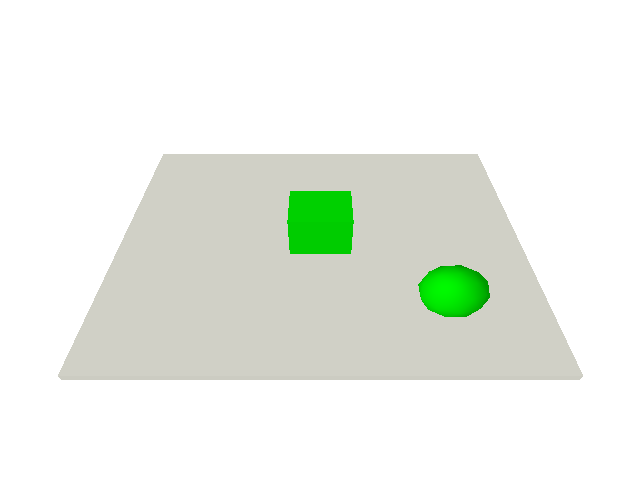}}
    \subfigure[Ex:1-step-100]{\includegraphics[width=0.24\linewidth]{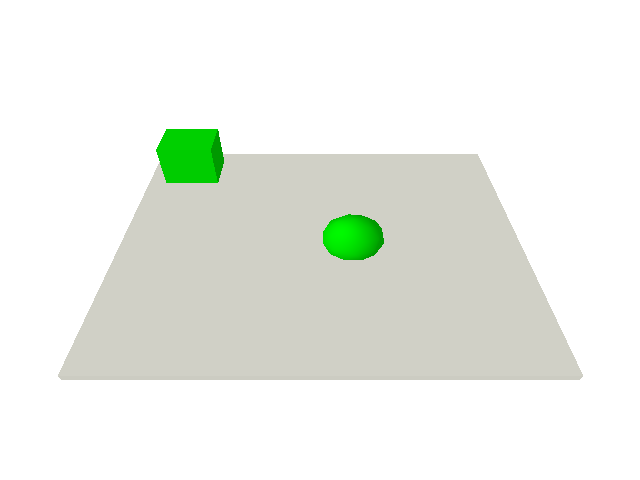}}
    \subfigure[Ex:2-step-100]{\includegraphics[width=0.24\linewidth]{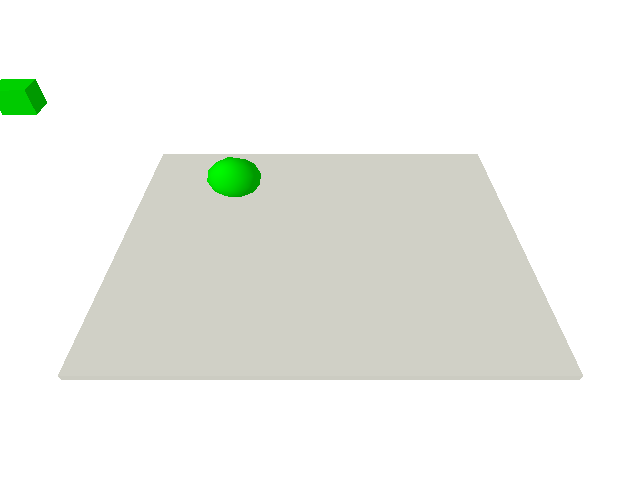}}
    \subfigure[Ex:3-step-100]{\includegraphics[width=0.24\linewidth]{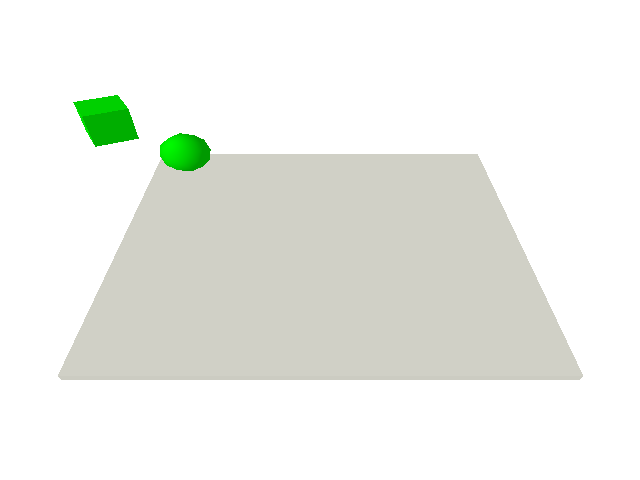}}
    \caption{Left most: First frame in Phase 1 of CLEVRER-LLMPhy. Subsequent frames show the final states from three examples in the 100-th step for varied physics parameters. }
    \label{fig:clevrer-phase1}
\end{figure*}

\noindent\textbf{Phase 1:} In this phase, the task of $\llmphy$ is to estimate the physics parameters of objects in the scene using a setup that has a single sphere and a cube, with the sphere moving towards the cube, and colliding with it. The initial velocity of the sphere is assumed given. The physics parameters for the examples in the dataset were randomly chosen. The trajectories of center of gravities of motion of the objects are extracted from the simulator. The goal of $\llmphy$ is to minimize the trajectory error for the sphere and the cube as was described for TraySim using the LLM to produce estimates of the physics in a potentially intuitive but trial-and-error black-box manner. See Figure~\ref{fig:clevrer-phase1} for some examples of Phase 1 setup showing the first and last frames.

\noindent\textbf{Phase 2:} In this phase, the goal of $\llmphy$ is to use the estimated physics parameters to conduct higher-level complex reasoning using the simulator. The input to $\llmphy$ is a top-down and side-view image of the initial setup of the scene and consist of multiple instances of objects arranged in a grid, with the goal to estimate the layout using program synthesis. Specifically, $\llmphy$ is prompted to produce the type, color, and location of each object on the grid from the two images. The objects and their positions on the grid were randomly chosen, however under the condition that the location (1,1) on the grid must be a sphere, as it is expected to initiate the task (similar to the pusher in TraySim).

\begin{figure*}
    \centering
    \subfigure[Ex:1-step-1]{\includegraphics[width=0.24\linewidth]{figs/suppl/ex1_p2_rgb_front_1.png}}
    \subfigure[Ex:1-step-100]{\includegraphics[width=0.24\linewidth]{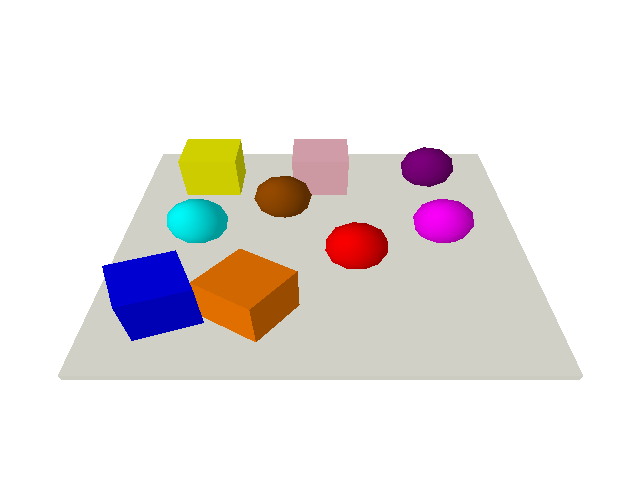}}
    \subfigure[Ex:2-step-1]{\includegraphics[width=0.24\linewidth]{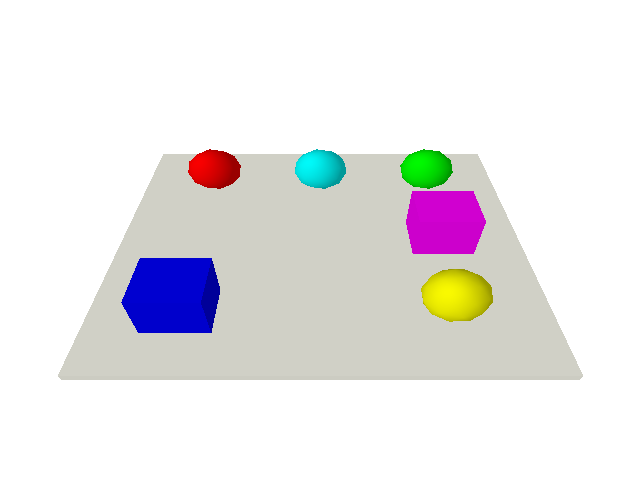}}
    \subfigure[Ex:2-step-100]{\includegraphics[width=0.24\linewidth]{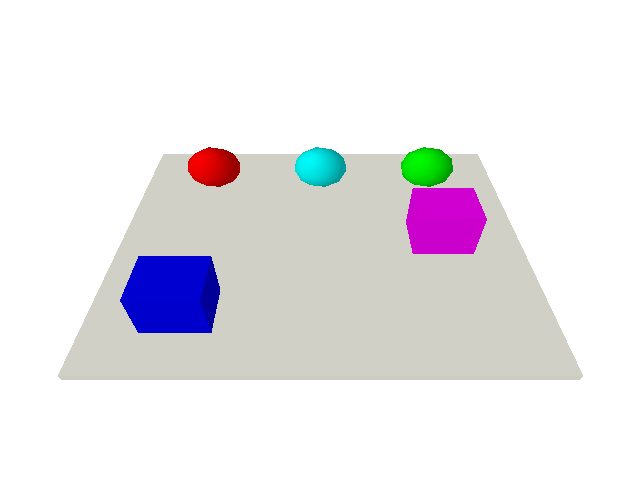}}
    \caption{Left two: First and last frame in Phase 2 of CLEVRER-LLMPhy. Right two: the same for another sequence. In the first pair, the red sphere is the initiating collider while in the second pair, it is the yellow sphere. Note that }
    \label{fig:clevrer-phase2}
\end{figure*}

\noindent\textbf{Task:} In the CLEVRER-LLMPhy, a sphere located at the grid location (1, 1) in the Phase 2 initial scene is assumed to move with a given velocity, and the goal is to reason which of the objects in the scene will involve in a collision either to the initiating sphere or the subsequent mutual collisions ensued. This problem is challenging as the motion trajectory of the initiating sphere and other objects are controlled by their varied physics parameters. The performance of a model is computed in a multiple choice format, with a given set of 5 answers for objects in collisions and their colors, and a 6th option for 'None'. The final performance on a CLEVRER-LLMPhy example is computed using the mean intersection-over-union metric that computes the intersection over the predicted options against their union by $\llmphy$ against the ground truth answer set. We also provide a 'Precise' accuracy, which calculates the frequency at which the answers predicted by a model matches precisely with the ground truth answer.  See Figure~\ref{fig:clevrer-phase2} for two examples from Phase 2 demonstrating the task. A reasoning model only sees the first frame (left ones) and the Phase 1 sequence, and is expected to reason using the simulator to produce the last frame in Phase 2 from which the objects undergoing collisions can be extracted.  

\noindent\textbf{Dataset Statistics:} Using the above setup, we created a dataset with 10 sequences, the numbers for which was reported in the main paper. Below we provide results using an additional 15 sequences. Each sequence used 100 frames captured at 240 fps. The initial velocity of Phase 1 sphere is fixed at (5,-5) m/s while for Phase 2, the velocity is randomly selected from [-10, 10] m/s for each 2D coordinate.


\subsection{Experimental Results}
\begin{figure*}[htbp]
    \centering
    \subfigure[Convergence Phase 1]{\includegraphics[width=0.32\linewidth]{figs/clevrer_mean_phase1_plot.png}}
    \subfigure[Conv. Phase 1 (w/ variance)]{\includegraphics[width=0.32\linewidth]{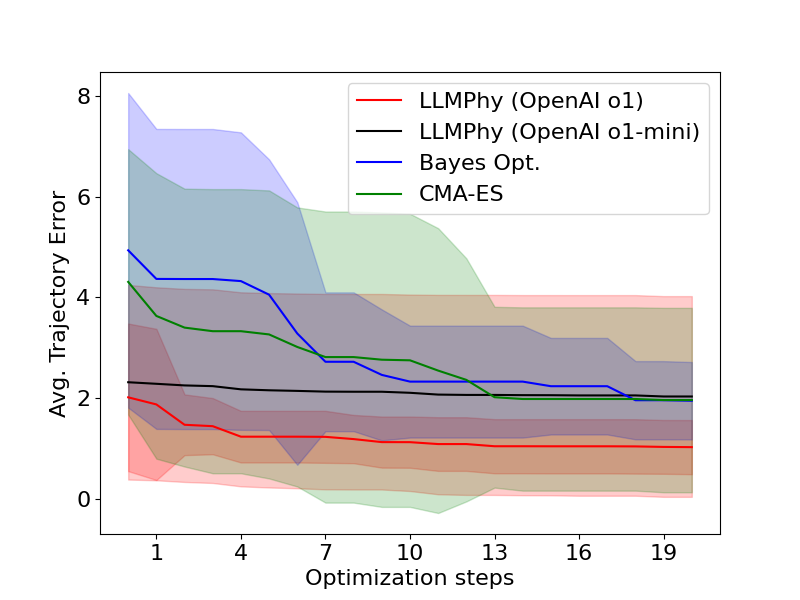}}
    \subfigure[Conv. Phase 2 (w/ variance)]{\includegraphics[width=0.32\linewidth]{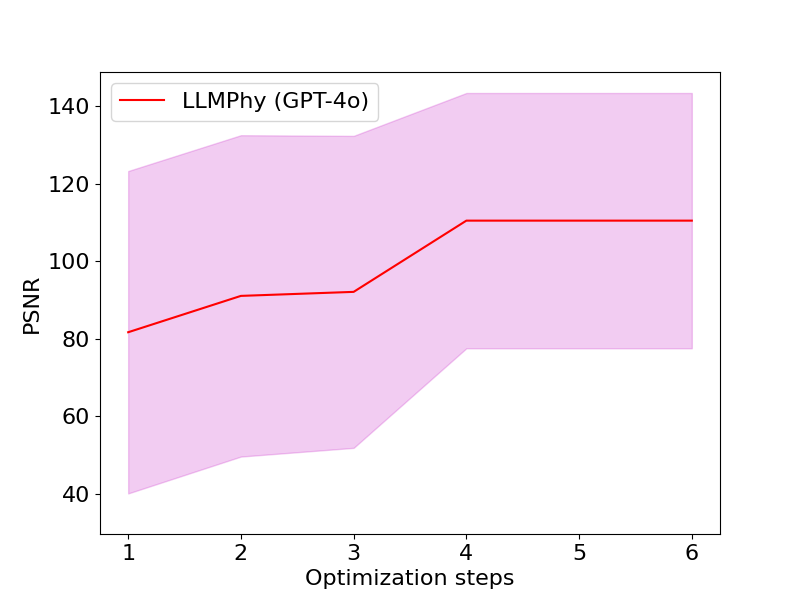}}
    \caption{(a) shows comparison of convergence when using various state-of-the-art LLMs in $\llmphy$ against Bayesian optimization and CMA-ES, including the OpenAI o1 model. We plot the minimum loss computed thus far in the optimization process against the number of optimization steps.  (b) plots show the convergence of $\llmphy$ and the error variance for Phase 1. (c) plots the convergence in Phase 2.}
    \label{fig:clevrer-convergence}
\end{figure*}

\noindent\textbf{Convergence Plots}
In Figures~\ref{fig:clevrer-convergence}, we plot the convergence of $\llmphy$ optimization for phases 1 and 2 on the CLEVRER-LLMPhy dataset. We compare against o1 and o1-mini, the latter is used in our main experiments in the paper. As is clear, using more powerful LLMs such as o1 demonstrates significantly better convergence than o1-mini, and even better than using BO or CMA-ES. This was also seen in our experiments on TraySim as in Figure 3 of the main paper. In Figure~\ref{fig:clevrer-convergence}(c), we plot the convergence in Phase 2 on the CLEVRER-LLMPhy dataset. As was seen on the TraySim, we see that the iterative methodology improves the layout estimation accuracy.

\begin{figure*}[htbp]
    \centering
    \begin{tabular}{>{\centering\arraybackslash}m{0.22\textwidth}
                  >{\centering\arraybackslash}m{0.22\textwidth}
                  >{\centering\arraybackslash}m{0.22\textwidth}
                  >{\centering\arraybackslash}m{0.22\textwidth}}

        \textbf{GT \#1} & \textbf{GT \#100} & \textbf{CMA-ES \#100} & \textbf{LLMPhy \#100} \\
        \includegraphics[width=\linewidth]{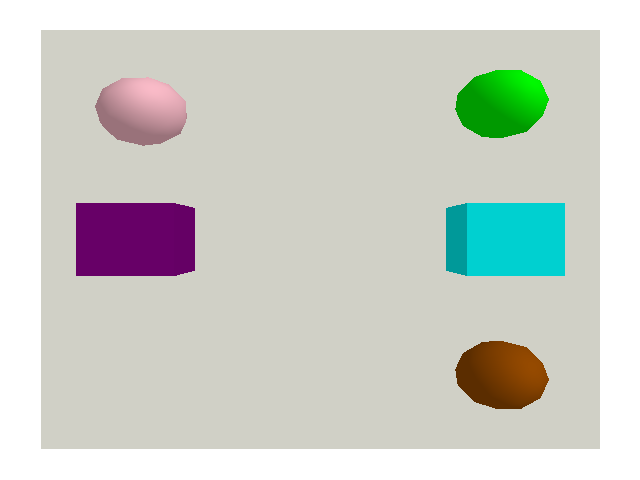} & 
        \includegraphics[width=\linewidth]{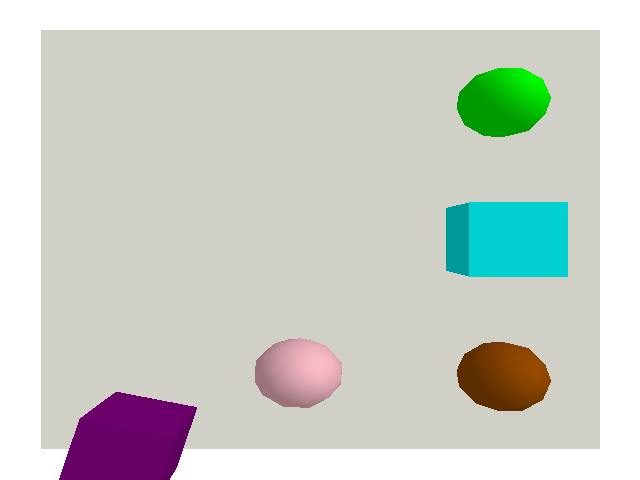} &
        \includegraphics[width=\linewidth]{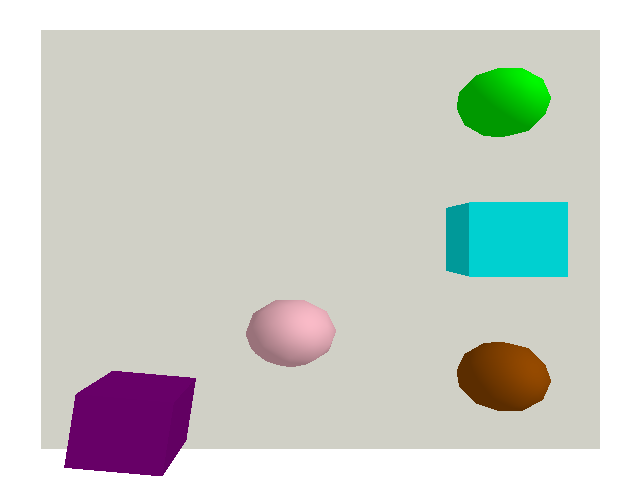} &
        \includegraphics[width=\linewidth]{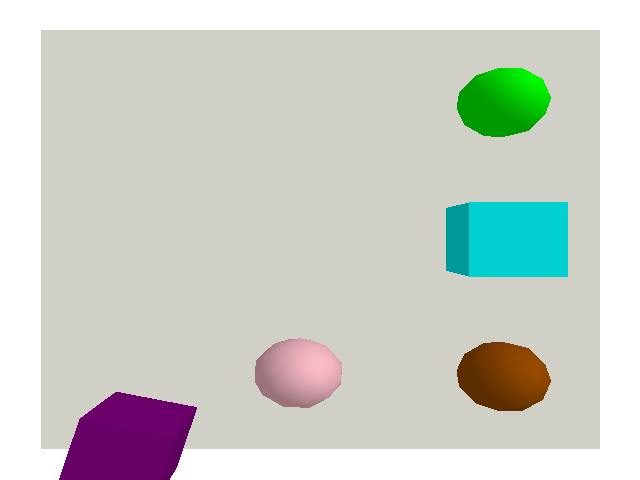}\\[0.5em]
        
        \includegraphics[width=\linewidth]{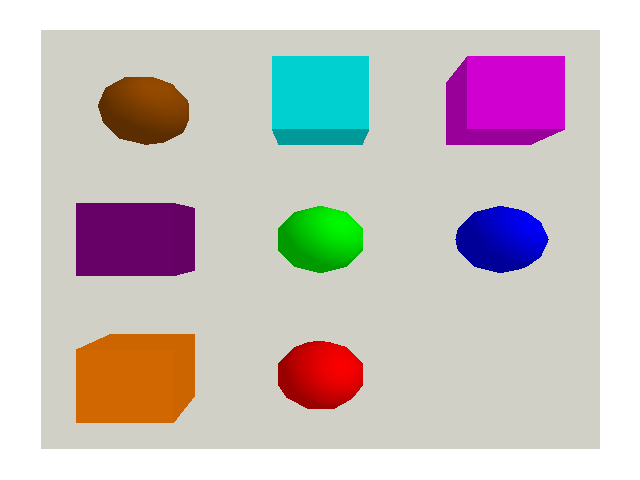} & 
        \includegraphics[width=\linewidth]{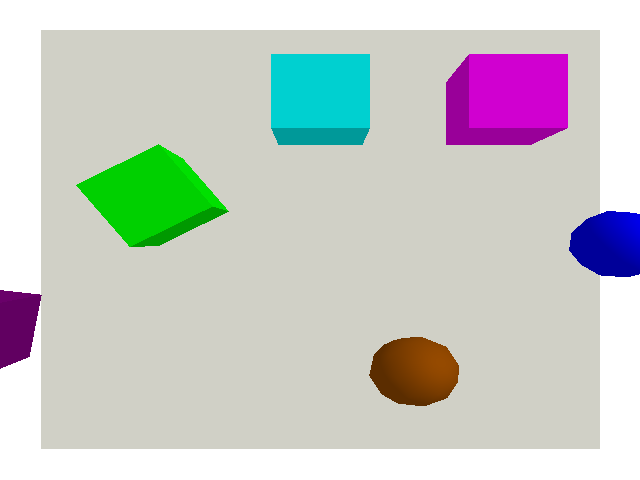} &
        \includegraphics[width=\linewidth]{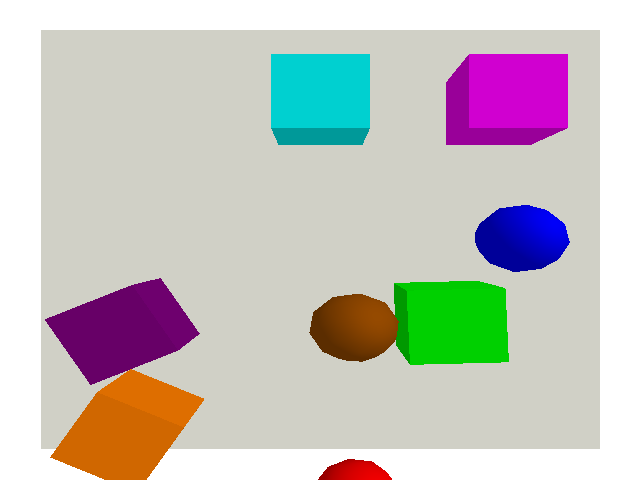} &
        \includegraphics[width=\linewidth]{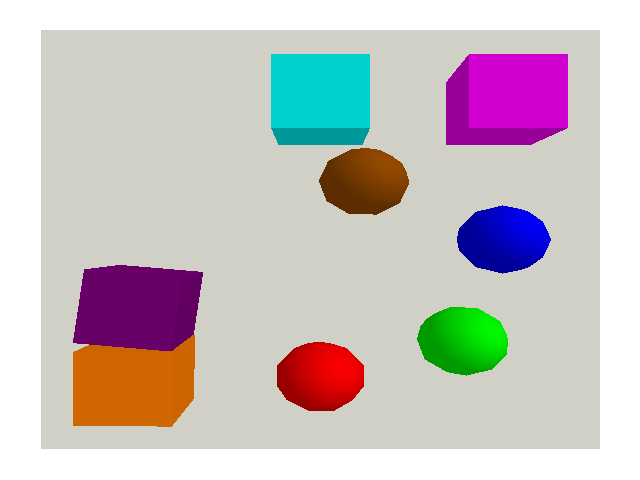}\\[0.5em]
        
        \includegraphics[width=\linewidth]{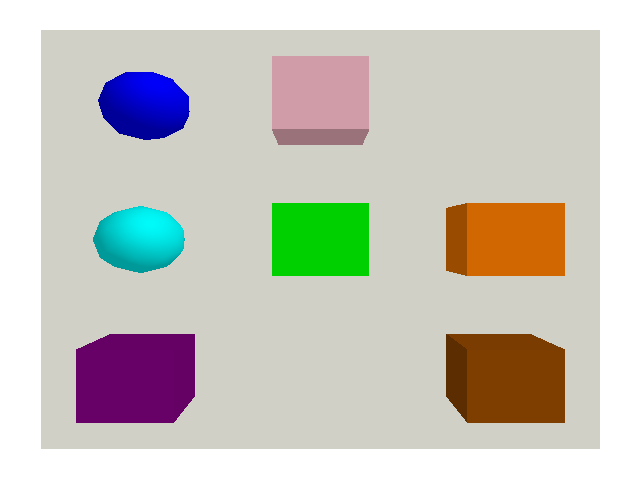} &
        \includegraphics[width=\linewidth]{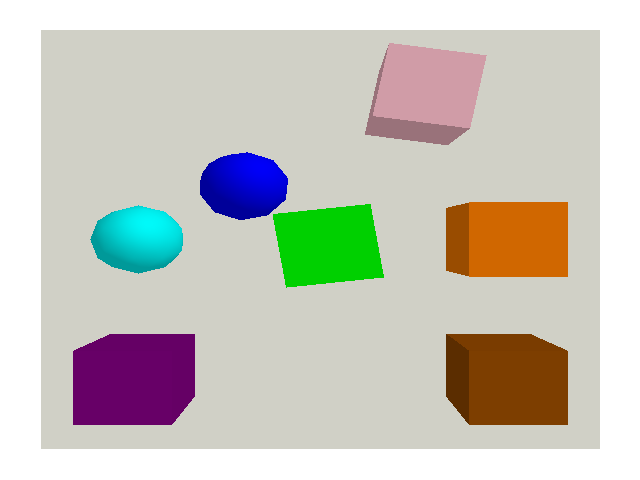} &
        \includegraphics[width=\linewidth]{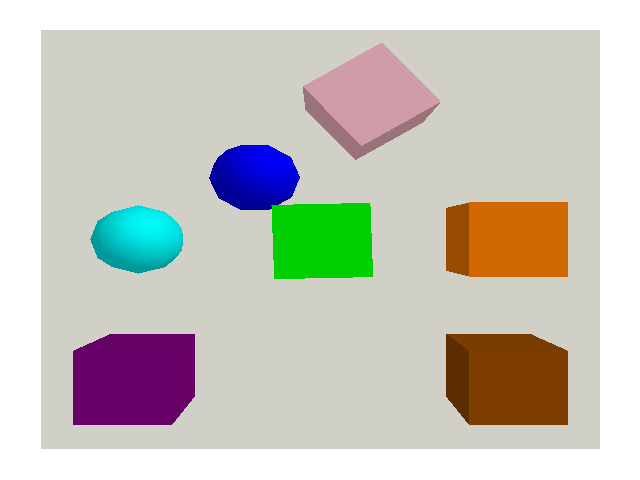} &
        \includegraphics[width=\linewidth]{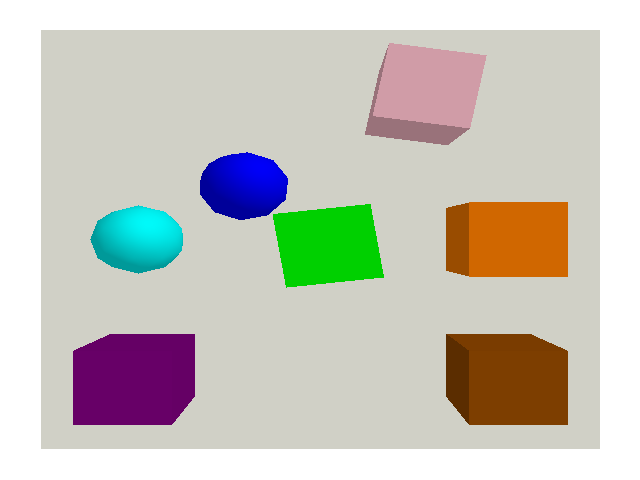}\\[0.5em]
        
        \includegraphics[width=\linewidth]{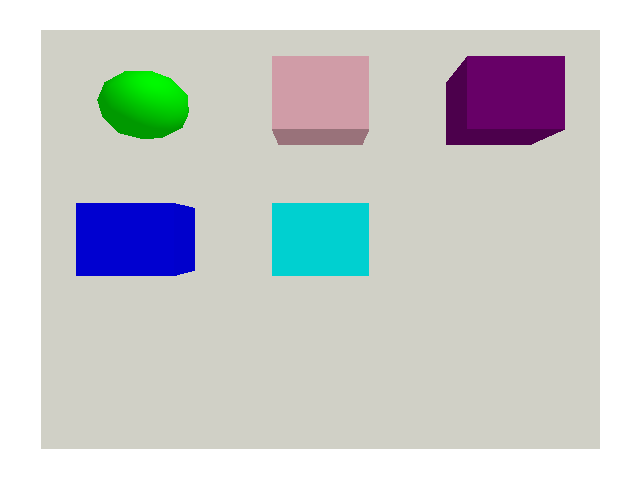}&
        \includegraphics[width=\linewidth]{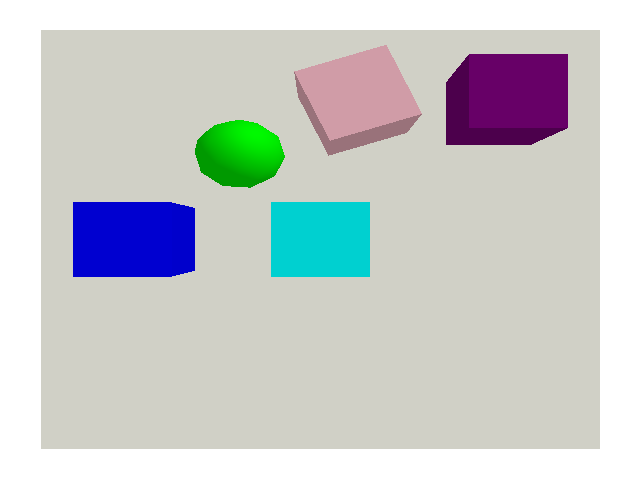}&
        \includegraphics[width=\linewidth]{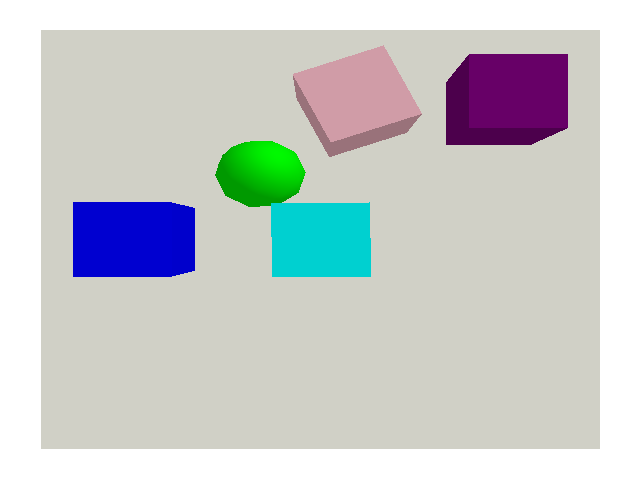}&
        \includegraphics[width=\linewidth]{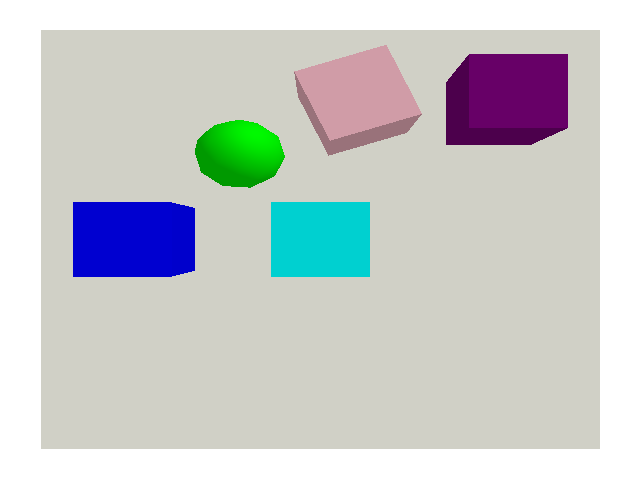}\\[0.5em]

        \includegraphics[width=\linewidth]{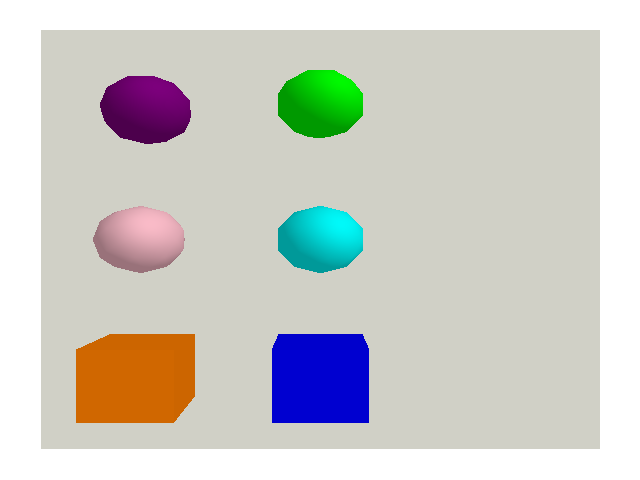}&
        \includegraphics[width=\linewidth]{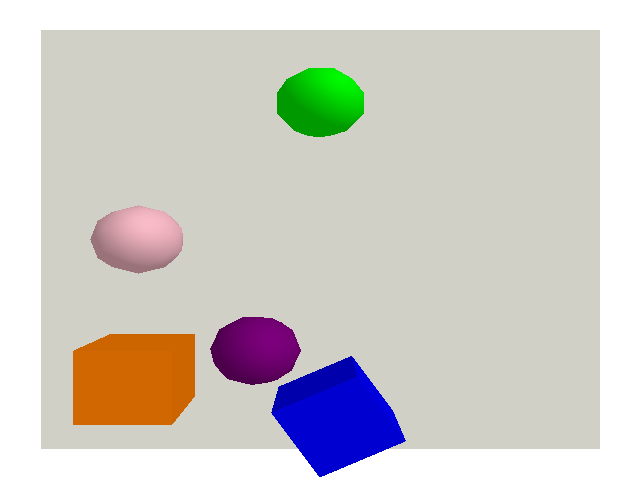}&
        \includegraphics[width=\linewidth]{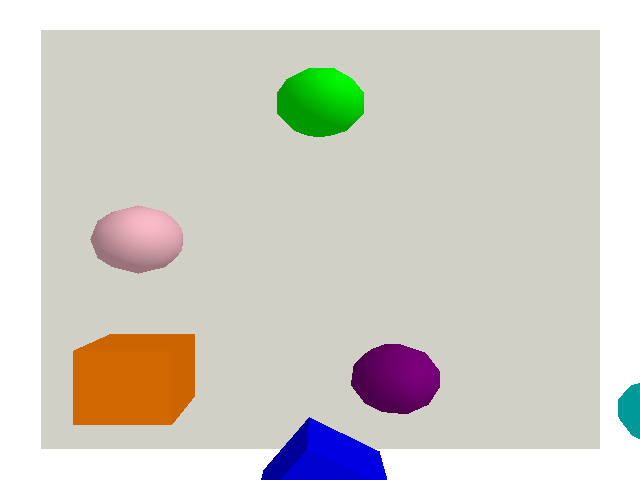}&
        \includegraphics[width=\linewidth]{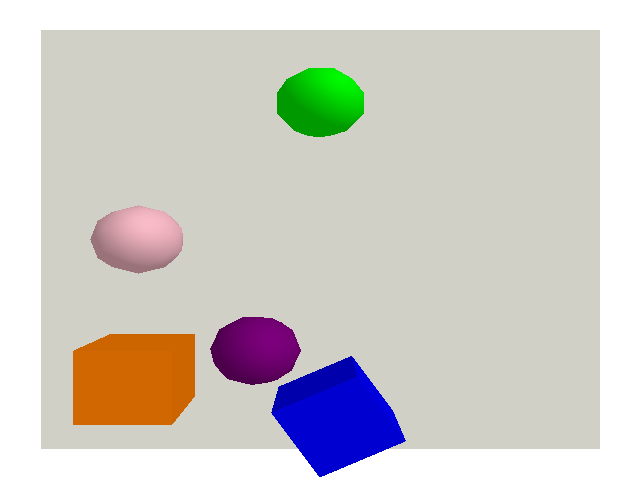}
    \end{tabular}
    \caption{Qualitative comparisons between LLMPhy and other methods on the CLEVRER-LLMPhy task. In all the sequences, the sphere on the top-left is the collision initiating object. We show the ground truth frames 1 and 100, and the 100-th (final) frames from CMA-ES and $\llmphy$. Only the first frame is input to Phase 2 of $\llmphy$ and the physics is learned from the auxiliary sequences, one for each task example. As is clear from the second and last columns, $\llmphy$ learns the physics better than CMA-ES as depicted by the collision positions of the various objects. Note that LLMhy uses GPT models in both phases while CMA-ES uses GT for Phase 2. We show top-down view of the scene for better clarity of the final states. Note that both top-down and side-views are given to $\llmphy$ in Phase 2.}
    \label{fig:qual-results-clevrer}
\end{figure*}

\noindent\textbf{Qualitative Results}
In Figure~\ref{fig:qual-results-clevrer}, we provide several qualitative results of $\llmphy$ on the CLEVRER-LLMPhy dataset against CMA-ES.

\section{REAL WORLD EXPERIMENTS}
\label{sec:realworld}
The goal of this experiment is to demonstrate the performance of $\llmphy$ in real world settings, specifically to explore the feasibility of estimating the underlying physics of the scene from RGB camera frames. To this end, we programmed a Kinova Gen3 collaborative robot with 7DOF and a wrist camera. The robot used a 
Robotiq 2F-85 gripper (used as a rigid object pusher with fingers fully closed). We use the ROS Noetic robotic system for controlling the robot arm using velocity control in 6D. The computer vision setup used one RealSense D515 camera publishing depth and color images directly over the ROS network, however we used only the RGB images without depth in our experiments to make our setup similar to the one in TraySim. The setup had a tray being pushed by the gripper at speeds in [0.1, 0.4] m/s along a straight path. The camera is calibrated for its intrinsic and extrinsic parameters. As in the TraySim setup, our goal is to estimate the physics of the scene from the motions of objects on a tray when pushed by the pusher. To this end, we used two boxes as the objects, namely: an empty Granola cereal box and a RealSense camera box, the former is lighter and more dynamic than the latter.

From the ROS network we capture raw images and use computer vision techniques for segmenting, tracking, and computing the 3D trajectories of the objects. Specifically, we use MiDAS~\cite{Ranftl2022} for monocular depth estimation and SAM~\cite{kirillov2023segment} to find the tray and object segments. We use April Tags~\cite{wang2016apriltag} on the objects for their pose estimation and tracking, and use classical object trackers~\cite{shi1994good} when the April Tags are occluded or undergo motion blur. 

\begin{figure*}
    \centering
    \subfigure[Real Scene]{\includegraphics[width=0.24\linewidth]{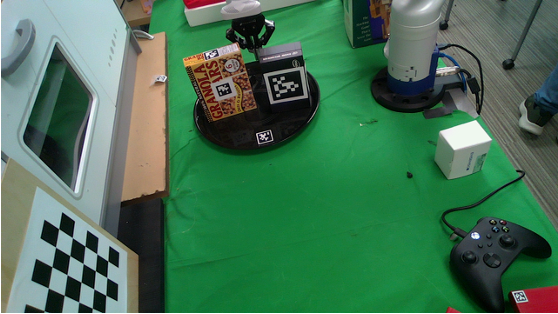}}
    \subfigure[Monocular Depth]{\includegraphics[width=0.24\linewidth]{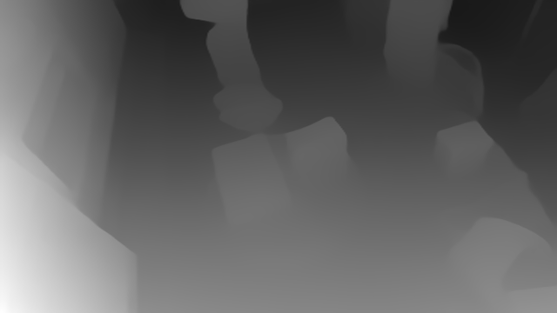}}
    \subfigure[SAM Segments]{\includegraphics[width=0.24\linewidth]{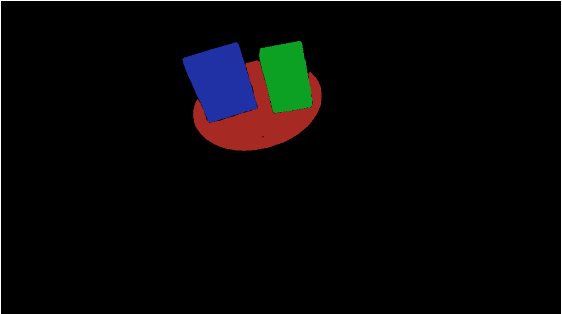}}
    \subfigure[MuJoCo Digital Twin]{\includegraphics[width=0.24\linewidth]{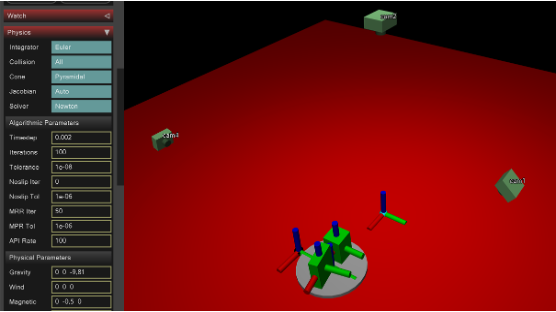}}
    \caption{Given an RGB video of real world physical reasoning scene, we use the April Tags, per-frame monocular depth estimation and SAM segmentation to reconstruct each frame independently in a MujoCo simulator. }
    \label{fig:real-scene-vision}

    \centering
    \includegraphics[width=0.3\linewidth,trim={10cm 4cm 6cm 0cm},clip]{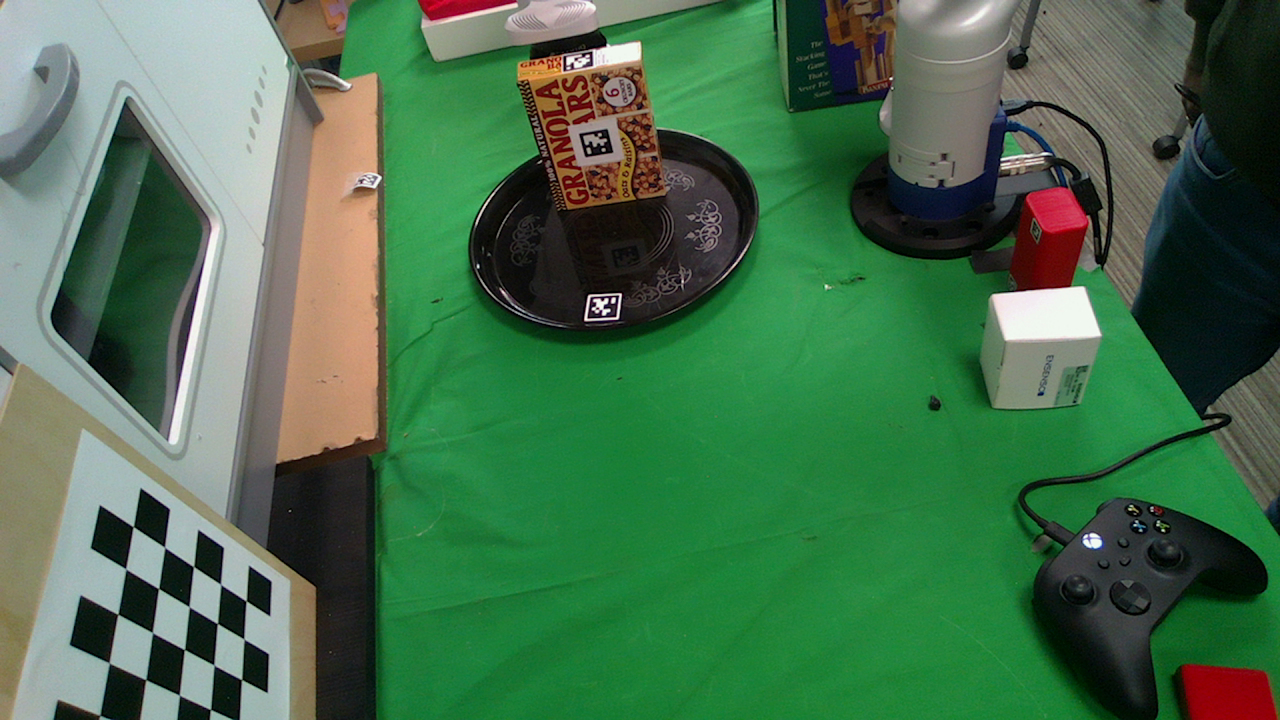}
    \includegraphics[width=0.3\linewidth,trim={10cm 4cm 6cm 0cm},clip]{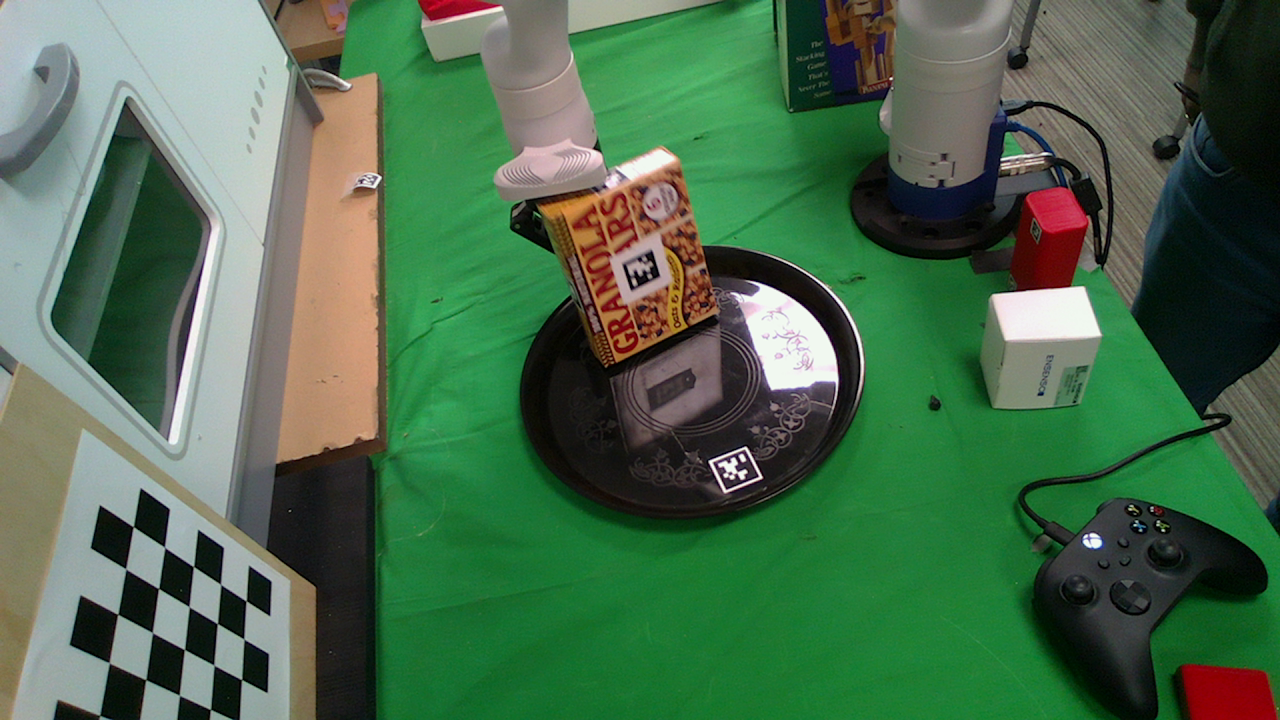}
    \includegraphics[width=0.3\linewidth,trim={10cm 4cm 6cm 0cm},clip]{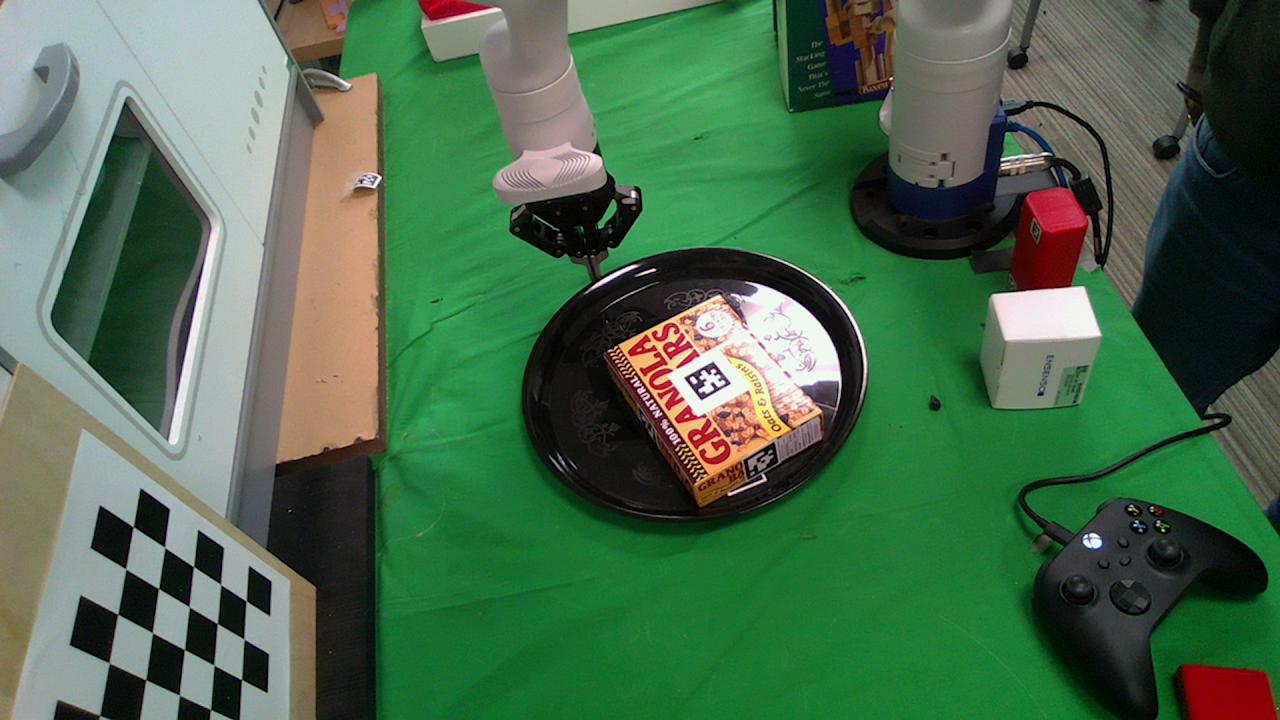}\\
    \includegraphics[width=0.3\linewidth]{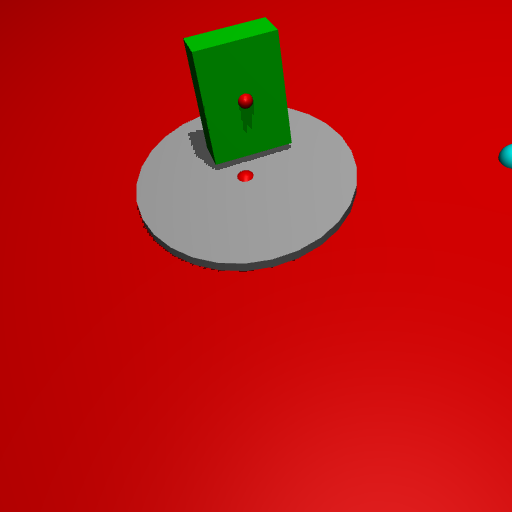}
    \includegraphics[width=0.3\linewidth]{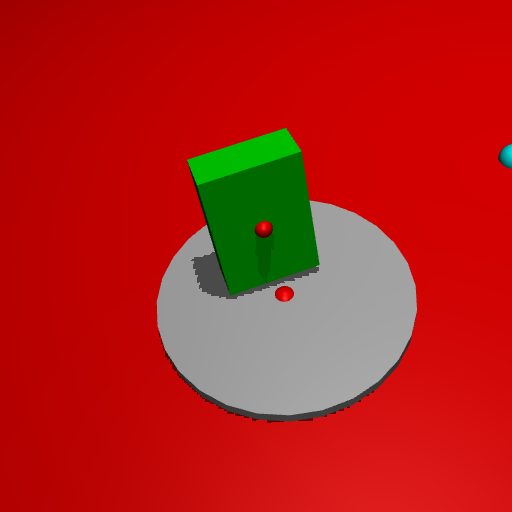}
    \includegraphics[width=0.3\linewidth]{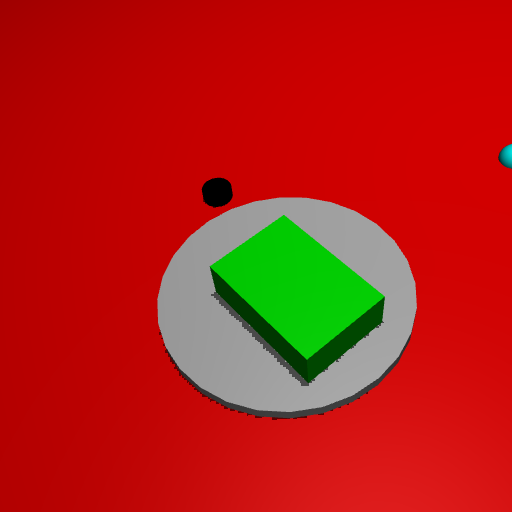}
    \\[1em]
     \includegraphics[width=0.3\linewidth,trim={10cm 4cm 6cm 0cm},clip]{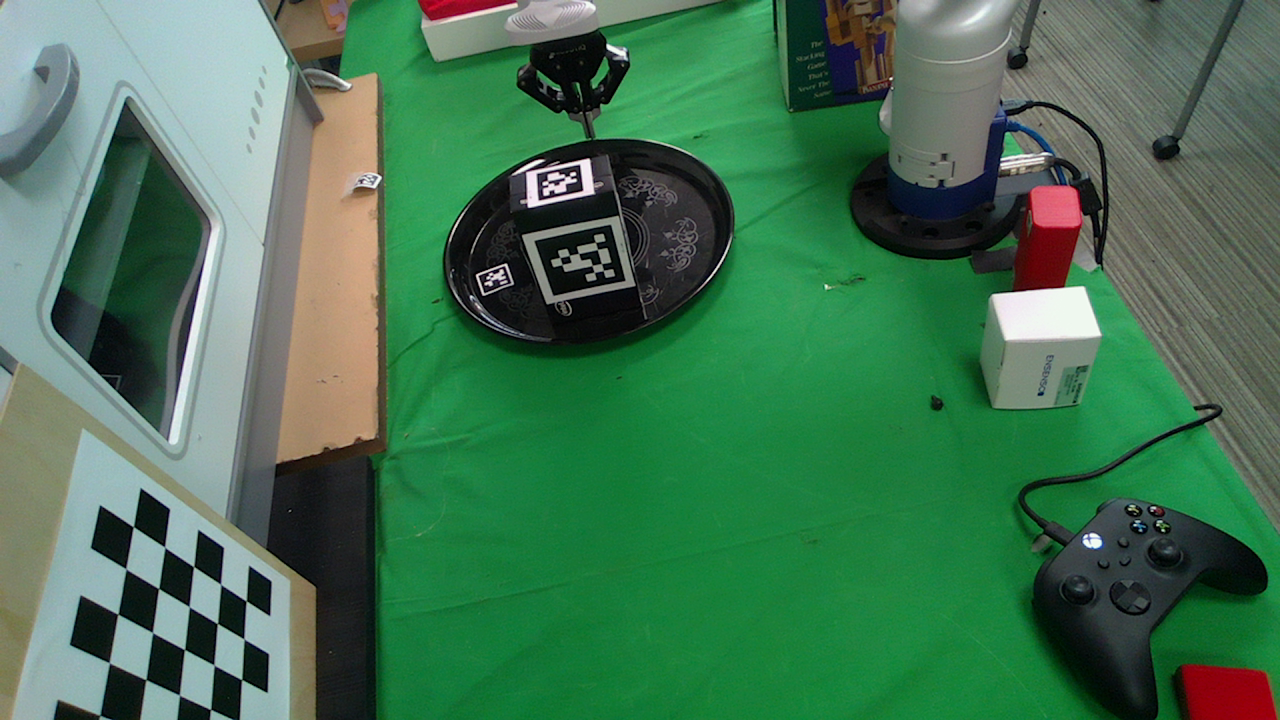}
    \includegraphics[width=0.3\linewidth,trim={10cm 4cm 6cm 0cm},clip]{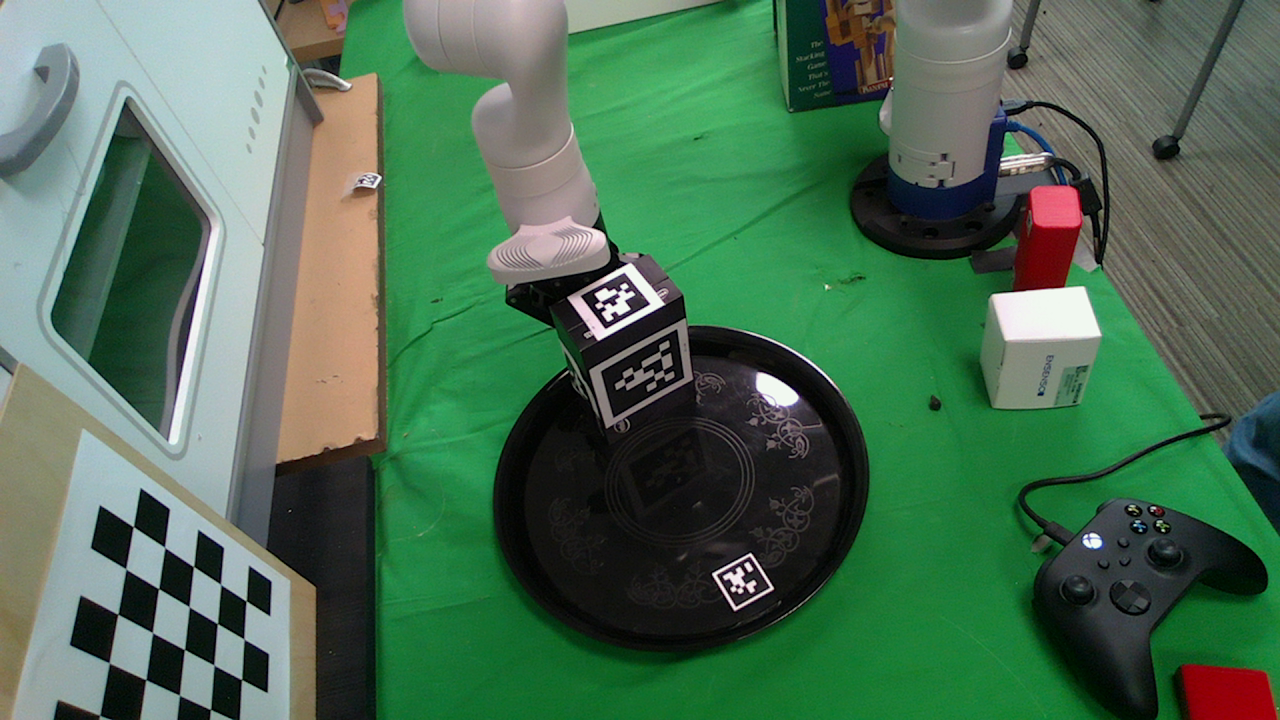}
    \includegraphics[width=0.3\linewidth,trim={10cm 4cm 6cm 0cm},clip]{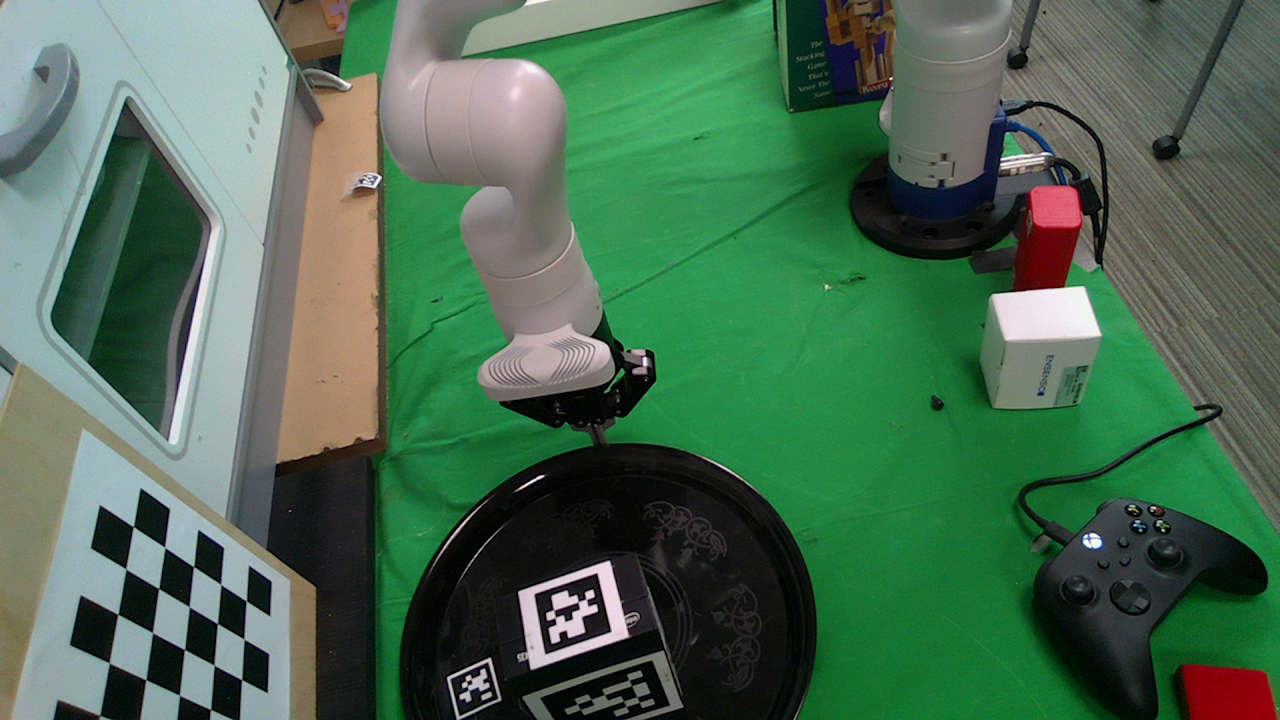}\\
    \includegraphics[width=0.3\linewidth]{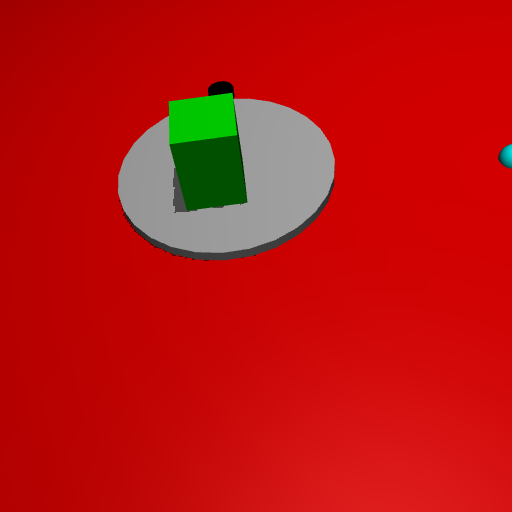}
    \includegraphics[width=0.3\linewidth]{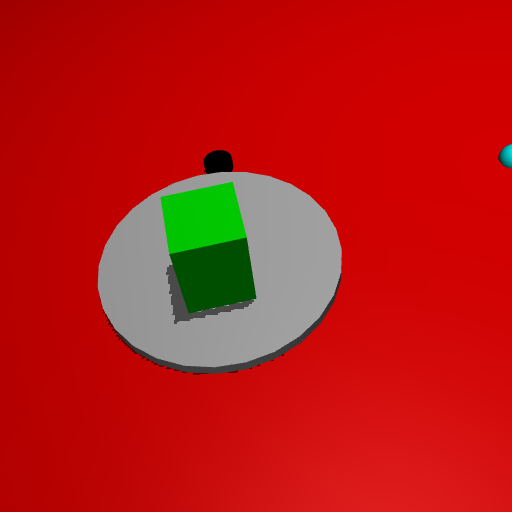}
    \includegraphics[width=0.3\linewidth]{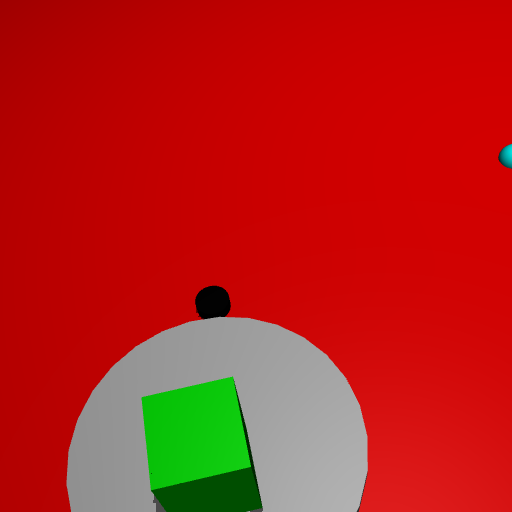}
    \caption{A digital twins of scenes in MuJoCo simulator. The blue sphere shows the world origin and we show the robotic gripper using the black pusher in the twin. The top scene uses a lighter cereal box and the below scene uses a heavier RealSense box. The top scene used a gripper speed of 0.3 m/s and below scene uses a speed of 0.4 m/s. Note that there are errors in the exact reconstruction of the scene in the digital twin, due to the various sources of noise, including depth estimation.}
    \label{fig:digital-twin-examples}
\end{figure*}

In Figure~\ref{fig:real-scene-vision}, we show the outputs from the computer vision modules for a scene. The segmentation module used LangSam~\cite{kirillov2023segment,luca2023langsegmentanything} for finding the segments corresponding to each of the boxes and the tray from text inputs of their names. The centroids of these segments are considered for depth estimation using the MiDAS. The computed depth is then transformed into 3D world coordinates using the camera calibration. The 3D pose of the boxes are extracted from the April Tags. The 3D object trajectories and their pose trajectories are then used to reconstruct the scene for every frame in the MuJoCo simulator to produce a digital twin of the real world scene. Specifically, the digital twin ensures that our setup is physically realizable within the simulator and  allows designing the API software interfaces to be invoked during $\llmphy$ execution. Note that, this twin of the real scene in the simulator is created for each frame separately, without involving any physical parameters, i.e., we reconstruct the real scene one frame at a time independent of its past or future object trajectories. Next, the 3D trajectories and pose sequences of the objects from the digital twin are used in $\llmphy$ with the goal of $\llmphy$ being to estimate the physics of the scene that will reproduce the frame-by-frame object motions. Particularly, $\llmphy$ must estimate the physical parameters of the scene such that the program it generates incorporating these parameters, when executed in the simulator must reproduce a simulated video sequence as close as possible to the independent frame-level reconstructions used in digital twin of the real world. Note that we only use Phase 1 of $\llmphy$ in this setup as Phase 2 (of inferring the locations of objects) is directly handled using our vision setup. 

\subsection{Experimental Results:}
\begin{figure*}
    \centering
    \begin{minipage}[c]{0.05\textwidth}
    \centering
    \rotatebox{90}{\small Original}
    \end{minipage}%
    \begin{minipage}[c]{0.95\textwidth}
        \centering
        \includegraphics[width=0.24\linewidth,trim={10cm 4cm 6cm 0cm},clip]{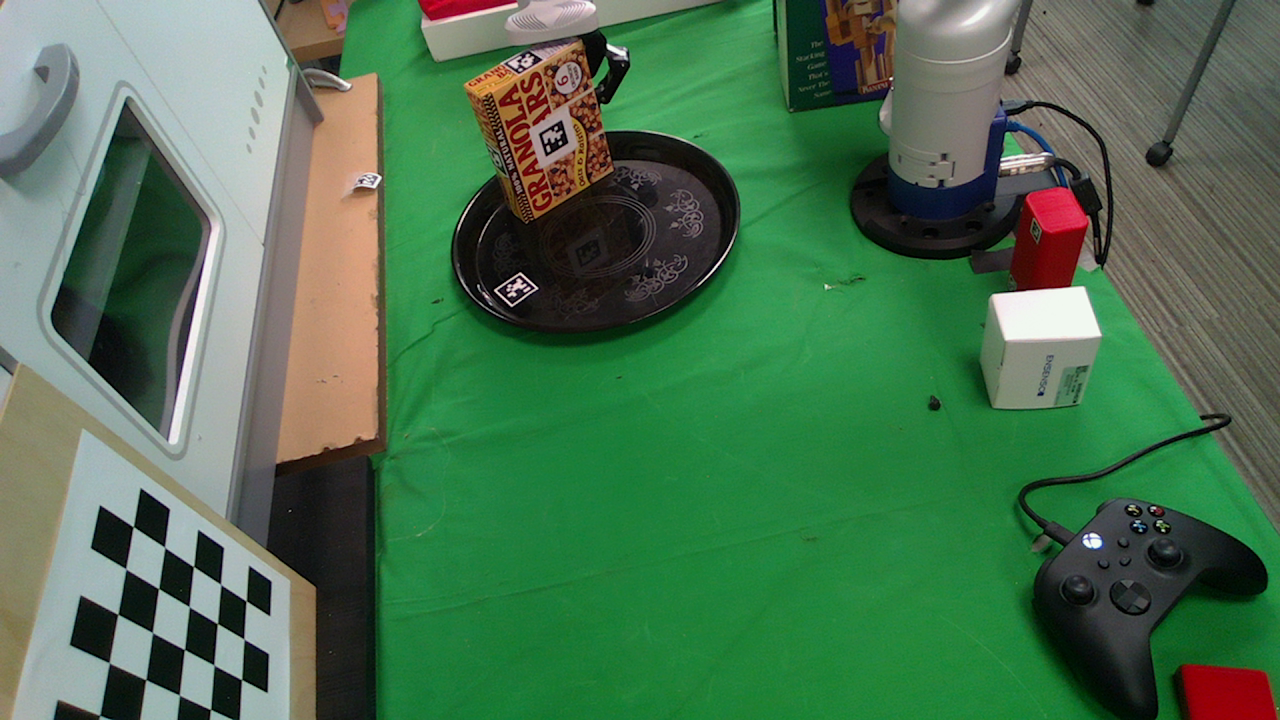}
        \includegraphics[width=0.24\linewidth,trim={10cm 4cm 6cm 0cm},clip]{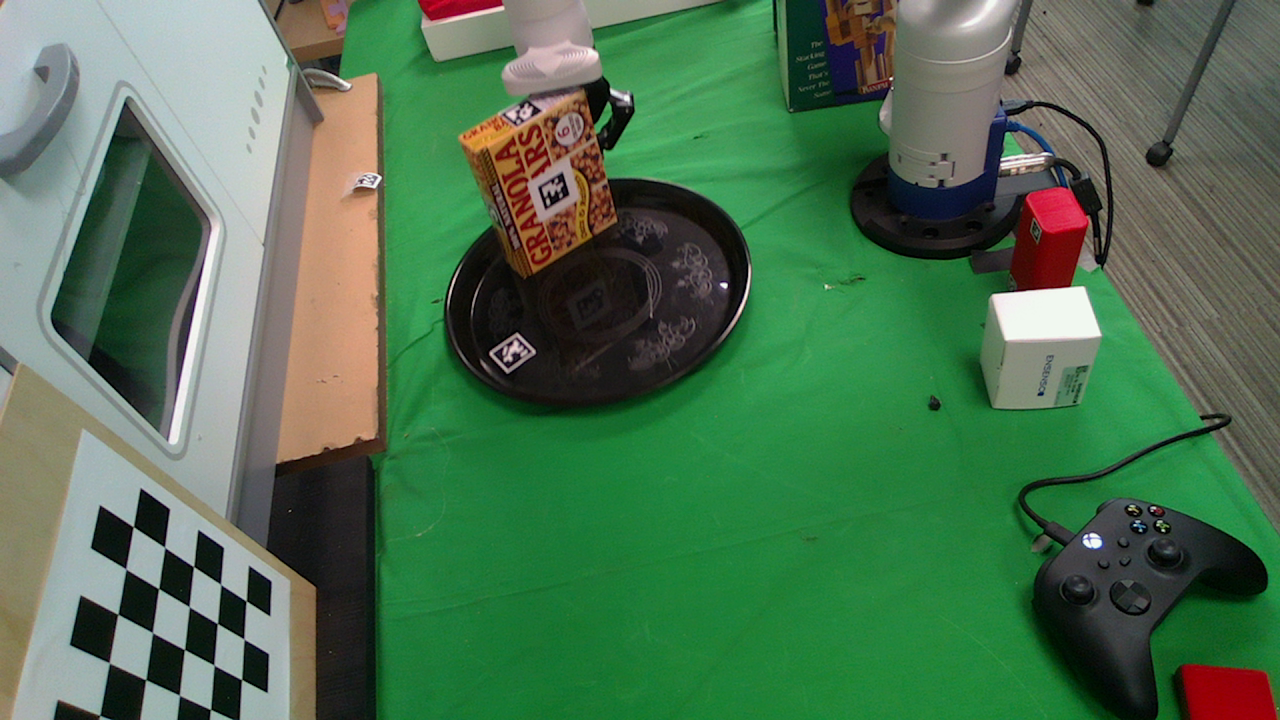}
        \includegraphics[width=0.24\linewidth,trim={10cm 4cm 6cm 0cm},clip]{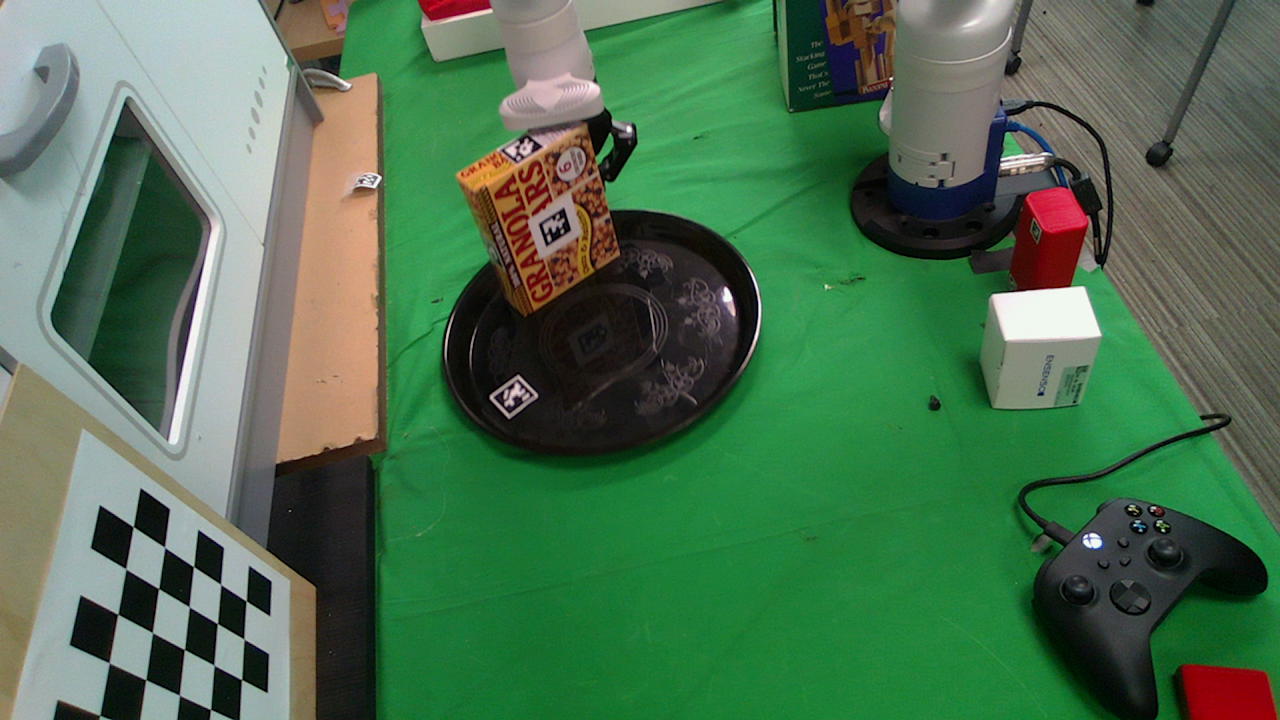}
        \includegraphics[width=0.24\linewidth,trim={10cm 4cm 6cm 0cm},clip]{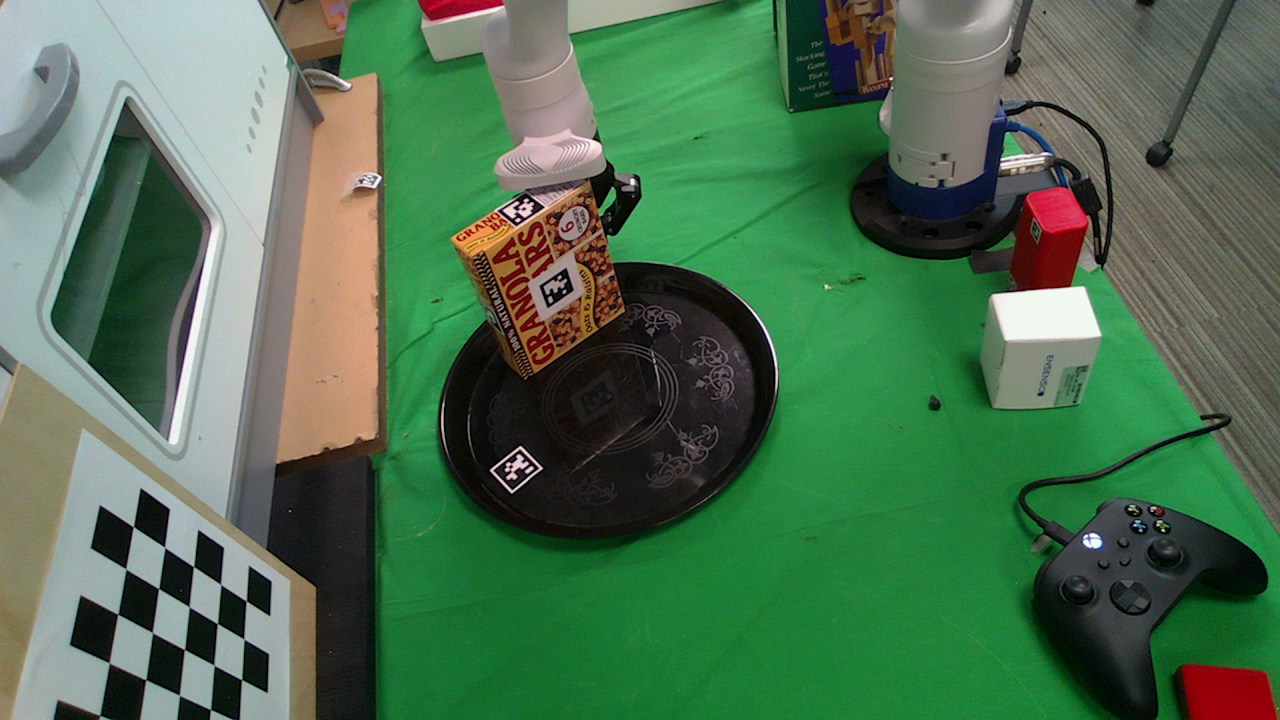}
    \end{minipage}
    \vspace{1em}
    \begin{minipage}[c]{0.05\textwidth}
    \centering
    \rotatebox{90}{\small Bayes Opt.}
    \end{minipage}%
    \begin{minipage}[c]{0.95\textwidth}
        \centering
        \includegraphics[width=0.24\linewidth]{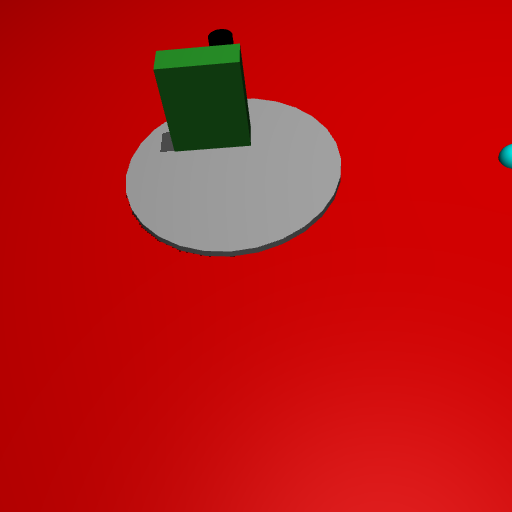}
        \includegraphics[width=0.24\linewidth]{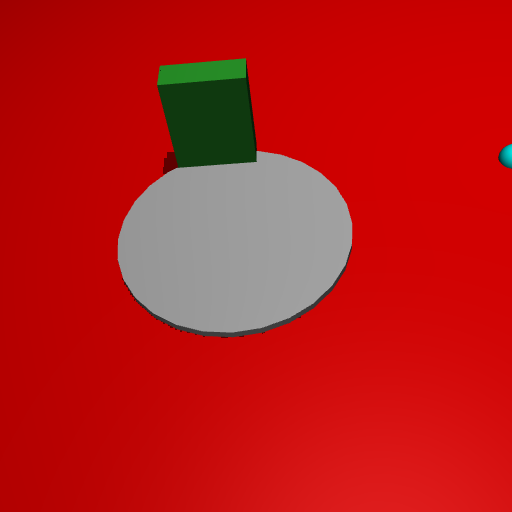}
        \includegraphics[width=0.24\linewidth]{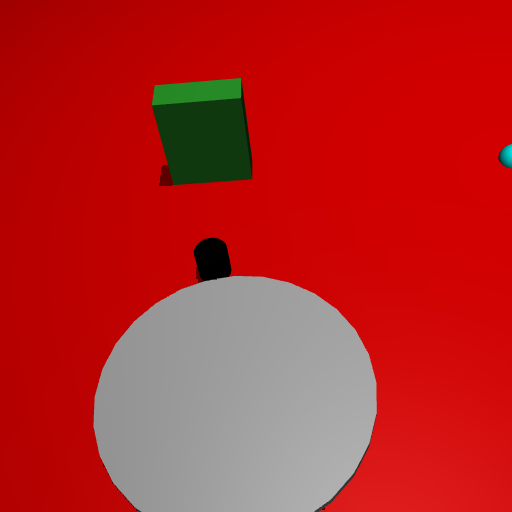}
        \includegraphics[width=0.24\linewidth]{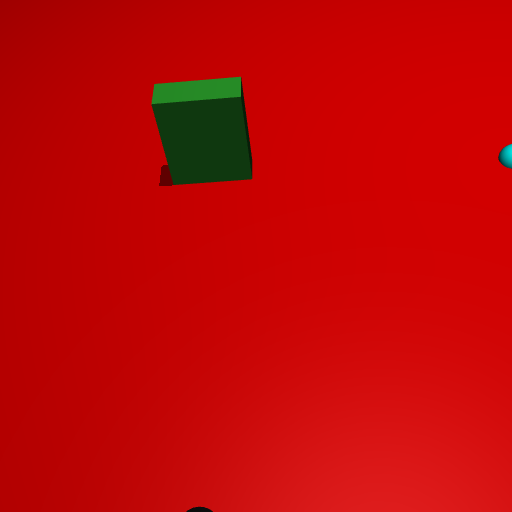}
    \end{minipage}
    \vspace{1em}
    \begin{minipage}[c]{0.05\textwidth}
    \centering
    \rotatebox{90}{\small CMA-ES}
    \end{minipage}%
    \begin{minipage}[c]{0.95\textwidth}
        \centering
        \includegraphics[width=0.24\linewidth]{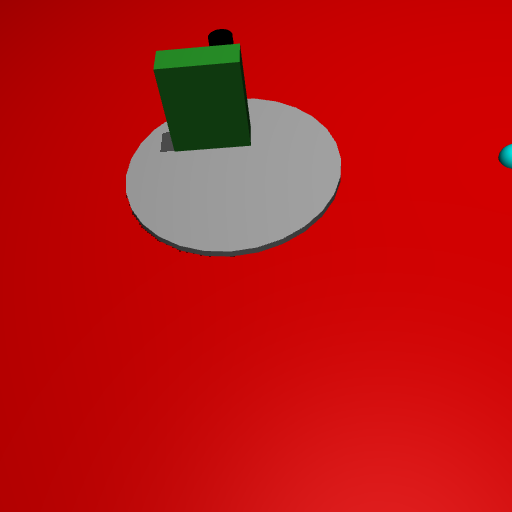}
        \includegraphics[width=0.24\linewidth]{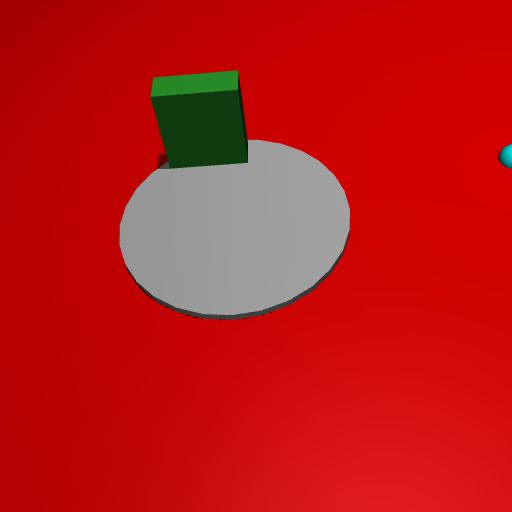}
        \includegraphics[width=0.24\linewidth]{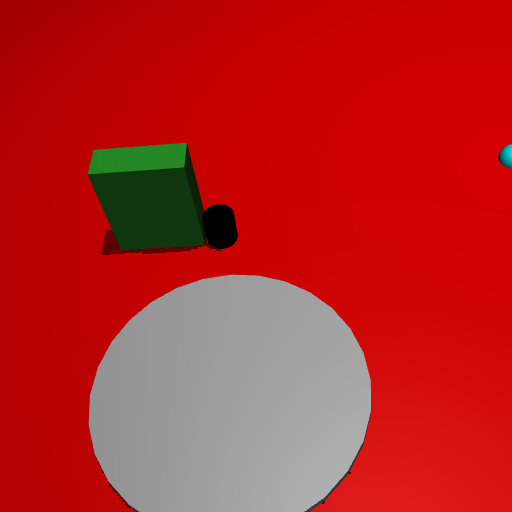}
        \includegraphics[width=0.24\linewidth]{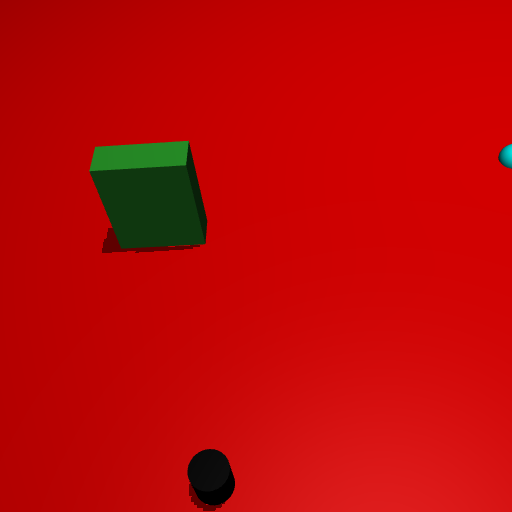}
    \end{minipage}
    \vspace{1em}
    \begin{minipage}[c]{0.05\textwidth}
    \centering
    \rotatebox{90}{\small LLMPhy}
    \end{minipage}%
    \begin{minipage}[c]{0.95\textwidth}
        \centering
        \includegraphics[width=0.24\linewidth]{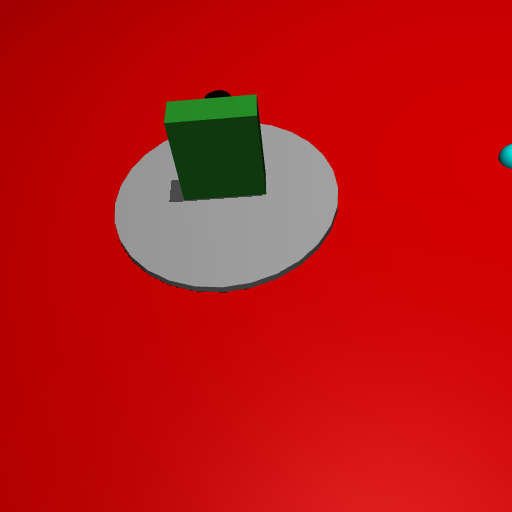}
        \includegraphics[width=0.24\linewidth]{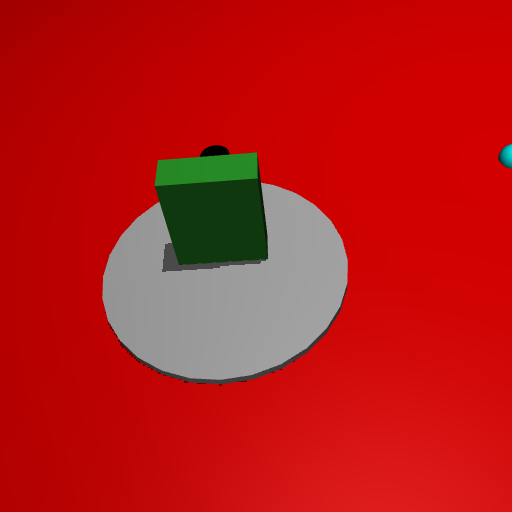}
        \includegraphics[width=0.24\linewidth]{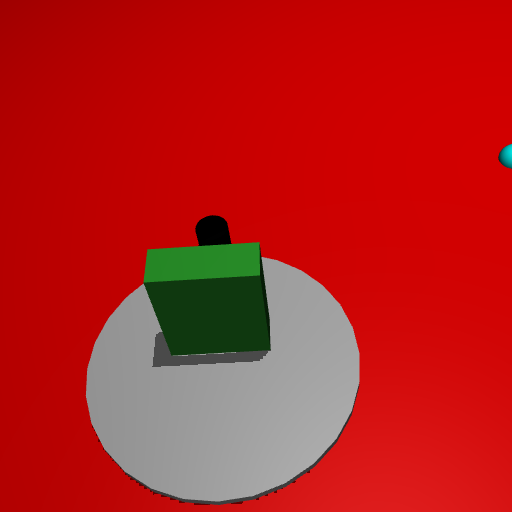}
        \includegraphics[width=0.24\linewidth]{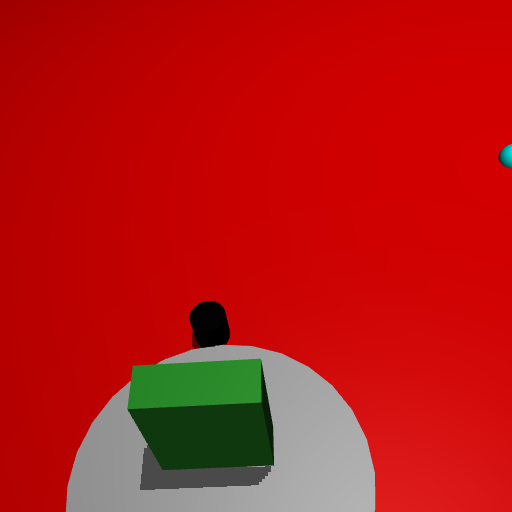}
    \end{minipage}
    \caption{Qualitative comparisons between LLMPhy and other methods on a real sequence. The physics parameters were estimated using a sequence where the pusher moved at 0.1 m/s and the above sequence uses a pusher velocity of 0.2 m/s. Note that the frames are in temporal order but does not correspond to the same time step as different methods show disparate dynamics for the corresponding estimated physics parameters. However, the time steps of original and LLMPhy are closely aligned.}
    \label{fig:qual-results-real-granola}
\end{figure*}

\begin{figure*}
    \centering
    \begin{minipage}[c]{0.05\textwidth}
    \centering
    \rotatebox{90}{\small Original}
    \end{minipage}%
    \begin{minipage}[c]{0.95\textwidth}
        \centering
        \includegraphics[width=0.24\linewidth,trim={10cm 4cm 6cm 0cm},clip]{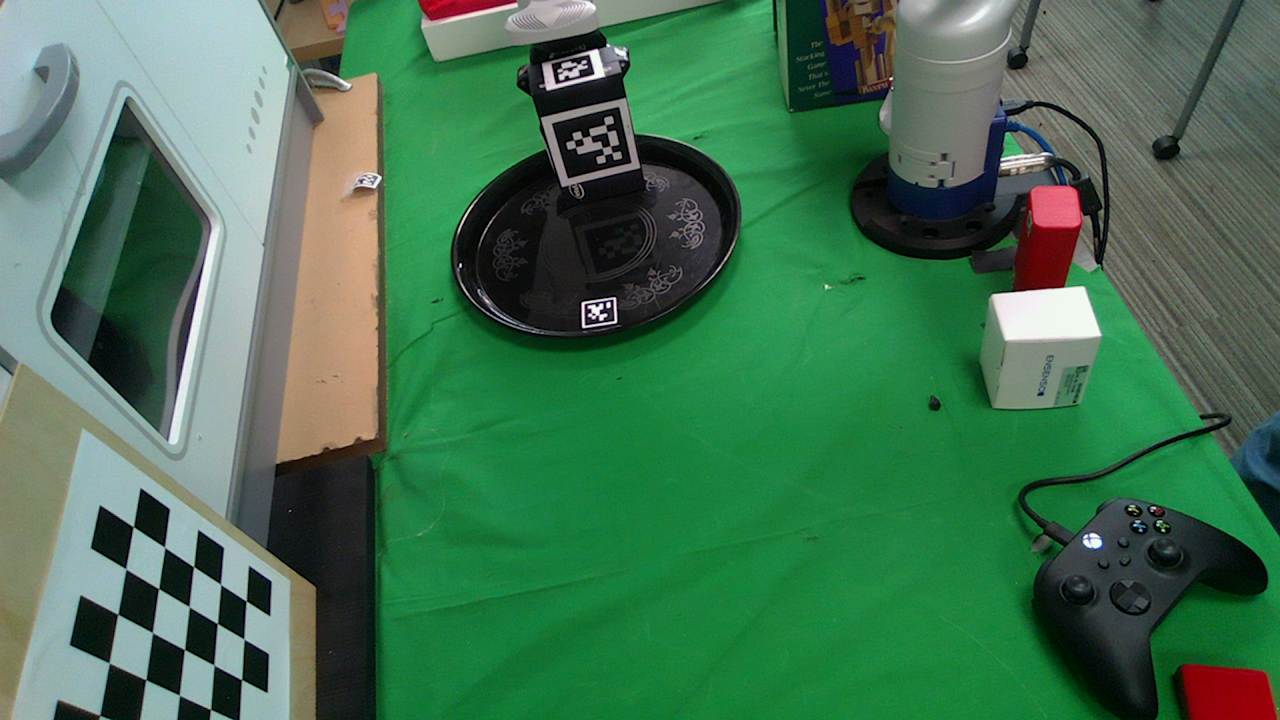}
        \includegraphics[width=0.24\linewidth,trim={10cm 4cm 6cm 0cm},clip]{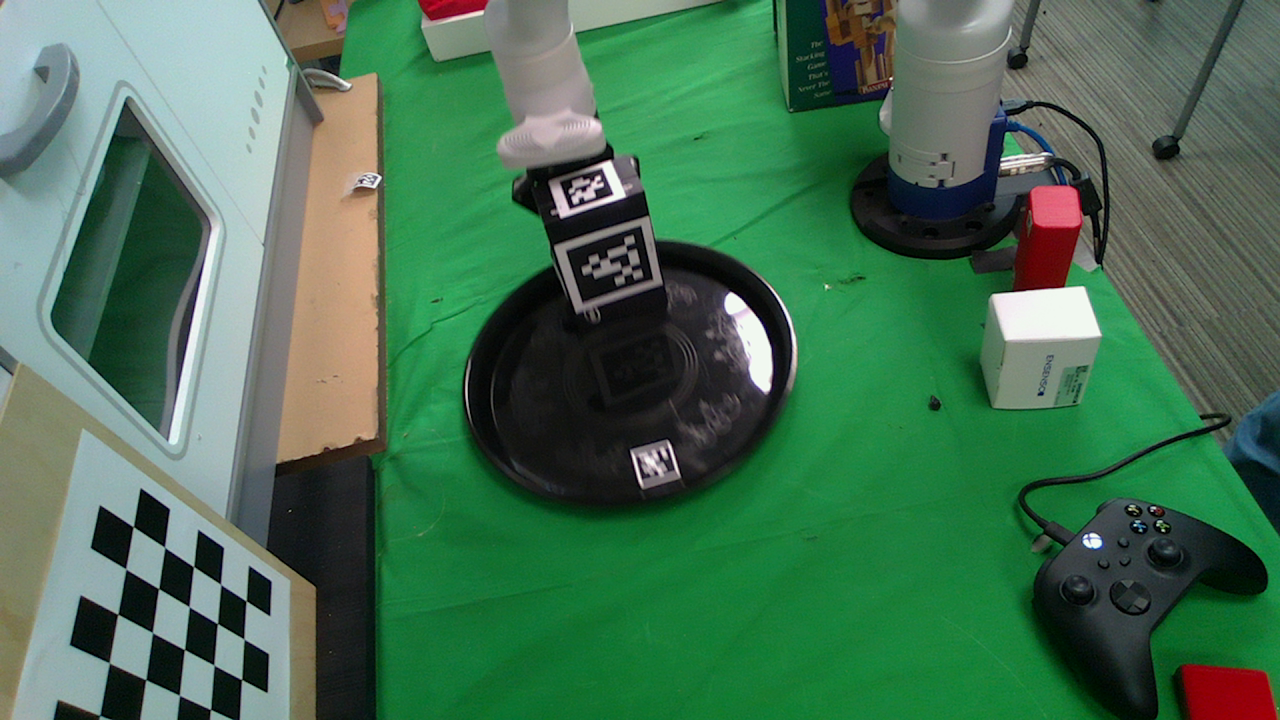}
        \includegraphics[width=0.24\linewidth,trim={10cm 4cm 6cm 0cm},clip]{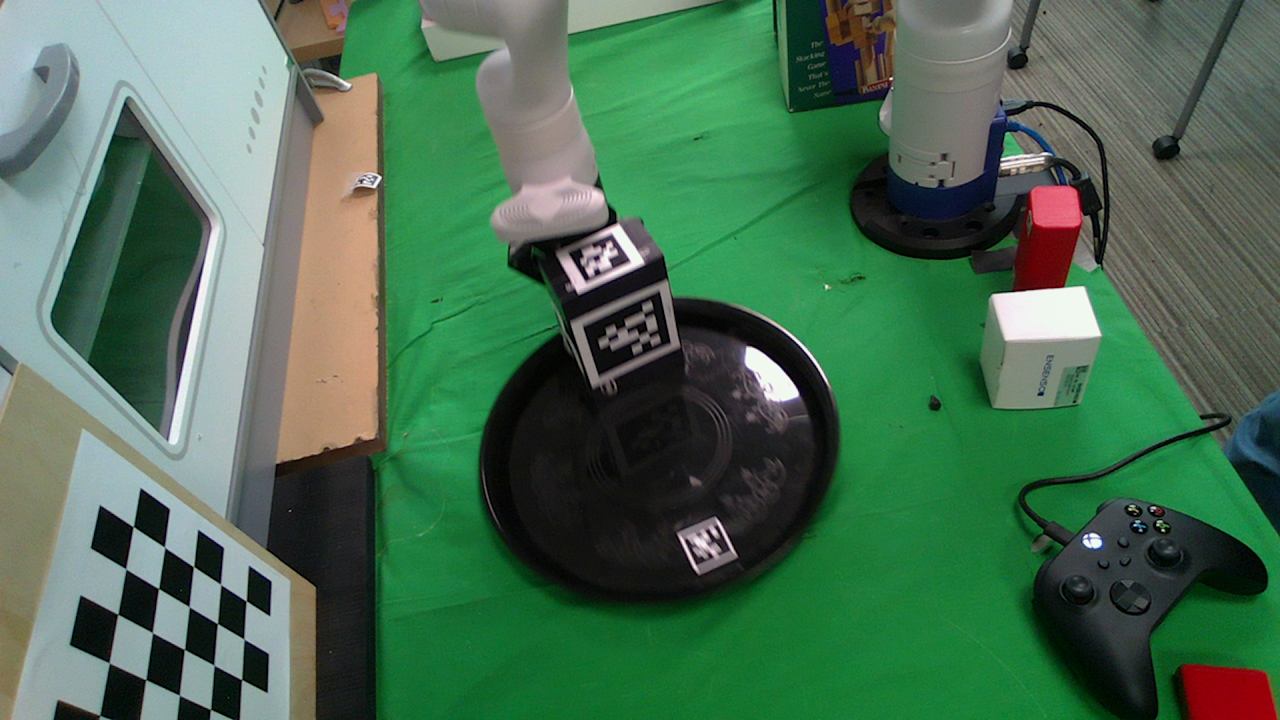}
        \includegraphics[width=0.24\linewidth,trim={10cm 4cm 6cm 0cm},clip]{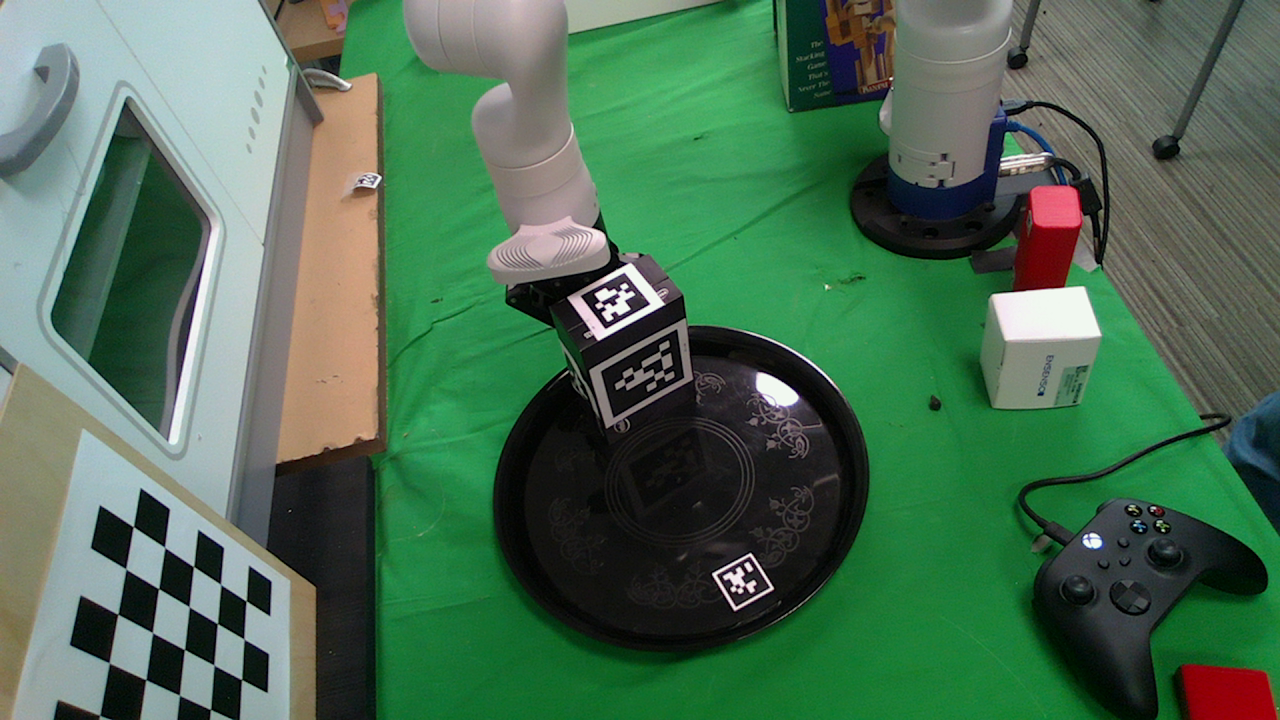}
    \end{minipage}
    \vspace{1em}
    \begin{minipage}[c]{0.05\textwidth}
    \centering
    \rotatebox{90}{\small Bayes Opt.}
    \end{minipage}%
    \begin{minipage}[c]{0.95\textwidth}
        \centering
        \includegraphics[width=0.24\linewidth]{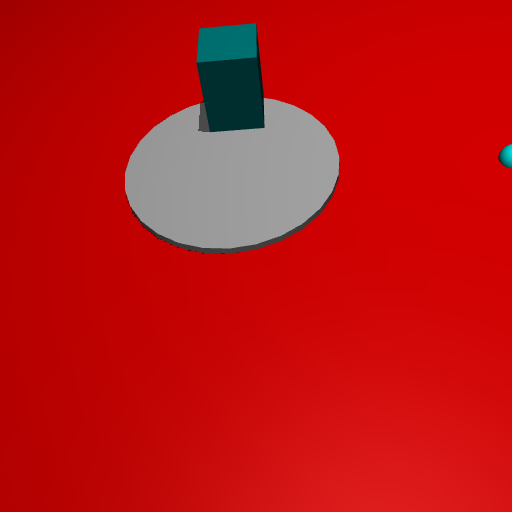}
        \includegraphics[width=0.24\linewidth]{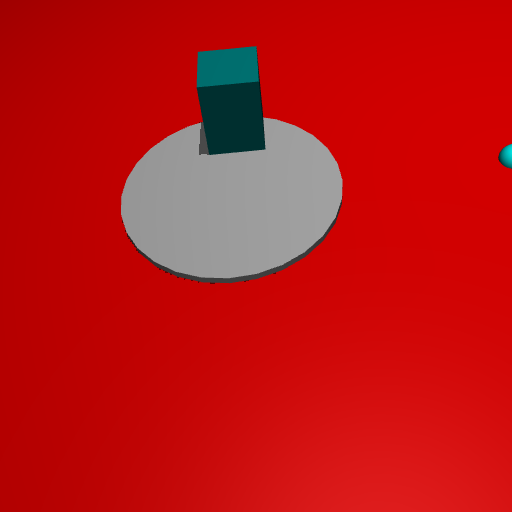}
        \includegraphics[width=0.24\linewidth]{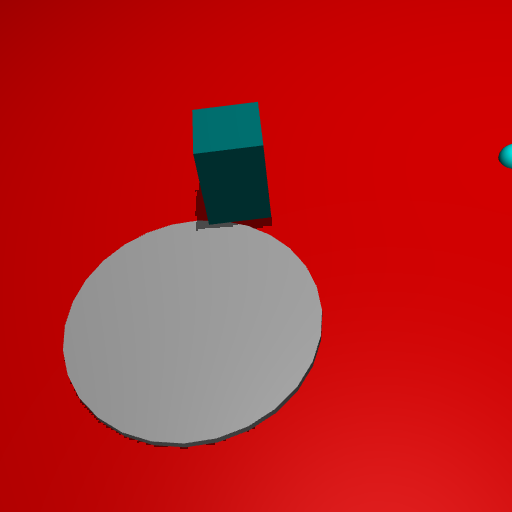}
        \includegraphics[width=0.24\linewidth]{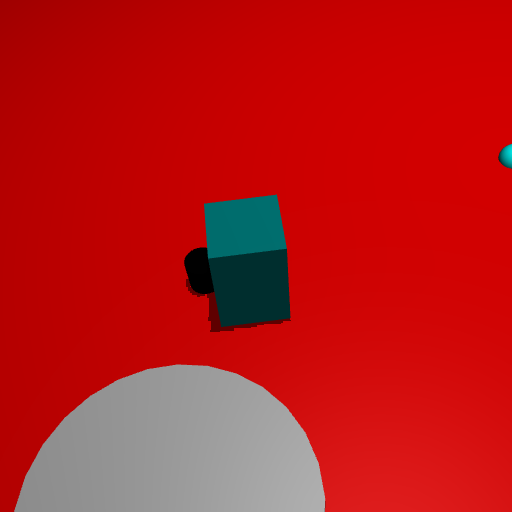}
    \end{minipage}
    \vspace{1em}
    \begin{minipage}[c]{0.05\textwidth}
    \centering
    \rotatebox{90}{\small CMA-ES}
    \end{minipage}%
    \begin{minipage}[c]{0.95\textwidth}
        \centering
        \includegraphics[width=0.24\linewidth]{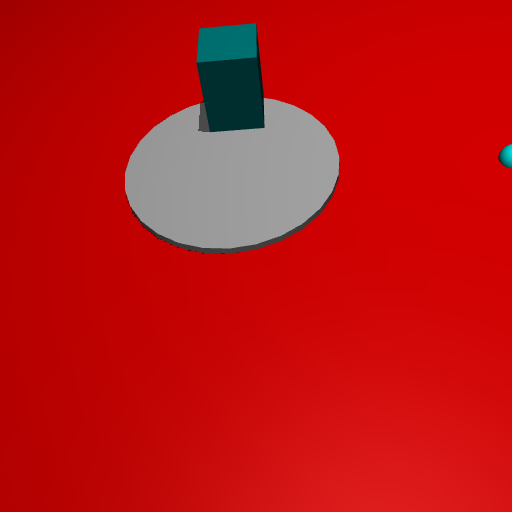}
        \includegraphics[width=0.24\linewidth]{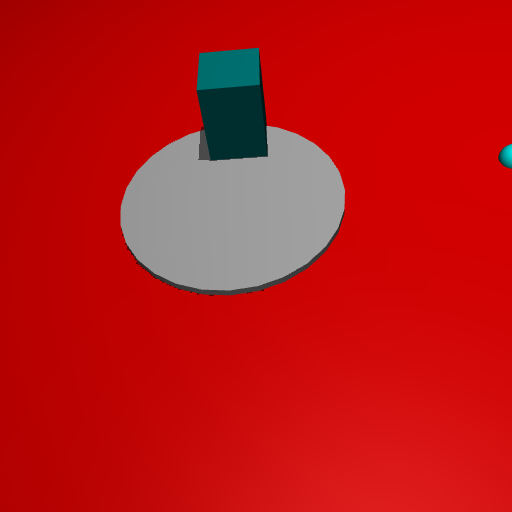}
        \includegraphics[width=0.24\linewidth]{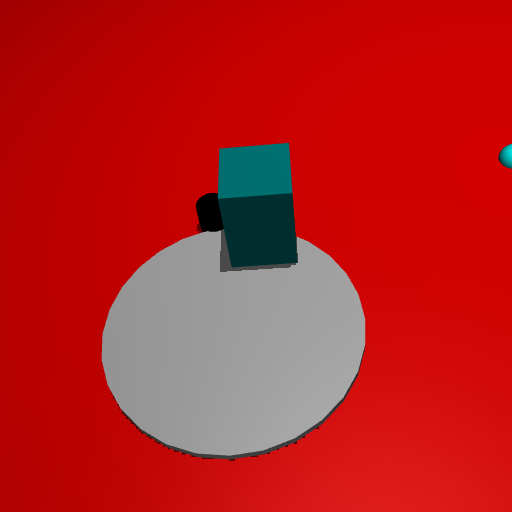}
        \includegraphics[width=0.24\linewidth]{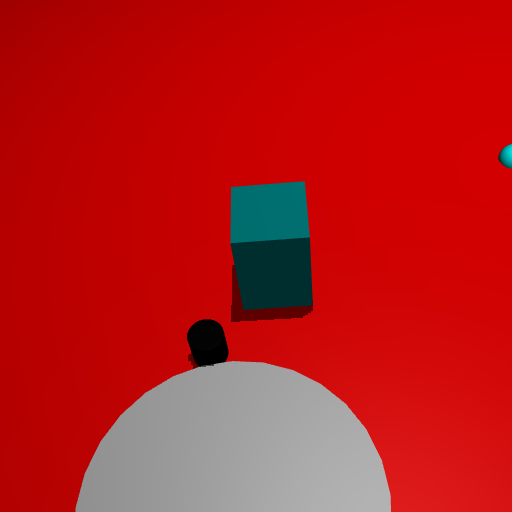}
    \end{minipage}
    \vspace{1em}
    \begin{minipage}[c]{0.05\textwidth}
    \centering
    \rotatebox{90}{\small LLMPhy}
    \end{minipage}%
    \begin{minipage}[c]{0.95\textwidth}
        \centering
        \includegraphics[width=0.24\linewidth]{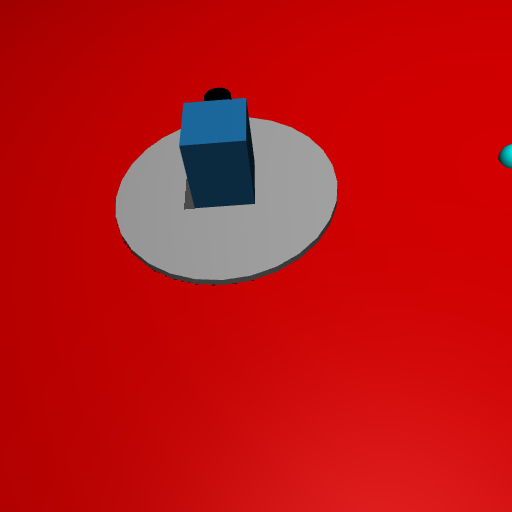}
        \includegraphics[width=0.24\linewidth]{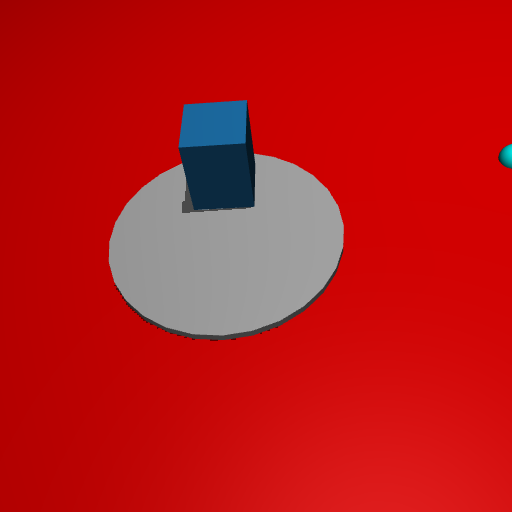}
        \includegraphics[width=0.24\linewidth]{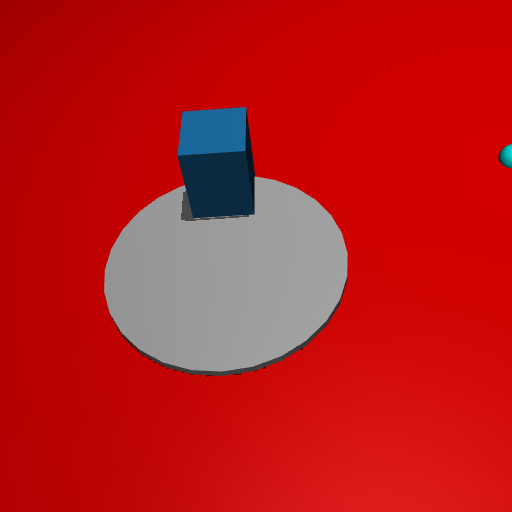}
        \includegraphics[width=0.24\linewidth]{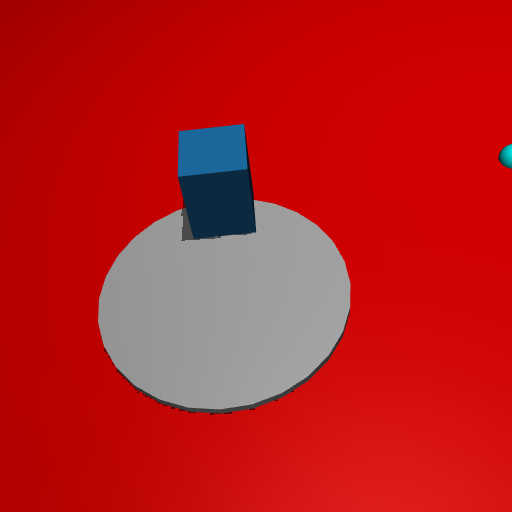}
    \end{minipage}
    \caption{Qualitative comparisons between LLMPhy and other methods on a real sequence. The physics parameters were estimated using a sequence where the pusher moved at 0.2 m/s and the above sequence uses a pusher velocity of 0.3 m/s. Note that the frames are in temporal order but does not correspond to the same time step as different methods show disparate dynamics for the corresponding estimated physics parameters. However, the time steps of original and LLMPhy are closely aligned.}
    \label{fig:qual-results-real-realsense}
\end{figure*}

\noindent\textbf{Qualitative Results}
See Figures~\ref{fig:qual-results-real-granola} and~\ref{fig:qual-results-real-realsense} show qualitative results from our real-world experiments. Please see the attachment videos in the zip file for qualitative videos. 

\begin{figure*}[h]
\centering
\begin{tcolorbox}[colback=gray!5,colframe=black,title=LLMPhy Phase 1: Prompt - Preamble]
The given scene has a tray with three objects (a bottle, a wine\_glass, and a martini\_glass) on it. The radius of the tray is 1.8 and its center of gravity is 0.05 above the ground with a sliding friction of 0.1 and no spin or roll friction. The radius of bottle is 0.4 and its center of gravity is 1.1 above the ground. The center of gravity of the martini\_glass is at a height of 0.5. The center of gravity of the wine\_glass is 0.9 above the ground. The tray is impacted by a pusher and the tray with the objects on it moves. Python code in example\_code\_1.py creates the scene and runs the simulation. The trajectories in object\_traj\_example\_1.txt show the motion of the center of gravity of the objects when running the simulation. Your task is to analyze the given example and then write similar code to produce the trajectories given in `problem\_trajectories.txt`.
\\
\\
You must assume the scene is similar to the one given; however, the physics between the tray and the objects are different. That is, the sliding friction, damping, stiffness, and armature must be adjusted for all the `physical\_parameters\_for\_object\_id\_*` dictionaries in `example\_code\_1.py` so as to reproduce the trajectories in `problem\_trajectories.txt`. You must assume that the physics of the tray with the ground remains the same, as does the external force applied on the tray by the pusher. The trajectories use a timestep of 0.2s. Do not attempt to change the physics parameters beyond their first significant digit. Your written code should strictly follow the same code structure as provided in `example\_code\_1.py`. You may further assume that multiple instances of the same object will have the same physical parameters.
\\
\\
You must not change the `mass` of the objects in your generated code. Do not include the object trajectories in your generated code, as that will fail our simulator.
\end{tcolorbox}
\caption{Preamble prompt for Phase 1 of LLMPhy.}
\label{fig:prompt-phase-1-preamble}
\end{figure*}

\begin{figure*}[t]
\centering
\begin{tcolorbox}[colback=gray!5,colframe=black,title=LLMPhy Phase 1 Prompt -- Program API Interfaces]

Note that the simulation trajectory in \texttt{problem\_trajectories.txt} may use instances of \texttt{bottle}, \texttt{martini\_glass}, and \texttt{wine\_glass}. The object names are provided in the \texttt{problem\_trajectories.txt} file. The masses for the objects are as follows: \texttt{wine\_glass} is 4.0, \texttt{martini\_glass} is 10.0, and \texttt{bottle} is 20.0.

\medskip
\textbf{\texttt{example\_code\_1.py}}
\begin{lstlisting}[style=mypython, basicstyle=\ttfamily\scriptsize]
sim = SIMULATOR_MODEL()
sim.create_pusher('3.0 3.0 0.05')

physical_parameters_for_object_id_tray = {
    'sliding-friction': 0.1,
    'armature': 0.1,
    'stiffness': 0.0,
    'mass': 0.5,
    'damping': 20
}
sim.create_tray(object_physics=physical_parameters_for_object_id_tray)

physical_parameters_for_object_id_1 = {
    'sliding-friction': 0.1,
    'armature': 0.2,
    'stiffness': 0.3,
    'mass': 20.0,  # 'mass' is 20.0 for bottle, 10.0 for martini_glass, and 4.0 for wine_glass
    'damping': 5.7
}
sim.create_object(
    object_id=1,
    object_name='bottle',
    object_location=('row_1', 'column_3'),
    object_color='orange',
    object_physics=physical_parameters_for_object_id_1
)

physical_parameters_for_object_id_2 = {
    'sliding-friction': 0.5,
    'armature': 0.4,
    'stiffness': 1.0,
    'mass': 10.0,  # 'mass' is 20.0 for bottle, 10.0 for martini_glass, and 4.0 for wine_glass
    'damping': 8.8
}
sim.create_object(
    object_id=2,
    object_name='martini_glass',
    object_location=('row_1', 'column_2'),
    object_color='orange',
    object_physics=physical_parameters_for_object_id_2
)

...

sim.create_scene()
sim_out = sim.run_simulation()
del sim
\end{lstlisting}

\medskip
\textbf{\texttt{object\_traj\_example\_1.txt}}

\begin{lstlisting}[basicstyle=\ttfamily\scriptsize,frame=single,breaklines=true]
tray_motion_trajectory (x, y, z) = [(0.0, 0.0, 0.1), (-0.8, -0.8, 0.1), (-1.4, -1.4, 0.1), (-1.8, -1.8, 0.1), (-2.1, -2.1, 0.1), (-2.3, -2.3, 0.1), (-2.4, -2.5, 0.1), (-2.6, -2.6, 0.1), (-2.7, -2.7, 0.1)]

bottle_motion_trajectory (x, y, z) = [(-1.1, -1.1, 1.1), (-1.1, -1.1, 1.1), (-1.1, -1.1, 1.1), (-1.1, -1.1, 1.1), (-1.2, -1.2, 1.1), (-1.3, -1.3, 1.1), (-1.4, -1.5, 1.1), (-1.5, -1.6, 1.1), (-1.6, -1.7, 1.1)]

...
\end{lstlisting}

\end{tcolorbox}
\caption{Phase 1 Prompt Python Examples to inform the LLM about the task, API interfaces, program structure, and input details. }
\label{fig:prompt-phase-1-example}
\end{figure*}

\begin{figure*}[t]
\centering
\begin{tcolorbox}[colback=gray!5,colframe=black,title=LLMPhy Phase 1 Prompt -- Example]
\medskip
\textbf{\texttt{object\_traj\_example\_1.txt}}
\begin{lstlisting}[basicstyle=\ttfamily\scriptsize,frame=single,breaklines=true]

problem\_trajectories.txt

tray\_motion\_trajectory (x, y, z) = [(0.0, 0.0, 0.1), (-0.8, -0.8, 0.1), (-1.3, -1.3, 0.1), (-1.7, -1.7, 0.1), (-2.0, -1.9, 0.1), (-2.2, -2.1, 0.1), (-2.3, -2.2, 0.1), (-2.4, -2.4, 0.1), (-2.5, -2.5, 0.1)]

bottle\_motion\_trajectory (x, y, z) = [(-1.1, -1.1, 1.1), (-1.1, -1.1, 1.2), (-1.2, -1.2, 1.2), (-1.3, -1.3, 1.1), (-1.4, -1.5, 1.1), (-1.6, -1.6, 1.1), (-1.8, -1.8, 1.1), (-1.9, -2.0, 1.1), (-2.1, -2.2, 1.1)]

martini\_glass\_motion\_trajectory (x, y, z) = [(-1.0, 0.0, 0.5), (-1.1, -0.0, 0.6), (-1.3, -0.2, 0.6), (-1.5, -0.4, 0.6), (-1.7, -0.7, 0.6), (-1.9, -0.9, 0.5), (-2.1, -1.1, 0.5), (-2.2, -1.2, 0.5), (-2.3, -1.3, 0.5)]
\end{lstlisting}

\end{tcolorbox}
\caption{Phase 1 Prompt continuation -- We provide the problem trajectories that the model needs to generate program for, following the program structure in Figure~\ref{fig:prompt-phase-1-example}.}
\label{fig:prompt-phase-1-problem-trajectories}
\end{figure*}

\begin{figure*}[h]
\centering
\begin{tcolorbox}[colback=gray!5,colframe=black,title=LLMPhy Phase 2: Prompt - Preamble]
Attached are two images: \\
'example\_1\_top\_down\_view\_1.png' (top-down view) and \\
'example\_1\_side\_view\_2.png' (side view) of the same scene. \\
The top-down view shows a scene arranged roughly on a 3x3 grid. The scene was rendered using the code in 'example\_code\_1.py'. Objects in the scene belong to one of the following classes: {martini\_glass, wine\_glass, bottle} and can be one of the following colors: {purple, red, green, blue, olive, cyan, brown, pink, orange, gray}. Each color appears only once in the scene. Can you interpret the provided code using the images? Use the top-down image to determine the arrangement and color of the objects, and correlate this with the side view to identify the object classes. Each object instance has a unique color, helping you identify the same object across different views.
\\
example\_1\_top\_down\_view\_1.png\\
Image: url\\
\\
example\_1\_side\_view\_2.png\\
Image: url\\
\\
\end{tcolorbox}
\caption{Preamble prompt for Phase 2 of LLMPhy.}
\label{fig:prompt-phase-2-preamble}
\end{figure*}

\begin{figure*}[t]
\centering
\begin{tcolorbox}[colback=gray!5,colframe=black,title=LLMPhy Phase 1 Prompt -- Program API Interfaces]

\medskip
\textbf{\texttt{example\_code\_1.py}}
\begin{lstlisting}[style=mypython, basicstyle=\ttfamily\scriptsize]

#example_code_1.py

sim = SIMULATOR_MODEL()
sim.create_pusher('3.0 3.0 0.05')
physical_parameters_for_object_id_tray = { 
             'sliding-friction': 0.1,
             'armature': 0.1,
             'stiffness': 0.0,
             'mass': 0.5,
             'damping': 20
         }
sim.create_tray(object_physics = physical_parameters_for_object_id_tray)
physical_parameters_for_object_id_1 = { 
             'sliding-friction': 0.1,
             'armature': 0.2,
             'stiffness': 0.3,
             'mass': 20.0, # 'mass' is 20.0 for bottle, 10.0 for martini_glass, and 5.0 for wine_glass
             'damping': 5.7
         }
sim.create_object(object_id=1, object_name='bottle', object_location=('row_2', 'column_3'), object_color='brown', object_physics=physical_parameters_for_object_id_1)

physical_parameters_for_object_id_2 = { 
             'sliding-friction': 0.6,
             'armature': 0.8,
             'stiffness': 0.6,
             'mass': 4.0, # 'mass' is 20.0 for bottle, 10.0 for martini_glass, and 5.0 for wine_glass
             'damping': 8.3
         }
sim.create_object(object_id=2, object_name='wine_glass', object_location=('row_3', 'column_2'), object_color='pink', object_physics=physical_parameters_for_object_id_2)

...

sim.create_scene()
sim_out=sim.run_simulation()
del sim
\end{lstlisting}

Using the above information, write code similar to 'example\_code\_1.py' to reproduce the two images given below for a different scene? The images are named: 'task\_image\_top\_down\_view\_1.png' for the top-down view of the scene and 'task\_image\_side\_view\_2.png' for the side-view of the same scene. Note that not all positions on the grid need to have an object. \\
task\_image\_top\_view\_1.png\\
Image: url\\
task\_image\_side\_view\_2.png\\
Image: url\\
You should further use the following set of physical attributes for the respective objects in the scene when generating the code. (these were estimated in Phase 1). Note that all the instances of the same object use the same physical attributes.\\ 
object\_name: bottle, mass: 20.0, 'sliding-friction': 0.2, 'armature': 0.2, 'stiffness': 0.3, 'damping': 5.7 
\\
object\_name: martini\_glass, mass: 10.0, 'sliding-friction': 0.4, 'armature': 0.4, 'stiffness': 1.0, 'damping': 8.8 \\
object\_name: wine\_glass, mass: 4.0, 'sliding-friction': 0.5, 'armature': 0.8, 'stiffness': 0.6, 'damping': 8.3 \\
\\
\end{tcolorbox}
\caption{Phase 2 Prompt detailing the Python APIs, and the phase goal. The input images are provided as part of the inputs, and not shown. }
\label{fig:prompt-phase-2-problem}
\end{figure*}
\clearpage
\onecolumn
\input{appendix}
\clearpage

\end{document}

%% file: definitions.tex
\newcommand{\AC}[1]{{#1}}

\newcommand{\fr}{\mathbf{x}^v}
\newcommand{\gfr}{\mathbf{x}_g}
\newcommand{\gfrv}{\mathbf{x}^v_g}
\newcommand{\videov}{\mathbf{X}^v}
\newcommand{\pvideo}{\mathbf{X}_{\mathcal{T}}}
\newcommand{\video}{\mathbf{X}}
\newcommand{\videoset}{\mathcal{X}}
\newcommand{\viewset}{\mathcal{V}}
\newcommand{\obj}{o}
\newcommand{\objectset}{\mathcal{C}}
\newcommand{\tobjectset}{\mathcal{C}_{\mathcal{T}}}
\newcommand{\pusher}{p}
\newcommand{\pvel}{\pusher_s}
\newcommand{\task}{\mathcal{T}}
\newcommand{\tuple}[1]{\langle{#1}\rangle}
\newcommand{\set}[1]{\{{#1}\}}
\newcommand{\reals}[1]{\mathbb{R}^{#1}}
\newcommand{\phy}{\Phi}
\newcommand{\question}{Q}
\newcommand{\options}{\mathcal{O}}
\newcommand{\answerset}{\mathcal{A}}
\newcommand{\incontext}{\mathcal{I}}
\newcommand{\layout}{\Psi}
\newcommand{\program}{\pi}
\newcommand{\programset}{\Pi}
\newcommand{\traj}{\tau}

\newcommand{\enorm}[1]{\|{#1}\|}
\newcommand{\psnr}[1]{\PSNR{\!\bigl(#1\bigr)}}
\newcommand{\out}{\Lambda}

\DeclareMathOperator{\llm}{LLM}
\DeclareMathOperator{\simu}{SIM}
\DeclareMathOperator{\llmphy}{LLMPhy}
\DeclareMathOperator{\prompt}{prompt}
\DeclareMathOperator{\error}{error}
\DeclareMathOperator{\concat}{concat}
\DeclareMathOperator{\PSNR}{PSNR}

%% file: appendix.tex
\section{LLMPHY OPTIMIZATION TRACES, PROGRAM SYNTHESIS, AND LLM INTERACTIONS}
\label{sec:llm_opt}
Below, we present the exact prompts we used for the LLM in our experiments for Phases 1 and 2, as well as depicting the programs LLM generate. These results use the TraySim dataset.

\noindent\textbf{Phase 1 Prompt:} 

{\fontfamily{qcr}\selectfont The given scene has a tray with three objects (a bottle, a wine\_glass, and a martini\_glass) on it. The radius of the tray is 1.8 and its center of gravity is 0.05 above the ground with a sliding friction of 0.1 and no spin or roll friction. The radius of bottle is 0.4 and its center of gravity is 1.1 above the ground. The center of gravity of the martini\_glass is at a height of 0.5.The center of gravity of the wine\_glass is 0.9 above the ground. The tray is impacted by a pusher and the tray with the objects on it moves. Python code in example\_code\_1.py creates the scene and runs the simulation. The trajectories in object\_traj\_example\_1.txt show the motion of the center of gravity of the objects when running the simulation. Your task is to analyze the given example and then write similar code to produce the trajectories given in 'problem\_trajectories.txt'. 

You must assume the scene is similar to the one given, however the physics between the tray and the objects are different, that is, the sliding-friction, damping, stiffness, and armature need to be adjusted for all the physical\_parameters\_for\_object\_id\_* dictionaries in the example\_code\_1.py so as to reproduce the trajectories in 'problem\_trajectories.txt'. You must assume that the physics of the tray with the ground remains the same and so is the external force applied on the tray by the pusher. The trajectories use a time step of 0.2s. Do not attempt to change the physics parameters beyond their first significant digit. Your written code should strictly follow the same code structure as provided in example\_code\_1.py. You may further assume that multiple instances of the same object will have the same physical parameters.

You must not change the 'mass' of the objects in your generated code. Do not include the object trajectories in your generated code as that will fail our simulator.

Note that the simulation trajectory in problem\_trajectories.txt may use instances of bottle, martini\_glass, and wine\_glass. The name of the objects is provided in the problem\_trajectories.txt file. The mass for the objects are as follows: wine\_glass is 4.0, martini\_glass is 10.0 and bottle is 20.0.''
}
\begin{verbatim}
\# nexample\_code\_1.py
sim = SIMULATOR_MODEL()
sim.create_pusher('3.0 3.0 0.05')
physical_parameters_for_object_id_tray = { 
             'sliding-friction': 0.1,
             'armature': 0.1,
             'stiffness': 0.0,
             'mass': 0.5,
             'damping': 20
         }
sim.create_tray(object_physics = physical_parameters_for_object_id_tray)
physical_parameters_for_object_id_1 = { 
             'sliding-friction': 0.1, 
             'armature': 0.2, 
             'stiffness': 0.3, 
             'mass': 20.0, 
             'damping': 5.7
         }
sim.create_object(object_id=1, object_name='bottle', 
object_location=('row_1', 'column_3'), object_color='orange', 
object_physics=physical_parameters_for_object_id_1)
...

sim.create_scene()
sim_out=sim.run_simulation()
del sim
\end{verbatim}

\begin{verbatim}
# object\_traj\_example\_1.txt
...

bottle_motion_trajectory (x, y, z) = [(-1.1, -1.1, 1.1), (-1.1, -1.1, 
1.1), (-1.1, -1.1, 1.1), (-1.1, -1.1, 1.1), (-1.2, -1.2, 1.1), (-1.3, 
-1.3, 1.1), (-1.4, -1.5, 1.1), (-1.5, -1.6, 1.1), (-1.6, -1.7, 1.1)]

martini_glass_motion_trajectory (x, y, z) = [(-1.0, 0.0, 0.5), (-1.1, 
-0.0, 0.6), (-1.2, -0.1, 0.6), (-1.4, -0.4, 0.5), (-1.6, -0.6, 0.5), 
(-1.8, -0.8, 0.5), (-2.0, -0.9, 0.5), (-2.1, -1.0, 0.5), (-2.2, -1.1, 
0.5)]

...

\end{verbatim}

\noindent\textbf{Phase 2 Prompt:}
\label{sec:phase2_example}
\begin{lstlisting}
Attached are two images: 'example_1_top_down_view_1.png' (top-down view) 
and 'example_1_side_view_2.png' (side view) of the same scene. The top-
down view shows a scene arranged roughly on a 3x3 grid. The scene was 
rendered using the code in 'example_code_1.py'. Objects in the scene 
belong to one of the following classes: {martini_glass, wine_glass,
bottle} and can be one of the following colors: {purple, red, green,
blue, olive, cyan, brown, pink, orange, gray}. Each color appears only
once in the scene. Can you interpret the provided code using the images?
Use the top-down image to determine the arrangement and color of the
objects, and correlate this with the side view to identify the object
classes. Each object instance has a unique color, helping you identify
the same object across different views.

example_1_top_down_view_1.png
Image: top-down-image url
example_1_side_view_2.png
Image: side-view image url

example_code_1.py

sim = SIMULATOR_MODEL()
sim.create_pusher('3.0 3.0 0.05')
physical_parameters_for_object_id_tray = { 
             'sliding-friction': 0.1,
             'armature': 0.1,
             'stiffness': 0.0,
             'mass': 0.5,
             'damping': 20
         }
sim.create_tray(object_physics = physical_parameters_for_object_id_tray)
physical_parameters_for_object_id_1 = { 
             'sliding-friction': 0.1,
             'armature': 0.2,
             'stiffness': 0.3,
             'mass': 20.0, 
             'damping': 5.7
         }
sim.create_object(object_id=1, object_name='bottle', 
    object_location=('row_2', 'column_3'), object_color='brown', 
    object_physics=physical_parameters_for_object_id_1)

physical_parameters_for_object_id_2 = { 
             'sliding-friction': 0.6,
             'armature': 0.8,
             'stiffness': 0.6,
             'mass': 4.0, 
             'damping': 8.3
         }
sim.create_object(object_id=2, object_name='wine_glass', 
    object_location=('row_3', 'column_2'), object_color='pink', 
    object_physics=physical_parameters_for_object_id_2)

physical_parameters_for_object_id_3 = { 
             'sliding-friction': 0.1,
             'armature': 0.2,
             'stiffness': 0.3,
             'mass': 20.0, 
             'damping': 5.7
         }
sim.create_object(object_id=3, object_name='bottle', 
    object_location=('row_1', 'column_1'), object_color='purple', 
    object_physics=physical_parameters_for_object_id_3)

physical_parameters_for_object_id_4 = { 
             'sliding-friction': 0.1,
             'armature': 0.2,
             'stiffness': 0.3,
             'mass': 20.0, 
             'damping': 5.7
         }
sim.create_object(object_id=4, object_name='bottle', 
    object_location=('row_1', 'column_2'), object_color='olive', 
    object_physics=physical_parameters_for_object_id_4)

physical_parameters_for_object_id_5 = { 
             'sliding-friction': 0.1,
             'armature': 0.2,
             'stiffness': 0.3,
             'mass': 20.0, 
             'damping': 5.7
         }
sim.create_object(object_id=5, object_name='bottle', 
    object_location=('row_3', 'column_1'), object_color='orange', 
    object_physics=physical_parameters_for_object_id_5)

physical_parameters_for_object_id_6 = { 
             'sliding-friction': 0.5,
             'armature': 0.4,
             'stiffness': 1.0,
             'mass': 10.0, 
             'damping': 8.8
         }
sim.create_object(object_id=6, object_name='martini_glass', 
    object_location=('row_2', 'column_2'), 
    object_color='cyan', object_physics=physical_parameters_for_object_id_6)

physical_parameters_for_object_id_7 = { 
             'sliding-friction': 0.5,
             'armature': 0.4,
             'stiffness': 1.0,
             'mass': 10.0, 
             'damping': 8.8
         }
sim.create_object(object_id=7, object_name='martini_glass', 
    object_location=('row_2', 'column_1'), 
    object_color='gray', object_physics=physical_parameters_for_object_id_7)

physical_parameters_for_object_id_8 = { 
             'sliding-friction': 0.5,
             'armature': 0.4,
             'stiffness': 1.0,
             'mass': 10.0, 
             'damping': 8.8
         }
sim.create_object(
    object_id=8,
    object_name='martini_glass',
    object_location=('row_3', 'column_3'),
    object_color='green',
    object_physics=physical_parameters_for_object_id_8
)

physical_parameters_for_object_id_9 = { 
             'sliding-friction': 0.1,
             'armature': 0.2,
             'stiffness': 0.3,
             'mass': 20.0, 
             'damping': 5.7
         }
sim.create_object(object_id=9, object_name='bottle', 
    object_location=('row_1', 'column_3'), 
    object_color='blue', object_physics=physical_parameters_for_object_id_9)

sim.create_scene()
sim_out=sim.run_simulation()
del sim

Using the above information, can you write code similar to 'example_code_1.py' 
to reproduce the 
two images given below for a different scene? The images are named: 
'task_image_top_down_view_1.png' 
for the top-down view of the scene and 'task_image_side_view_2.png' for 
the side-view of the same 
scene. Note that not all positions on the grid need to have an object. 
task_image_top_view_1.png
Image: top-down image url
task_image_side_view_2.png
Image: side-view image url
You should further use the following set of physical attributes for the
respective objects in the scene when generating the code. Note that all
the instances of the same object use the same physical attributes.
object_name: bottle, mass: 20.0, 'sliding-friction': 0.3, 'armature': 0.5, 
'stiffness': 0.4, 'damping': 6.3 

object_name: martini_glass, mass: 10.0, 'sliding-friction': 0.5, 
'armature': 0.3, 'stiffness': 1.2, 'damping': 8.3 

object_name: wine_glass, mass: 4.0, 'sliding-friction': 0.6, 
'armature': 0.7, 'stiffness': 0.8, 'damping': 7.8 

The chat history below shows various attempts of GPT-4o in generating
Python code to reproduce the task images (task_image_top_view_1.png,
task_image_side_view_2.png). For each attempt, we ran the GPT-4o 
generated code in our simulator and found mistakes. Below we provide the
code GPT produced in each attempt, titled GPT Attempt #K, where K is the
attempt number, as well as the PSNR of the generated image against the
given top-down image. Can you refine your code to reproduce the task
images correctly? You should not change any part of the code
corresponding to correctly inferred objects.

GPT Attempt #0:

GPT generated code: 

sim = SIMULATOR_MODEL()
sim.create_pusher('3.0 3.0 0.05')

physical_parameters_for_object_id_tray = {
    'sliding-friction': 0.1,
    'armature': 0.1,
    'stiffness': 0.0,
    'mass': 0.5,
    'damping': 20
}
sim.create_tray(object_physics=physical_parameters_for_object_id_tray)

# Defining the physical parameters for each class of objects
physical_parameters_for_bottle = {
    'sliding-friction': 0.3,
    'armature': 0.5,
    'stiffness': 0.4,
    'mass': 20.0,
    'damping': 6.3
}

physical_parameters_for_wine_glass = {
    'sliding-friction': 0.6,
    'armature': 0.7,
    'stiffness': 0.8,
    'mass': 4.0,
    'damping': 7.8
}

physical_parameters_for_martini_glass = {
    'sliding-friction': 0.5,
    'armature': 0.3,
    'stiffness': 1.2,
    'mass': 10.0,
    'damping': 8.3
}

# Creating objects based on analysis of the task image
sim.create_object(
    object_id=1,
    object_name='bottle',
    object_location=('row_2', 'column_1'),
    object_color='gray',
    object_physics=physical_parameters_for_bottle,
)
sim.create_object(
    object_id=2,
    object_name='wine_glass',
    object_location=('row_1', 'column_1'),
    object_color='yellow',
    object_physics=physical_parameters_for_wine_glass,
)
sim.create_object(
    object_id=3,
    object_name='wine_glass',
    object_location=('row_3', 'column_1'),
    object_color='blue',
    object_physics=physical_parameters_for_wine_glass,
)
sim.create_object(
    object_id=4,
    object_name='martini_glass',
    object_location=('row_2', 'column_2'),
    object_color='orange',
    object_physics=physical_parameters_for_martini_glass,
)
sim.create_object(
    object_id=5,
    object_name='bottle',
    object_location=('row_2', 'column_3'),
    object_color='red',
    object_physics=physical_parameters_for_bottle,
)
sim.create_object(
    object_id=6,
    object_name='bottle',
    object_location=('row_3', 'column_2'),
    object_color='pink',
    object_physics=physical_parameters_for_bottle,
)

# Adding the missing martini glass with cyan color at row 1, column 2
sim.create_object(
    object_id=7,
    object_name='martini_glass',
    object_location=('row_1', 'column_3'),
    object_color='cyan',
    object_physics=physical_parameters_for_martini_glass,
)

# Run the simulation
sim.create_scene()
sim_out = sim.run_simulation()
del sim

Colors of the objects in the code above that are misplaced: {'purple', 'cyan', 'pink'}

PSNR for the generated image against given top-down image = 40.9
Please check the locations of these objects in task_image_top_view_1.png and 
fix the code accordingly.

End of Prompt
\end{lstlisting}

\section{Example of Synthesized Programs}
\label{sec:example_pgms}
Below, we show an example of Python programs synthesized by $\llmphy$, which are then executed to invoke the Mujoco Simulator to run and render the dynamical sequence, to extract the outcome of the simulation. 

\noindent\textbf{Phase 1 program:}
\begin{verbatim}
sim = SIMULATOR_MODEL()
sim.create_pusher('3.0 3.0 0.05')
physical_parameters_for_object_id_tray = { 
             'sliding-friction': 0.1,
             'armature': 0.1,
             'stiffness': 0.0,
             'mass': 0.5,
             'damping': 20
         }
sim.create_tray(object_physics = physical_parameters_for_object_id_tray)

physical_parameters_for_object_id_1 = { 
    'sliding-friction': 0.18,  # finely adjusted as previous good attempt
    'armature': 0.1,           # keeping consistent based on the effective value
    'stiffness': 0.16,         # slight readjustment from previous effective value
    'mass': 20.0,              # mass remains 20.0 for bottle
    'damping': 4.5             # slight decrease for finer adjustment
}

physical_parameters_for_object_id_2 = { 
    'sliding-friction': 0.35,  # minor adjustment from better attempt
    'armature': 0.26,          # further decreasing for better control
    'stiffness': 0.75,         # slight decrease for better compliance
    'mass': 10.0,              # mass remains 10.0 for martini_glass
    'damping': 6.25            # slight adjustment for better trajectory control
}

physical_parameters_for_object_id_3 = { 
    'sliding-friction': 0.42,  # slight decrease for better fit
    'armature': 0.58,          # minor decrease for better control
    'stiffness': 0.39,         # minor decrease for better adjustment
    'mass': 4.0,               # mass remains 4.0 for wine_glass
    'damping': 6.55            # slight decrease for finer adjustment
}
sim.create_object(object_id=1, object_name='bottle', 
    object_location=('row_1', 'column_3'), 
    object_color='orange', 
    object_physics=physical_parameters_for_object_id_1)

sim.create_object(object_id=2, object_name='martini_glass',    
    object_location=('row_1', 'column_2'), 
    object_color='orange', 
    object_physics=physical_parameters_for_object_id_2)

sim.create_object(object_id=3, object_name='wine_glass', 
    object_location=('row_1', 'column_1'), 
        object_color='orange', 
        object_physics=physical_parameters_for_object_id_3)

sim.create_scene()
sim_out = sim.run_simulation()
del sim
\end{verbatim} 

\noindent\textbf{Phase 2 program:}
\begin{verbatim}
sim = SIMULATOR_MODEL()
sim.create_pusher('3.0 3.0 0.05')

# Set physical attributes for the tray
physical_parameters_for_object_id_tray = {
    'sliding-friction': 0.1,
    'armature': 0.1,
    'stiffness': 0.0,
    'mass': 0.5,
    'damping': 20
}
sim.create_tray(object_physics=physical_parameters_for_object_id_tray)

# Physical parameters for each class of objects
physical_parameters_bottle = {
    'sliding-friction': 0.9,
    'armature': 0.2,
    'stiffness': 0.3,
    'mass': 20.0,
    'damping': 6.2
}

physical_parameters_wine_glass = {
    'sliding-friction': 0.7,
    'armature': 0.4,
    'stiffness': 0.6,
    'mass': 4.0,
    'damping': 3.4
}

physical_parameters_martini_glass = {
    'sliding-friction': 0.7,
    'armature': 0.6,
    'stiffness': 0.8,
    'mass': 10.0,
    'damping': 1.1
}

# Create objects with corrected positions and colors
sim.create_object(object_id=1, object_name='wine_glass', 
    object_location=('row_1', 'column_1'), 
    object_color='gray',
    object_physics=physical_parameters_wine_glass)
    
sim.create_object(object_id=2, object_name='wine_glass', 
    object_location=('row_1', 'column_2'), 
    object_color='red', 
    object_physics=physical_parameters_wine_glass)
    
sim.create_object(object_id=3, object_name='wine_glass', 
    object_location=('row_1', 'column_3'), 
    object_color='pink', 
    object_physics=physical_parameters_wine_glass)
    
sim.create_object(object_id=4, object_name='martini_glass', 
    object_location=('row_2', 'column_1'), 
    object_color='cyan', 
    object_physics=physical_parameters_martini_glass)

sim.create_object(object_id=5, object_name='bottle', 
    object_location=('row_2', 'column_2'), 
    object_color='yellow', 
    object_physics=physical_parameters_bottle)

sim.create_object(object_id=6, object_name='bottle', 
    object_location=('row_2', 'column_3'), 
    object_color='purple', 
    object_physics=physical_parameters_bottle)

sim.create_object(object_id=7, object_name='martini_glass', 
    object_location=('row_3', 'column_1'), object_color='blue', 
    object_physics=physical_parameters_martini_glass)

sim.create_object(object_id=8, object_name='bottle', 
    object_location=('row_3', 'column_2'), 
    object_color='orange', 
    object_physics=physical_parameters_bottle)

sim.create_object(object_id=9, object_name='martini_glass', 
    object_location=('row_3', 'column_3'), 
    object_color='brown', 
    object_physics=physical_parameters_martini_glass)

sim.create_scene()
sim_out = sim.run_simulation()
del sim

\end{verbatim}

\section{LLMPhy Optimization and Interaction Trace (Phase1)}
\label{sec:trace_phase1}

Below, we provide the full log of the optimization iterations produced in phase 1 by $\llmphy$. Specifically, the text shows how the LLM uses the provided feedback to adjust the physical attributes, with explanation of the justification for the adjustments. Note that each response from the LLM is separated by strings "=======".

\begin{lstlisting}
In order to produce the desired trajectories given in `problem_trajectories.txt`, 
we need to adjust the physics parameters for the objects (`bottle`, 
`wine_glass`, and `martini_glass`) in the scene. Based on the instructions, 
we need to adjust the `sliding-friction`, `damping`, `stiffness`, and 
`armature` parameters. We should follow the given code structure closely and 
only change the parameters to their first significant digit.

Below is the adjusted Python code for `example_code_1.py` to produce the 
desired trajectories:

```python
# example_code_problem.py

sim = SIMULATOR_MODEL()
sim.create_pusher('3.0 3.0 0.05')

# The physics of the tray with the ground remains the same
physical_parameters_for_object_id_tray = { 
             'sliding-friction': 0.1,
             'armature': 0.1,
             'stiffness': 0.0,
             'mass': 0.5,
             'damping': 20
         }
sim.create_tray(object_physics = physical_parameters_for_object_id_tray)

# Adjusted physics parameters for the bottle
physical_parameters_for_object_id_1 = { 
             'sliding-friction': 0.2,
             'armature': 0.3,
             'stiffness': 0.4,
             'mass': 20.0,  # mass remains unchanged
             'damping': 6.5
         }
sim.create_object(
    object_id=1,
    object_name='bottle',
    object_location=('row_1', 'column_3'),
    object_color='orange',
    object_physics=physical_parameters_for_object_id_1
)

# Adjusted physics parameters for the martini_glass
physical_parameters_for_object_id_2 = { 
             'sliding-friction': 0.6,
             'armature': 0.5,
             'stiffness': 1.1,
             'mass': 10.0,  # mass remains unchanged
             'damping': 9.0
         }
sim.create_object(
    object_id=2,
    object_name='martini_glass',
    object_location=('row_1', 'column_2'),
    object_color='orange',
    object_physics=physical_parameters_for_object_id_2
)

# Adjusted physics parameters for the wine_glass
physical_parameters_for_object_id_3 = { 
             'sliding-friction': 0.8,
             'armature': 0.9,
             'stiffness': 1.0,
             'mass': 4.0,  # mass remains unchanged
             'damping': 8.4
         }
sim.create_object(
    object_id=3,
    object_name='wine_glass',
    object_location=('row_1', 'column_1'),
    object_color='orange',
    object_physics=physical_parameters_for_object_id_3
)

sim.create_scene()
sim_out = sim.run_simulation()
del sim
```

In this adjusted code, I have modified the `sliding-friction`, `damping`, `stiffness`, 
and `armature` parameters for each object to produce the desired motion trajectories 
as suggested by the `problem_trajectories.txt` file. These modifications respect 
the constraint that parameters can only be adjusted up to their first significant digit.

You can now run this simulation code to generate the trajectories which should 
match closely to the ones provided in `problem_trajectories.txt`.

========

Let's see if we can refine the physical parameters for the objects to reduce the 
trajectory errors. To do this, we need to adjust the parameters more carefully 
while adhering to the given constraints.

Let's analyze the trajectories and try to make minor adjustments to the parameters:

1. **Bottle:** The predicted trajectory error is 1.9, implying the need for finer 
adjustments. The last attempt modified the parameters to:
    - 'sliding-friction': 0.2
    - 'armature': 0.3
    - 'stiffness': 0.4
    - 'damping': 6.5

2. **Martini_glass:** The predicted trajectory error is relatively high at 3.16. 
We will need to make significant adjustments.
    - 'sliding-friction': 0.6
    - 'armature': 0.5
    - 'stiffness': 1.1
    - 'damping': 9.0
    
   Let's reduce the damping a bit and adjust other parameters slightly.

3. **Wine_glass:** The predicted trajectory error is low at 0.24. Therefore, fine 
adjustments may be enough.
    - 'sliding-friction': 0.8
    - 'armature': 0.9
    - 'stiffness': 1.0
    - 'damping': 8.4

### New physical parameters:

```python
sim = SIMULATOR_MODEL()
sim.create_pusher('3.0 3.0 0.05')
physical_parameters_for_object_id_tray = { 
    'sliding-friction': 0.1,
    'armature': 0.1,
    'stiffness': 0.0,
    'mass': 0.5,
    'damping': 20
}

sim.create_tray(object_physics = physical_parameters_for_object_id_tray)

physical_parameters_for_object_id_1 = { 
    'sliding-friction': 0.25,   # slight increase in friction
    'armature': 0.4,            # slight increase in armature
    'stiffness': 0.5,           # slight increase in stiffness
    'mass': 20.0,               # mass remains unchanged
    'damping': 7.0              # slight increase in damping
}

sim.create_object(
    object_id=1,
    object_name='bottle',
    object_location=('row_1', 'column_3'),
    object_color='orange',
    object_physics=physical_parameters_for_object_id_1
)

physical_parameters_for_object_id_2 = { 
    'sliding-friction': 0.55,   # slight decrease in friction
    'armature': 0.6,            # slight increase in armature
    'stiffness': 1.2,           # slight increase in stiffness
    'mass': 10.0,               # mass remains unchanged
    'damping': 8.5              # decrease in damping
}

sim.create_object(
    object_id=2,
    object_name='martini_glass',
    object_location=('row_1', 'column_2'),
    object_color='orange',
    object_physics=physical_parameters_for_object_id_2
)

physical_parameters_for_object_id_3 = { 
    'sliding-friction': 0.75,   # slight decrease in friction
    'armature': 1.0,            # increase in armature
    'stiffness': 0.9,           # slight decrease in stiffness
    'mass': 4.0,                # mass remains unchanged
    'damping': 8.2              # slight decrease in damping
}

sim.create_object(
    object_id=3,
    object_name='wine_glass',
    object_location=('row_1', 'column_1'),
    object_color='orange',
    object_physics=physical_parameters_for_object_id_3,
)

sim.create_scene()
sim_out=sim.run_simulation()
del sim
```

Please use the above code to run the simulation again and see if the trajectory 
errors decrease. Adjustments have been made based on the analysis of the previous attempt.

========

Based on the trend and error information from the previous attempts, we'll refine the 
parameters further to attempt a better match for the trajectories in 
'problem_trajectories.txt'. The key insights from previous attempts suggest that we need finer adjustments, 
particularly to friction, damping, and stiffness.

Here's an updated attempt:

```python
# Code structure and constants same as provided

sim = SIMULATOR_MODEL()
sim.create_pusher('3.0 3.0 0.05')
physical_parameters_for_object_id_tray = {
    'sliding-friction': 0.1,
    'armature': 0.1,
    'stiffness': 0.0,
    'mass': 0.5,
    'damping': 20
}
sim.create_tray(object_physics=physical_parameters_for_object_id_tray)

# Refine bottle parameters
physical_parameters_for_object_id_1 = {
    'sliding-friction': 0.23,   # slight increase in friction
    'armature': 0.45,           # increase in armature
    'stiffness': 0.55,          # slight increase in stiffness
    'mass': 20.0,               # mass remains unchanged
    'damping': 7.5              # increase in damping
}
sim.create_object(
    object_id=1,
    object_name='bottle',
    object_location=('row_1', 'column_3'),
    object_color='orange',
    object_physics=physical_parameters_for_object_id_1
)

# Refine martini_glass parameters
physical_parameters_for_object_id_2 = {
    'sliding-friction': 0.54,   # slight decrease in friction
    'armature': 0.65,           # increase in armature
    'stiffness': 1.25,          # increase in stiffness
    'mass': 10.0,               # mass remains unchanged
    'damping': 8.3              # slight decrease in damping
}
sim.create_object(
    object_id=2,
    object_name='martini_glass',
    object_location=('row_1', 'column_2'),
    object_color='orange',
    object_physics=physical_parameters_for_object_id_2
)

# Refine wine_glass parameters
physical_parameters_for_object_id_3 = {
    'sliding-friction': 0.72,   # slight decrease in friction
    'armature': 1.05,           # increase in armature
    'stiffness': 0.95,          # slight decrease in stiffness
    'mass': 4.0,                # mass remains unchanged
    'damping': 8.1              # decrease in damping
}
sim.create_object(
    object_id=3,
    object_name='wine_glass',
    object_location=('row_1', 'column_1'),
    object_color='orange',
    object_physics=physical_parameters_for_object_id_3,
)

# Run the simulation
sim.create_scene()
sim_out = sim.run_simulation()
del sim
```

This refinement aims to balance the dynamics and interactions of the objects with 
the tray, considering the error metrics from previous attempts. Monitoring the 
changes in each property helps to iteratively converge towards a more accurate 
trajectory.

========

Based on the provided attempts and their errors, let's attempt to refine the physical 
parameters for the objects in such a way to specifically address the notable errors. 
Here's another attempt:

1. The bottle (`object_id=1`) has a relatively high trajectory error, so we may need to 
adjust its parameters further to get a close match.
2. The martini_glass (`object_id=2`) has the highest trajectory error, requiring 
focus on reducing its discrepancy.
3. The wine_glass (`object_id=3`) has the least error, suggesting that its parameters 
are close to what's needed but might still need slight adjustments to further align 
trajectories.

Considering these points, we'll make small but precise adjustments, particularly 
targeting increased damping for the bottle, slightly decreasing friction for the 
martini_glass, and incrementally changing the armature and stiffness for both 
martini_glass and wine_glass.

Here is the code:

```python
# Generated Python code to accurately reproduce the trajectories

sim = SIMULATOR_MODEL()
sim.create_pusher('3.0 3.0 0.05')

physical_parameters_for_object_id_tray = { 
    'sliding-friction': 0.1,
    'armature': 0.1,
    'stiffness': 0.0,
    'mass': 0.5,
    'damping': 20
}
sim.create_tray(object_physics = physical_parameters_for_object_id_tray)

physical_parameters_for_object_id_1 = {
    'sliding-friction': 0.22,  # slight decrease in friction to adjust movement
    'armature': 0.43,          # slight decrease in armature
    'stiffness': 0.53,         # slight decrease in stiffness
    'mass': 20.0,              # mass remains unchanged
    'damping': 7.4             # slight decrease in damping
}
sim.create_object(
    object_id=1,
    object_name='bottle',
    object_location=('row_1', 'column_3'),
    object_color='orange',
    object_physics=physical_parameters_for_object_id_1
)

physical_parameters_for_object_id_2 = {
    'sliding-friction': 0.52,  # slight decrease in friction to reduce error
    'armature': 0.62,          # slight decrease in armature
    'stiffness': 1.23,         # slight decrease in stiffness
    'mass': 10.0,              # mass remains unchanged
    'damping': 8.1             # slight decrease in damping to reduce discrepancy
}
sim.create_object(
    object_id=2,
    object_name='martini_glass',
    object_location=('row_1', 'column_2'),
    object_color='orange',
    object_physics=physical_parameters_for_object_id_2
)
physical_parameters_for_object_id_3 = {
    'sliding-friction': 0.71,  # slight decrease in friction for close alignment
    'armature': 1.04,          # slight decrease in armature
    'stiffness': 0.94,         # slight decrease in stiffness
    'mass': 4.0,               # mass remains unchanged
    'damping': 8.0             # slight decrease in damping for close alignment
}
sim.create_object(
    object_id=3,
    object_name='wine_glass',
    object_location=('row_1', 'column_1'),
    object_color='orange',
    object_physics=physical_parameters_for_object_id_3
)

sim.create_scene()
sim_out=sim.run_simulation()
del sim
```

Make sure to run this code through your simulation environment to verify the 
trajectory errors. The adjustments are subtle yet precise to better align with the required trajectories in 
'problem_trajectories.txt'.

========

To refine the parameters and reduce the trajectory prediction error, I will make 
gradual and strategic adjustments while making sure not to alter beyond the first 
significant digit. Below are the physical parameters for each object, followed by 
the refined code that will be used to generate the trajectories.

```python
# refined_code.py

sim = SIMULATOR_MODEL()
sim.create_pusher('3.0 3.0 0.05')

physical_parameters_for_object_id_tray = { 
    'sliding-friction': 0.1,
    'armature': 0.1,
    'stiffness': 0.0,
    'mass': 0.5,
    'damping': 20
}
sim.create_tray(object_physics=physical_parameters_for_object_id_tray)

physical_parameters_for_object_id_1 = { 
    'sliding-friction': 0.21,  # slight decrease in friction for finer adjustment
    'armature': 0.42,          # slight decrease in armature
    'stiffness': 0.54,         # slight adjustment in stiffness
    'mass': 20.0,              # mass remains unchanged
    'damping': 7.3             # slight decrease in damping
}
sim.create_object(
    object_id=1,
    object_name='bottle',
    object_location=('row_1', 'column_3'),
    object_color='orange',
    object_physics=physical_parameters_for_object_id_1
)
physical_parameters_for_object_id_2 = { 
    'sliding-friction': 0.51,  # slight decrease in friction to further reduce error
    'armature': 0.61,          # slight decrease in armature
    'stiffness': 1.22,         # slight adjustment in stiffness
    'mass': 10.0,              # mass remains unchanged
    'damping': 8.0             # slight decrease in damping
}
```python
sim.create_object(
    object_id=2,
    object_name='martini_glass',
    object_location=('row_1', 'column_2'),
    object_color='orange',
    object_physics=physical_parameters_for_object_id_2,
)
```

physical_parameters_for_object_id_3 = { 
    'sliding-friction': 0.73,  # slight increase in friction for better alignment
    'armature': 1.03,          # slight decrease in armature
    'stiffness': 0.96,         # slight increase in stiffness
    'mass': 4.0,               # mass remains unchanged
    'damping': 7.9             # slight decrease in damping
}
sim.create_object(
    object_id=3,
    object_name='wine_glass',
    object_location=('row_1', 'column_1'),
    object_color='orange',
    object_physics=physical_parameters_for_object_id_3,
)
# Create and run the scene
sim.create_scene()
sim_out = sim.run_simulation()
del sim
```

The above adjustments are strategic, aiming for finer tuning of the sliding-friction, 
armature, stiffness, and damping parameters to get results closer to the 
desired trajectories. By changing only up to the first significant digit, 
these increments offer more precise control over the movements of the objects 
in the simulation.

You can run this refined code in your simulation environment to compare 
the actual trajectory errors and see if the refinements have reduced the prediction 
errors further.

========

To refine the parameters further, let's focus on making smaller and yet 
significant adjustments based on the previous attempts and their respective errors. 
We will perform slight tweaking to the parameters to minimize the trajectory 
errors further. 

Given that the previous attempts were gradually reducing errors, let's make slight 
fine-tuning adjustments to improve the accuracy even further:

- Slightly tweaking the sliding-friction for each object to balance the motion.
- Small adjustments to armature, stiffness, and damping to achieve optimal results.
- Keeping all changes within the first significant digit, as instructed.

Here is the adjusted version of the code that aims to further refine the 
object parameters:

```python
sim = SIMULATOR_MODEL()
sim.create_pusher('3.0 3.0 0.05')

physical_parameters_for_object_id_tray = {
    'sliding-friction': 0.1,
    'armature': 0.1,
    'stiffness': 0.0,
    'mass': 0.5,
    'damping': 20
}
sim.create_tray(object_physics=physical_parameters_for_object_id_tray)

physical_parameters_for_object_id_1 = {
    'sliding-friction': 0.22,  # slight increase in friction
    'armature': 0.41,          # slight decrease in armature
    'stiffness': 0.52,         # slight adjustment in stiffness
    'mass': 20.0,              # mass remains unchanged
    'damping': 7.1             # slight decrease in damping
}
sim.create_object(
    object_id=1,
    object_name='bottle',
    object_location=('row_1', 'column_3'),
    object_color='orange',
    object_physics=physical_parameters_for_object_id_1
)

physical_parameters_for_object_id_2 = {
    'sliding-friction': 0.53,  # slight decrease in friction
    'armature': 0.64,          # slight increase in armature
    'stiffness': 1.24,         # slight adjustment in stiffness
    'mass': 10.0,              # mass remains unchanged
    'damping': 8.1             # slight decrease in damping
}
```python
sim.create_object(
    object_id=2,
    object_name='martini_glass',
    object_location=('row_1', 'column_2'),
    object_color='orange',
    object_physics=physical_parameters_for_object_id_2,
)
```

physical_parameters_for_object_id_3 = {
    'sliding-friction': 0.72,  # slight decrease in friction
    'armature': 1.02,          # slight decrease in armature
    'stiffness': 0.93,         # slight decrease in stiffness
    'mass': 4.0,               # mass remains unchanged
    'damping': 7.8             # slight decrease in damping
}
sim.create_object(
    object_id=3,
    object_name='wine_glass',
    object_location=('row_1', 'column_1'),
    object_color='orange',
    object_physics=physical_parameters_for_object_id_3
)

sim.create_scene()
sim_out = sim.run_simulation()

del sim
```

This adjustment attempts to balance the parameters more precisely by making small 
changes while focusing on the physical behavior that impacts the simulation. The 
goal remains to accurately reproduce the trajectories provided in 
'problem_trajectories.txt'.

========

To further refine the simulation parameters and achieve lower trajectory 
prediction errors, we will use small adjustments based on the previous attempts.
Here is the updated code based on a careful analysis of the prior errors and corresponding 
changes:

```python
sim = SIMULATOR_MODEL()
sim.create_pusher('3.0 3.0 0.05')

physical_parameters_for_object_id_tray = { 
    'sliding-friction': 0.1,
    'armature': 0.1,
    'stiffness': 0.0,
    'mass': 0.5,
    'damping': 20
}
sim.create_tray(object_physics = physical_parameters_for_object_id_tray)

physical_parameters_for_object_id_1 = { 
    'sliding-friction': 0.22,  # increased slightly
    'armature': 0.40,          # decreased slightly
    'stiffness': 0.53,         # adjusted slightly
    'mass': 20.0,              # mass remains unchanged
    'damping': 7.2             # decreased slightly
}
sim.create_object(
    object_id=1,
    object_name='bottle',
    object_location=('row_1', 'column_3'),
    object_color='orange',
    object_physics=physical_parameters_for_object_id_1
)

physical_parameters_for_object_id_2 = { 
    'sliding-friction': 0.54,  # decreased slightly
    'armature': 0.63,          # adjusted slightly
    'stiffness': 1.23,         # adjusted slightly
    'mass': 10.0,              # mass remains unchanged
    'damping': 8.2             # adjusted slightly
}
sim.create_object(
    object_id=2,
    object_name='martini_glass',
    object_location=('row_1', 'column_2'),
    object_color='orange',
    object_physics=physical_parameters_for_object_id_2
)

physical_parameters_for_object_id_3 = { 
    'sliding-friction': 0.73,  # increased slightly
    'armature': 1.04,          # adjusted slightly
    'stiffness': 0.94,         # adjusted slightly
    'mass': 4.0,               # mass remains unchanged
    'damping': 8.0             # adjusted slightly
}
sim.create_object(
    object_id=3,
    object_name='wine_glass',
    object_location=('row_1', 'column_1'),
    object_color='orange',
    object_physics=physical_parameters_for_object_id_3
)

sim.create_scene()
sim_out=sim.run_simulation()
del sim
```

Filename: `example_code_refined.py`

Explanation of Changes:
1. **Bottle (object_id_1)**:
   - Increased sliding friction from 0.21 to 0.22 to further adjust movement.
   - Decreased armature from 0.42 to 0.40 for a finer balance.
   - Adjusted stiffness slightly from 0.54 to 0.53.
   - Decreased damping from 7.3 to 7.2 for finer tuning.

2. **Martini Glass (object_id_2)**:
   - Decreased sliding friction to 0.54 for slight adjustment.
   - Increased armature slightly from 0.61 to 0.63.
   - Adjusted stiffness slightly to 1.23.
   - Decreased damping to 8.2 for finer tuning.

3. **Wine Glass (object_id_3)**:
   - Increased sliding friction slightly to 0.73 for better alignment.
   - Adjusted armature to 1.04 for balance.
   - Adjusted stiffness to 0.94.
   - Kept damping at 8.0 for slight adjustment.

This should result in a better alignment of the predicted trajectories with the ones 
given in `problem_trajectories.txt`.

========

To refine the physical parameters further and reduce the trajectory prediction errors, 
let's analyze the errors from previous attempts and try to make incremental adjustments. 

Given the previous errors and the adjustments made, it is noticed that attempts have 
incrementally changed both the sliding-friction and the armature values slightly, 
while trying to balance out the trajectories of the bottle, martini_glass, and wine_glass.

Let's make further fine-tuned adjustments to these values in an effort to bring 
the prediction errors closer to the intended trajectories.

```python
sim = SIMULATOR_MODEL()
sim.create_pusher('3.0 3.0 0.05')
physical_parameters_for_object_id_tray = { 
             'sliding-friction': 0.1,
             'armature': 0.1,
             'stiffness': 0.0,
             'mass': 0.5,
             'damping': 20
         }
         
sim.create_tray(object_physics = physical_parameters_for_object_id_tray)

physical_parameters_for_object_id_1 = { 
             'sliding-friction': 0.23, # slight increase
             'armature': 0.39,         # slight decrease
             'stiffness': 0.53,        # slight adjustment
             'mass': 20.0,             # mass remains unchanged
             'damping': 7.2            # slight adjustment
         }
sim.create_object(
    object_id=1,
    object_name='bottle',
    object_location=('row_1', 'column_3'),
    object_color='orange',
    object_physics=physical_parameters_for_object_id_1
)

physical_parameters_for_object_id_2 = { 
             'sliding-friction': 0.53, # decrease slightly
             'armature': 0.66,         # increase slightly
             'stiffness': 1.23,        # adjusted slightly
             'mass': 10.0,             # mass remains unchanged
             'damping': 8.2            # slight adjustment
         }
sim.create_object(
    object_id=2,
    object_name='martini_glass',
    object_location=('row_1', 'column_2'),
    object_color='orange',
    object_physics=physical_parameters_for_object_id_2
)

physical_parameters_for_object_id_3 = { 
             'sliding-friction': 0.72, # slight decrease
             'armature': 1.05,         # slight increase
             'stiffness': 0.96,        # slight adjustment
             'mass': 4.0,              # mass remains unchanged
             'damping': 7.9            # slight adjustment
         }
sim.create_object(
    object_id=3,
    object_name='wine_glass',
    object_location=('row_1', 'column_1'),
    object_color='orange',
    object_physics=physical_parameters_for_object_id_3
)

sim.create_scene()
sim_out=sim.run_simulation()
del sim
```

This updated code aims to fine-tune the physical parameters (sliding-friction, armature, 
stiffness, and damping) based on the errors observed in previous attempts. The 
adjustments aim for minimal yet effective changes to reduce the trajectory prediction
errors. You should use this code and verify the resulting trajectory errors to see 
if they are minimized.

========

Below is a refined code structure to simulate the scene, taking into consideration 
the errors from previous attempts and aiming to minimize the overall trajectory 
prediction error. The goal is to fine-tune the physical parameters for each object 
to closely match the provided `problem_trajectories.txt`.

```python
# refined_code.py

sim = SIMULATOR_MODEL()
sim.create_pusher('3.0 3.0 0.05')

# Tray parameters remain unchanged as per the instructions
physical_parameters_for_tray = {
    'sliding-friction': 0.1,
    'armature': 0.1,
    'stiffness': 0.0,
    'mass': 0.5,
    'damping': 20
}
sim.create_tray(object_physics=physical_parameters_for_tray)

# New calculated parameters for bottle
physical_parameters_for_object_id_1 = {
    'sliding-friction': 0.22,   # slight decrease in friction for finer adjustment
    'armature': 0.42,           # slight adjustment in armature
    'stiffness': 0.54,          # slight increase in stiffness
    'mass': 20.0,               # mass remains unchanged
    'damping': 7.3              # slight decrease in damping
}
sim.create_object(object_id=1,
                  object_name='bottle',
                  object_location=('row_1', 'column_3'),
                  object_color='orange',
                  object_physics=physical_parameters_for_object_id_1)

# New calculated parameters for martini_glass
physical_parameters_for_object_id_2 = {
    'sliding-friction': 0.51,   # slight decrease in friction for reduction in error
    'armature': 0.61,           # adjusted slightly
    'stiffness': 1.21,          # adjusted slightly
    'mass': 10.0,               # mass remains unchanged
    'damping': 8.0              # adjusted slightly
}
```python
sim.create_object(
    object_id=2,
    object_name='martini_glass',
    object_location=('row_1', 'column_2'),
    object_color='orange',
    object_physics=physical_parameters_for_object_id_2,
)
```

# New calculated parameters for wine_glass
physical_parameters_for_object_id_3 = {
    'sliding-friction': 0.73,   # adjusted for better alignment
    'armature': 1.03,           # slight adjustment
    'stiffness': 0.94,          # slight increase in stiffness
    'mass': 4.0,                # mass remains unchanged
    'damping': 7.9              # slight adjustment
}
sim.create_object(
    object_id=3,
    object_name='wine_glass',
    object_location=('row_1', 'column_1'),
    object_color='orange',
    object_physics=physical_parameters_for_object_id_3
)

sim.create_scene()
sim_out = sim.run_simulation()
del sim
```

### Key Adjustments Made:
- **Bottle (ID 1)**:
  - Slight decrease in sliding friction to 0.22.
  - Adjusted armature to 0.42.
  - Increased stiffness to 0.54.
  - Decreased damping to 7.3.

- **Martini Glass (ID 2)**:
  - Decreased sliding friction to 0.51.
  - Adjusted armature to 0.61.
  - Stiffness set to 1.21.
  - Damping decreased to 8.0.

- **Wine Glass (ID 3)**:
  - Sliding friction adjusted up to 0.73.
  - Armature adjusted to 1.03.
  - Stiffness increased to 0.94.
  - Damping adjusted to 7.9.

These adjustments target reducing the overall prediction error based on the last 
feedback received. Running this simulation with these parameters should 
yield predictions closer to the proposed `problem_trajectories.txt`.

========

To further refine the physical parameters and reduce the trajectory prediction 
error, let's carefully adjust each parameter based on the insights from 
previous attempts. It's crucial to keep the adjustments minimal and focus 
on specific discrepancies between the predicted and actual trajectories to 
fine-tune the parameters more effectively.

Here is the Python code for `example_code_1.py` with updated physical parameters:

```python
sim = SIMULATOR_MODEL()
sim.create_pusher('3.0 3.0 0.05')

# Physics parameters for the tray (unchanged)
physical_parameters_for_object_id_tray = { 
    'sliding-friction': 0.1,
    'armature': 0.1,
    'stiffness': 0.0,
    'mass': 0.5,
    'damping': 20
}
sim.create_tray(object_physics = physical_parameters_for_object_id_tray)

# Adjusted physical parameters for the bottle
physical_parameters_for_object_id_1 = {
    'sliding-friction': 0.21,  # slight decrease
    'armature': 0.43,          # slight adjustment
    'stiffness': 0.53,         # slight adjustment
    'mass': 20.0,              # mass remains unchanged
    'damping': 7.2             # slight adjustment
}
sim.create_object(
    object_id=1,
    object_name='bottle',
    object_location=('row_1', 'column_3'),
    object_color='orange',
    object_physics=physical_parameters_for_object_id_1
)

# Adjusted physical parameters for the martini_glass
physical_parameters_for_object_id_2 = {
    'sliding-friction': 0.52,  # slight adjustment
    'armature': 0.64,          # slight adjustment
    'stiffness': 1.21,         # slight adjustment
    'mass': 10.0,              # mass remains unchanged
    'damping': 8.1             # slight adjustment
}
sim.create_object(
    object_id=2,
    object_name='martini_glass',
    object_location=('row_1', 'column_2'),
    object_color='orange',
    object_physics=physical_parameters_for_object_id_2
)

# Adjusted physical parameters for the wine_glass
physical_parameters_for_object_id_3 = {
    'sliding-friction': 0.73,  # slight adjustment
    'armature': 1.03,          # slight adjustment
    'stiffness': 0.93,         # slight adjustment
    'mass': 4.0,               # mass remains unchanged
    'damping': 7.8             # slight adjustment
}
sim.create_object(
    object_id=3,
    object_name='wine_glass',
    object_location=('row_1', 'column_1'),
    object_color='orange',
    object_physics=physical_parameters_for_object_id_3
)

sim.create_scene()
sim_out = sim.run_simulation()
del sim
```

Explanation of refinement:
1. **Sliding Friction**: Slightly decreased for the bottle and martini_glass based 
on their sensitivity.
2. **Armature**: Tweaked by a small degree for each object considering their 
resistance to movement.
3. **Stiffness**: Adjusted to fine-tune the object's resistance to deformation.
4. **Damping**: Decreased slightly to improve the motion accuracy of all objects.

This fine-tuned approach aims to reduce the prediction error while ensuring minimal 
changes to the parameters, focusing on enhancing the alignment of the trajectories 
with the given problem_trajectories.txt. Adjust only to their first significant 
digit to conform to the constraints.

========

Given the previous attempts and their errors, it seems that the adjustments have 
incrementally improved but not drastically reduced the errors. To reduce errors 
further, we should continue fine-tuning the parameters while considering the 
inherent dynamics of each object. Here is another attempt with refined parameters:

```python
# example_code_1.py

sim = SIMULATOR_MODEL()
sim.create_pusher('3.0 3.0 0.05')
physical_parameters_for_object_id_tray = { 
             'sliding-friction': 0.1,
             'armature': 0.1,
             'stiffness': 0.0,
             'mass': 0.5,
             'damping': 20
         }
sim.create_tray(object_physics = physical_parameters_for_object_id_tray)
physical_parameters_for_object_id_1 = { 
             'sliding-friction': 0.21,  # slight decrease for finer control
             'armature': 0.41,          # adjusted armature
             'stiffness': 0.55,         # slight increase in stiffness
             'mass': 20.0,              # mass remains unchanged
             'damping': 7.0             # slight decrease in damping
         }
sim.create_object(
    object_id=1,
    object_name='bottle',
    object_location=('row_1', 'column_3'),
    object_color='orange',
    object_physics=physical_parameters_for_object_id_1
)

physical_parameters_for_object_id_2 = { 
             'sliding-friction': 0.52,  # adjusted friction
             'armature': 0.63,          # slight adjustment in armature
             'stiffness': 1.22,         # slight adjustment in stiffness
             'mass': 10.0,              # mass remains unchanged
             'damping': 8.0             # slight increase in damping
         }
sim.create_object(
    object_id=2,
    object_name='martini_glass',
    object_location=('row_1', 'column_2'),
    object_color='orange',
    object_physics=physical_parameters_for_object_id_2,
)

physical_parameters_for_object_id_3 = { 
             'sliding-friction': 0.72,  # slight adjustment
             'armature': 1.02,          # slight decrease
             'stiffness': 0.94,         # slight adjustment in stiffness
             'mass': 4.0,               # mass remains unchanged
             'damping': 7.9             # slight adjustment
         }
sim.create_object(
    object_id=3,
    object_name='wine_glass',
    object_location=('row_1', 'column_1'),
    object_color='orange',
    object_physics=physical_parameters_for_object_id_3
)

sim.create_scene()
sim_out = sim.run_simulation()
del sim
```

### Key Adjustments:
1. **Bottle (`object_id=1`):**
   - Reduced sliding friction further to improve response.
   - Increased stiffness for better stability.
   - Slightly reduced damping to allow more movement.

2. **Martini Glass (`object_id=2`):**
   - Adjusted sliding friction and armature for more controlled dynamics.
   - Slightly increased damping to prevent overshooting.
   - Marginal adjustments in stiffness.

3. **Wine Glass (`object_id=3`):**
   - Slightly adjusted sliding friction and armature for finer tuning.
   - Minor adjustments in stiffness for better alignment.
   - Damping adjusted for controlled oscillations.

Run this updated code in your simulator and check if the error is further reduced. 
It's crucial to analyze the trajectories post-simulation to ensure each parameter 
adjustment is aiding in convergence toward the desired trajectories.

========

To optimize the code further and reduce the trajectory error, I'll base my changes on 
the closest attempts with slight adjustments to each parameter to improve the alignment 
with 'problem_trajectories.txt'.

Let's try small incremental adjustments of the successful parameters from 
GPT Attempt #11:

```python
# example_code_optimized.py

sim = SIMULATOR_MODEL()
sim.create_pusher('3.0 3.0 0.05')

# Tray parameters remain unchanged
physical_parameters_for_object_id_tray = { 
    'sliding-friction': 0.1,
    'armature': 0.1,
    'stiffness': 0.0,
    'mass': 0.5,
    'damping': 20
}
sim.create_tray(object_physics = physical_parameters_for_object_id_tray)

# Optimized parameters based on closest previous attempt with slight adjustments
physical_parameters_for_object_id_1 = {  # bottle
    'sliding-friction': 0.22,  # slight increase for better alignment
    'armature': 0.42,          # slight adjustment
    'stiffness': 0.54,         # slight adjustment
    'mass': 20.0,              # mass remains unchanged
    'damping': 7.1             # slight adjustment
}

physical_parameters_for_object_id_2 = {  # martini_glass
    'sliding-friction': 0.53,  # slight adjustment
    'armature': 0.64,          # slight adjustment
    'stiffness': 1.23,         # slight adjustment
    'mass': 10.0,              # mass remains unchanged
    'damping': 8.2             # slight adjustment
}

physical_parameters_for_object_id_3 = {  # wine_glass
    'sliding-friction': 0.73,  # slight adjustment
    'armature': 1.03,          # slight adjustment
    'stiffness': 0.95,         # slight adjustment
    'mass': 4.0,               # mass remains unchanged
    'damping': 7.8             # slight adjustment
}

sim.create_object(
    object_id=1,
    object_name='bottle',
    object_location=('row_1', 'column_3'),
    object_color='orange',
    object_physics=physical_parameters_for_object_id_1,
)
sim.create_object(
    object_id=2,
    object_name='martini_glass',
    object_location=('row_1', 'column_2'),
    object_color='orange',
    object_physics=physical_parameters_for_object_id_2,
)
sim.create_object(
    object_id=3,
    object_name='wine_glass',
    object_location=('row_1', 'column_1'),
    object_color='orange',
    object_physics=physical_parameters_for_object_id_3,
)

sim.create_scene()
sim_out = sim.run_simulation()
del sim

```

In this adjustment, I've increased or decreased parameters slightly within the first 
significant digit and maintained the structure of the initial code. This approach 
should bring the predicted trajectories closer in alignment with the ones in 
'problem_trajectories.txt'.

========

Given the existing attempts and their respective trajectory prediction errors, 
let's incrementally adjust the physical parameters to better align with 
the desired trajectories in `problem_trajectories.txt`. We'll focus on making 
small adjustments to the individual parameters to minimize the error.

### Revised Code

```python
sim = SIMULATOR_MODEL()
sim.create_pusher('3.0 3.0 0.05')

physical_parameters_for_object_id_tray = { 
    'sliding-friction': 0.1,
    'armature': 0.1,
    'stiffness': 0.0,
    'mass': 0.5,
    'damping': 20
}
sim.create_tray(object_physics=physical_parameters_for_object_id_tray)

physical_parameters_for_object_id_1 = { 
    'sliding-friction': 0.21,  # decreased friction
    'armature': 0.41,          # slightly decreased armature
    'stiffness': 0.54,         # slight increase in stiffness
    'mass': 20.0,              # mass remains unchanged
    'damping': 7.2             # slightly decreased damping
}
sim.create_object(
    object_id=1,
    object_name='bottle',
    object_location=('row_1', 'column_3'),
    object_color='orange',
    object_physics=physical_parameters_for_object_id_1
)

physical_parameters_for_object_id_2 = { 
    'sliding-friction': 0.52,  # slightly decreased friction
    'armature': 0.62,          # slightly decreased armature
    'stiffness': 1.24,         # slight increase in stiffness
    'mass': 10.0,              # mass remains unchanged
    'damping': 8.0             # slight increase in damping
}
sim.create_object(
    object_id=2,
    object_name='martini_glass',
    object_location=('row_1', 'column_2'),
    object_color='orange',
    object_physics=physical_parameters_for_object_id_2
)

physical_parameters_for_object_id_3 = { 
    'sliding-friction': 0.73,  # slight increase in friction
    'armature': 1.02,          # slightly decreased armature
    'stiffness': 0.95,         # slight increase in stiffness
    'mass': 4.0,               # mass remains unchanged
    'damping': 7.9             # slight increase in damping
}
sim.create_object(
    object_id=3,
    object_name='wine_glass',
    object_location=('row_1', 'column_1'),
    object_color='orange',
    object_physics=physical_parameters_for_object_id_3
)

sim.create_scene()
sim_out=sim.run_simulation()
del sim
```

The changes applied here are small and precise:

1. For the `bottle`:
    - Minor decrease in `sliding-friction` to possibly tune finer control over motion.
    - Slight decrease in `armature` for more conservative moment inertia.
    - Small increase in `stiffness` to ensure some adjustment in impact response.
    - Slight decrease in `damping` to adjust energy dissipation.

2. For the `martini_glass`:
    - Slight decrease in `sliding-friction` to reduce resistance.
    - Slight decrease in `armature` to have fewer moments of inertia.
    - Small increase in `stiffness` to adjust the rigidity.
    - Slight increase in `damping` to tune the damping effect.

3. For the `wine_glass`:
    - Slight increase in `sliding-friction` to increase motion resistance 
    towards alignment.
    - Slight decrease in `armature` for finer inertia control.
    - Small increase in `stiffness` for more precise impact response.
    - Slight increase in `damping` for tuned energy dissipation.

By maintaining the careful balance between these suggested adjustments, 
we are likely to achieve a closer match to the desired trajectories in 
`problem_trajectories.txt`. The actual prediction should be evaluated by 
running this updated code in the provided simulator and examining the resulting 
trajectory errors.

========

Based on the trajectory prediction errors from previous attempts, it 
appears that the parameters for the `bottle` are quite close to 
accurate, whereas the `martini_glass` exhibits higher errors which need 
to be reduced. The `wine_glass` parameters seem to be nearly accurate.

To further reduce the trajectory prediction error, we will optimize the 
parameters while maintaining significant digits:

1. Slightly adjust the `sliding-friction` and `damping` for all objects.
2. Fine-tune `armature` and `stiffness` precisely for each object to better 
align with the desired trajectories.

Here is a suggested set of parameters for another attempt:

```python
sim = SIMULATOR_MODEL()
sim.create_pusher('3.0 3.0 0.05')
physical_parameters_for_object_id_tray = { 
             'sliding-friction': 0.1,
             'armature': 0.1,
             'stiffness': 0.0,
             'mass': 0.5,
             'damping': 20
         }
sim.create_tray(object_physics = physical_parameters_for_object_id_tray)

physical_parameters_for_object_id_1 = {
    'sliding-friction': 0.21,  # slight adjustment in friction
    'armature': 0.42,          # slight decrease in armature
    'stiffness': 0.53,         # slight decrease in stiffness
    'mass': 20.0,              # mass remains unchanged
    'damping': 7.3             # slight decrease in damping
}

physical_parameters_for_object_id_2 = {
    'sliding-friction': 0.54,  # slight adjustment to decrease friction
    'armature': 0.63,          # slight adjustment
    'stiffness': 1.24,         # slight adjustment
    'mass': 10.0,              # mass remains unchanged
    'damping': 8.1             # slight adjustment
}

physical_parameters_for_object_id_3 = {
    'sliding-friction': 0.74,  # slight increase to better align friction
    'armature': 1.01,          # slight adjustment
    'stiffness': 0.94,         # slight decrease in stiffness
    'mass': 4.0,               # mass remains unchanged
    'damping': 7.8             # slight decrease in damping
}

sim.create_object(object_id=1, object_name="bottle",
                  object_location=("row_1", "column_3"),
                  object_color="orange",
                  object_physics=physical_parameters_for_object_id_1)

sim.create_object(object_id=2, object_name="martini_glass",
                  object_location=("row_1", "column_2"),
                  object_color="orange",
                  object_physics=physical_parameters_for_object_id_2)

sim.create_object(object_id=3, object_name="wine_glass",
                  object_location=("row_1", "column_1"),
                  object_color="orange",
                  object_physics=physical_parameters_for_object_id_3)

sim.create_scene()
sim_out=sim.run_simulation()
del sim
```

This adjustment continues to follow the structure of the example Python code, 
and the changes made to the parameters are conservative to prevent large 
deviations while aiming to reduce the overall trajectory error. This should 
produce a better alignment of the trajectories to the given data in 
`problem_trajectories.txt`.

========

To better align the trajectories with those in 'problem_trajectories.txt', 
I will refine the physical parameters for the objects. Based on previous 
attempts and trajectory errors, I will focus on minor adjustments, especially 
honing in on the slight variations in friction, armature, stiffness, and 
damping.

Here's the revised code following the given structure:

```python
sim = SIMULATOR_MODEL()
sim.create_pusher('3.0 3.0 0.05')
physical_parameters_for_object_id_tray = { 
             'sliding-friction': 0.1,
             'armature': 0.1,
             'stiffness': 0.0,
             'mass': 0.5,
             'damping': 20
         }
sim.create_tray(object_physics = physical_parameters_for_object_id_tray)

physical_parameters_for_object_id_1 = {  # bottle
    'sliding-friction': 0.22,  # slight adjustment
    'armature': 0.42,          # fine-tune armature
    'stiffness': 0.54,         # fine-tune stiffness
    'mass': 20.0,              # mass remains unchanged
    'damping': 7.1             # minor adjustment
}
sim.create_object(
    object_id=1,
    object_name='bottle',
    object_location=('row_1', 'column_3'),
    object_color='orange',
    object_physics=physical_parameters_for_object_id_1
)

physical_parameters_for_object_id_2 = {  # martini_glass
    'sliding-friction': 0.52,  # slight adjustment
    'armature': 0.63,          # fine-tune armature
    'stiffness': 1.23,         # fine-tune stiffness
    'mass': 10.0,              # mass remains unchanged
    'damping': 8.1             # minor adjustment
}
sim.create_object(
    object_id=2,
    object_name='martini_glass',
    object_location=('row_1', 'column_2'),
    object_color='orange',
    object_physics=physical_parameters_for_object_id_2
)
physical_parameters_for_object_id_3 = {  # wine_glass
    'sliding-friction': 0.73,  # slight adjustment
    'armature': 1.04,          # fine-tune armature
    'stiffness': 0.95,         # fine-tune stiffness
    'mass': 4.0,               # mass remains unchanged
    'damping': 7.8             # minor adjustment
}
sim.create_object(
    object_id=3,
    object_name='wine_glass',
    object_location=('row_1', 'column_1'),
    object_color='orange',
    object_physics=physical_parameters_for_object_id_3
)

sim.create_scene()
sim_out = sim.run_simulation()
del sim
```

Explanation of changes:
- For the bottle, martini_glass, and wine_glass, slight adjustments are made to 
the sliding-friction, armature, and damping parameters in very small increments 
to better mimic the trajectories.
- The fine-tuning is based on a better alignment seen in the errors of 
previous attempts.

Please run this refined code in your simulator to evaluate the accuracy of 
the trajectories produced.

========

Based on the previous attempts and their resulting errors, it appears that 
fine-tuning the physical parameters can potentially bring the trajectory 
prediction error closer to the desired trajectories. Given that previous 
attempts were close but required further refinement, let's slightly adjust the 
parameters once more to seek better alignment.

Here is the refined attempt:

```python
sim = SIMULATOR_MODEL()
sim.create_pusher('3.0 3.0 0.05')

physical_parameters_for_object_id_tray = { 
    'sliding-friction': 0.1,
    'armature': 0.1,
    'stiffness': 0.0,
    'mass': 0.5,
    'damping': 20
}
sim.create_tray(object_physics=physical_parameters_for_object_id_tray)

physical_parameters_for_object_id_1 = {  # bottle
    'sliding-friction': 0.21,  # slight adjustment
    'armature': 0.40,          # fine-tune armature
    'stiffness': 0.54,         # fine-tune stiffness
    'mass': 20.0,              # mass remains unchanged
    'damping': 7.2             # minor adjustment
}

physical_parameters_for_object_id_2 = {  # martini_glass
    'sliding-friction': 0.52,  # slight adjustment
    'armature': 0.64,          # fine-tune armature
    'stiffness': 1.23,         # fine-tune stiffness
    'mass': 10.0,              # mass remains unchanged
    'damping': 8.1             # minor adjustment
}

physical_parameters_for_object_id_3 = {  # wine_glass
    'sliding-friction': 0.73,  # slight adjustment
    'armature': 1.04,          # fine-tune armature
    'stiffness': 0.94,         # fine-tune stiffness
    'mass': 4.0,               # mass remains unchanged
    'damping': 7.9             # minor adjustment
}

sim.create_object(object_id=1, object_name='bottle',
                   object_location=('row_1', 'column_3'),
                   object_color='orange',
                   object_physics=physical_parameters_for_object_id_1)
sim.create_object(object_id=2, object_name='martini_glass',
                   object_location=('row_1', 'column_2'),
                   object_color='orange',
                   object_physics=physical_parameters_for_object_id_2)
sim.create_object(object_id=3, object_name='wine_glass',
                   object_location=('row_1', 'column_1'),
                   object_color='orange',
                   object_physics=physical_parameters_for_object_id_3)

sim.create_scene()
sim_out = sim.run_simulation()
del sim
```

This code has been adjusted based on the trend observed in previous attempts. Small 
refinements are made to the parameters to attempt a better match to the target 
trajectories:

1. Slight adjustments in sliding friction, armature, stiffness, and damping for each 
object.
2. Ensured to keep all changes within the realm of minor adjustments, considering 
past errors.

These changes are aimed to further fine-tune the physics parameters closer to the 
desired trajectories. 
This should help in reducing the trajectory prediction error further.

========

It seems like all the example attempts produced code that follows the original
problem's constraints and shows some progress in reducing the trajectory
error. However, there is still room for improvement in reducing the
trajectory prediction error further. Let's adjust the parameters again
slightly, being mindful of not changing them beyond the first significant
digit, and pick a set that should help us achieve better accuracy:

Here's an updated set of parameters for the objects:

```python
sim = SIMULATOR_MODEL()
sim.create_pusher('3.0 3.0 0.05')
physical_parameters_for_object_id_tray = {
    'sliding-friction': 0.1,
    'armature': 0.1,
    'stiffness': 0.0,
    'mass': 0.5,
    'damping': 20
}
sim.create_tray(object_physics=physical_parameters_for_object_id_tray)

physical_parameters_for_object_id_1 = {  # bottle
    'sliding-friction': 0.21,  # slight adjustment in friction
    'armature': 0.40,          # slight decrease in armature
    'stiffness': 0.55,         # slight increase in stiffness
    'mass': 20.0,              # mass remains unchanged
    'damping': 7.2             # slight decrease in damping
}

physical_parameters_for_object_id_2 = {  # martini_glass
    'sliding-friction': 0.53,  # slight adjustment to decrease friction
    'armature': 0.63,          # slight adjustment in armature
    'stiffness': 1.22,         # slight increase in stiffness
    'mass': 10.0,              # mass remains unchanged
    'damping': 8.2             # slight decrease in damping
}

physical_parameters_for_object_id_3 = {  # wine_glass
    'sliding-friction': 0.72,  # slight adjustment to increase traction
    'armature': 1.04,          # slight decrease
    'stiffness': 0.96,         # slight adjustment in stiffness
    'mass': 4.0,               # mass remains unchanged
    'damping': 8.0             # slight decrease in damping
}

sim.create_object(
    object_id=1,
    object_name='bottle',
    object_location=('row_1', 'column_3'),
    object_color='orange',
    object_physics=physical_parameters_for_object_id_1,
)
sim.create_object(
    object_id=2,
    object_name='martini_glass',
    object_location=('row_1', 'column_2'),
    object_color='orange',
    object_physics=physical_parameters_for_object_id_2,
)
sim.create_object(
    object_id=3,
    object_name='wine_glass',
    object_location=('row_1', 'column_1'),
    object_color='orange',
    object_physics=physical_parameters_for_object_id_3,
)

sim.create_scene()
sim_out = sim.run_simulation()
del sim
```

In this iteration:
- We fine-tuned the `sliding-friction`, `armature`, `stiffness`, and `damping` 
values for each object, 
but ensured they remained within their first significant digit constraints.
- We used previous attempts' values as a base to make minute adjustments that 
could better align the 
trajectories with the given problem.

Once this new code set is tested, it should ideally show an improvement in 
error reduction. Please run this updated code using your simulator environment 
to verify the trajectory errors. If further 
adjustment is needed, making iterative small changes around these values can 
help further fine-tune 
and minimize errors.

========

To refine your code based on the trajectory prediction error results and better 
align with the trajectories given in 'problem_trajectories.txt', we need to 
follow a reasonable approach to adjust the physical parameters for each object 
slightly further. Below is a refined attempt based on the provided feedback:

### Refined Physical Parameters Code:
```python
# example_code_1.py

sim = SIMULATOR_MODEL()
sim.create_pusher('3.0 3.0 0.05')
physical_parameters_for_object_id_tray = { 
             'sliding-friction': 0.1,
             'armature': 0.1,
             'stiffness': 0.0,
             'mass': 0.5,
             'damping': 20
         }
sim.create_tray(object_physics = physical_parameters_for_object_id_tray)
physical_parameters_for_object_id_1 = {  # bottle
    'sliding-friction': 0.22,  # slight increase for better alignment
    'armature': 0.41,          # fine-tune armature
    'stiffness': 0.55,         # fine-tune stiffness a bit more
    'mass': 20.0,              # mass remains unchanged
    'damping': 7.1             # slight adjustment
}
sim.create_object(
    object_id=1,
    object_name='bottle',
    object_location=('row_1', 'column_3'),
    object_color='orange',
    object_physics=physical_parameters_for_object_id_1,
)

physical_parameters_for_object_id_2 = {  # martini_glass
    'sliding-friction': 0.52,  # slight decrease for alignment
    'armature': 0.62,          # slight fine-tune
    'stiffness': 1.21,         # slight fine-tune
    'mass': 10.0,              # mass remains unchanged
    'damping': 8.1             # slight fine-tune
}
```python
sim.create_object(
    object_id=2,
    object_name="martini_glass",
    object_location=("row_1", "column_2"),
    object_color="orange",
    object_physics=physical_parameters_for_object_id_2,
)
```

physical_parameters_for_object_id_3 = {  # wine_glass
    'sliding-friction': 0.73,  # increase slightly for better traction
    'armature': 1.03,          # adjusted slightly
    'stiffness': 0.95,         # fine-tune to keep closer to expected
    'mass': 4.0,               # mass remains unchanged
    'damping': 8.0             # fine-tune
}
sim.create_object(
    object_id=3,
    object_name='wine_glass',
    object_location=('row_1', 'column_1'),
    object_color='orange',
    object_physics=physical_parameters_for_object_id_3
)

demonstration_tray_motion_trajectory = [
    (0.0, 0.0, 0.1), (-0.7, -0.7, 0.1), (-1.2, -1.2, 0.1), (-1.6, -1.6, 0.1), 
    (-1.9, -2.0, 0.1), (-2.1, -2.3, 0.1), (-2.3, -2.5, 0.1), (-2.4, -2.6, 0.1), 
    (-2.6, -2.8, 0.1)
]

sim.create_scene()
sim_out = sim.run_simulation()
del sim
```

This refinement involves slight increments and decrements on the physical parameters 
of the objects based on the significant aberrations observed in the previous outputs. 
The adjustments focus on enhancing the sliding friction, armature, and damping more 
accurately towards the expected trajectories per the 'problem_trajectories.txt'. 

Ensure that you test this code using the same simulator constraints you have been 
using. If further fine-tuning is necessary, keep iterating with slight adjustments, 
focusing on realigning the most anomalous trajectory points first.

========

Given the simulation errors and the iterations you provided, it looks like adjustments to 
`sliding-friction`, `armature`, `stiffness`, and `damping` were evaluated. The challenge 
is to find a balance among these parameters to minimize the total trajectory error. Below, 
I've made refined adjustments based on the observed errors in previous attempts:

```python
sim = SIMULATOR_MODEL()
sim.create_pusher('3.0 3.0 0.05')

physical_parameters_for_object_id_tray = { 
             'sliding-friction': 0.1,
             'armature': 0.1,
             'stiffness': 0.0,
             'mass': 0.5,
             'damping': 20
         }
sim.create_tray(object_physics = physical_parameters_for_object_id_tray)

physical_parameters_for_object_id_1 = {  # bottle
    'sliding-friction': 0.22,  # slight alignment adjustment
    'armature': 0.42,          # fine-tune armature
    'stiffness': 0.54,         # fine-tune stiffness
    'mass': 20.0,              # mass remains unchanged
    'damping': 7.2             # slight fine-tune
}

physical_parameters_for_object_id_2 = {  # martini_glass
    'sliding-friction': 0.52,  # slight increase
    'armature': 0.63,          # slight adjustment
    'stiffness': 1.22,         # slight fine-tune
    'mass': 10.0,              # mass remains unchanged
    'damping': 8.1             # fine-tune damping
}

physical_parameters_for_object_id_3 = {  # wine_glass
    'sliding-friction': 0.73,  # increase slightly
    'armature': 1.03,          # slight adjustment
    'stiffness': 0.95,         # fine-tune stiffness
    'mass': 4.0,               # mass remains unchanged
    'damping': 8.0             # fine-tune damping
}

sim.create_object(object_id=1, object_name="bottle",
                  object_location=("row_1", "column_3"),
                  object_color="orange",
                  object_physics=physical_parameters_for_object_id_1)

sim.create_object(object_id=2, object_name="martini_glass",
                  object_location=("row_1", "column_2"),
                  object_color="orange",
                  object_physics=physical_parameters_for_object_id_2)

sim.create_object(object_id=3, object_name="wine_glass",
                  object_location=("row_1", "column_1"),
                  object_color="orange",
                  object_physics=physical_parameters_for_object_id_3)

sim.create_scene()
sim_out = sim.run_simulation()
del sim
```

These refined adjustments should help get closer to the trajectory provided in 
`problem_trajectories.txt`:

- Slightly increased `sliding-friction` and `damping` for all objects to 
match their interaction with the tray more closely.
- Fine-tuned `armature` and `stiffness` to slightly alter responses 
without overdoing adjustments.

Please test this configuration in your simulation environment to check 
for reduced trajectory prediction errors. Fine-tuning these configurations 
iteratively based on errors should help achieve the desired result.

========

Thank you for providing the detailed analysis and results. Based on the 
previous attempts, it appears that the adjustments to the sliding-friction, 
armature, stiffness, and damping parameters need further fine-tuning to achieve 
a closer match with the trajectories in 'problem_trajectories.txt'. I will 
make slight adjustments to these parameters to try to minimize the trajectory 
prediction error further.

**Adjusted Physical Parameters:**

```python
# refined_code.py

sim = SIMULATOR_MODEL()
sim.create_pusher('3.0 3.0 0.05')

physical_parameters_for_object_id_tray = {
    'sliding-friction': 0.1,
    'armature': 0.1,
    'stiffness': 0.0,
    'mass': 0.5,
    'damping': 20
}
sim.create_tray(object_physics=physical_parameters_for_object_id_tray)

physical_parameters_for_object_id_1 = {  # bottle
    'sliding-friction': 0.22,  # slight adjustment for better alignment
    'armature': 0.41,          # fine-tuning armature
    'stiffness': 0.54,         # fine-tuning stiffness
    'mass': 20.0,              # mass remains unchanged
    'damping': 7.3             # fine-tuning damping
}

physical_parameters_for_object_id_2 = {  # martini_glass
    'sliding-friction': 0.52,  # fine-tuning sliding-friction
    'armature': 0.63,          # fine-tuning armature
    'stiffness': 1.23,         # fine-tuning stiffness
    'mass': 10.0,              # mass remains unchanged
    'damping': 8.0             # fine-tuning damping
}

physical_parameters_for_object_id_3 = {  # wine_glass
    'sliding-friction': 0.72,  # fine-tuning sliding-friction
    'armature': 1.04,          # slight adjustment
    'stiffness': 0.94,         # fine-tuning stiffness
    'mass': 4.0,               # mass remains unchanged
    'damping': 8.0             # slight fine-tuning damping
}

sim.create_object(
    object_id=1,
    object_name='bottle',
    object_location=('row_1', 'column_3'),
    object_color='orange',
    object_physics=physical_parameters_for_object_id_1,
)
sim.create_object(
    object_id=2,
    object_name='martini_glass',
    object_location=('row_1', 'column_2'),
    object_color='orange',
    object_physics=physical_parameters_for_object_id_2,
)
sim.create_object(
    object_id=3,
    object_name='wine_glass',
    object_location=('row_1', 'column_1'),
    object_color='orange',
    object_physics=physical_parameters_for_object_id_3,
)

sim.create_scene()
sim_out = sim.run_simulation()
del sim
```

This refined code makes slight adjustments to the physical parameters, considering 
the previous results. The aim is to better align the simulated trajectories with 
the problem trajectories given in 'problem_trajectories.txt'. 

Let's run these refined parameters through the simulator and compare the 
generated trajectories with the reference ones. 

Continue this iterative process, making small adjustments within the 
significant digits, to minimize the error further.

========

Let's iteratively fine-tune the physical parameters for each object using the 
minimal changes approach. As each previous attempt had varied errors, I'll 
aim to adjust the parameters finely to achieve a closer alignment with the target 
trajectories.

Here is a refined attempt following the same code structure:

```python
# Generating refined code based on the provided feedback and previous attempts

# Initializing the simulator
sim = SIMULATOR_MODEL()
sim.create_pusher('3.0 3.0 0.05')

# Setting up the tray with the same parameters
physical_parameters_for_object_id_tray = { 
             'sliding-friction': 0.1,
             'armature': 0.1,
             'stiffness': 0.0,
             'mass': 0.5,
             'damping': 20
         }
sim.create_tray(object_physics = physical_parameters_for_object_id_tray)

# Setting up physical parameters for each object
physical_parameters_for_object_id_1 = {  # bottle, trying finer adjustments
    'sliding-friction': 0.22, 
    'armature': 0.41,         
    'stiffness': 0.54,        
    'mass': 20.0,             
    'damping': 7.2            
}

physical_parameters_for_object_id_2 = {  # martini_glass, slight adjustments
    'sliding-friction': 0.53, 
    'armature': 0.63,         
    'stiffness': 1.23,        
    'mass': 10.0,             
    'damping': 8.0            
}

physical_parameters_for_object_id_3 = {  # wine_glass, finer tuning
    'sliding-friction': 0.73, 
    'armature': 1.04,         
    'stiffness': 0.95,        
    'mass': 4.0,              
    'damping': 7.9            
}

# Creating objects on the tray
sim.create_object(
    object_id=1,
    object_name='bottle',
    object_location=('row_1', 'column_3'),
    object_color='orange',
    object_physics=physical_parameters_for_object_id_1,
)
sim.create_object(
    object_id=2,
    object_name='martini_glass',
    object_location=('row_1', 'column_2'),
    object_color='orange',
    object_physics=physical_parameters_for_object_id_2,
)
sim.create_object(
    object_id=3,
    object_name='wine_glass',
    object_location=('row_1', 'column_1'),
    object_color='orange',
    object_physics=physical_parameters_for_object_id_3,
)

# Creating the scene and running the simulation
sim.create_scene()
sim_out = sim.run_simulation()
del sim
```

The rationale behind these values is as follows:
- For `bottle`, the parameters were slightly fine-tuned to achieve balanced damping 
and appropriate friction for better alignment.
- For `martini_glass`, adjustments were made to sliding friction and damping 
to fine-tune the alignment.
- For `wine_glass`, tuning involved slight adjustments to friction, stiffness, 
and armature to refine the trajectory.

Let's run the simulation with these parameters to evaluate the trajectory 
prediction errors.
\end{lstlisting}

\section{LLMPhy Optimization and Interaction Trace (Phase2)}
\label{sec:example_p2_pgm}
Similar to above, we present an example optimization steps from a phase 2 $\llmphy$ optimization.

\begin{lstlisting}
    Sure, here's the code for creating the provided scene based on 
    'task_image_top_view_1.png' and 'task_image_side_view_2.png'. The object classes and 
    colors are arranged as follows, 
    based on the interpretation of the images.

### Arrangement of Objects on the 3x3 Grid:
- (1,1): Yellow (Bottle)
- (1,2): Cyan (Bottle)
- (1,3): Orange (Bottle)
- (2,1): Brown (Martini Glass)
- (2,2): Blue (Martini Glass)
- (2,3): Pink (Martini Glass)
- (3,1): Purple (Wine Glass)
- (3,2): None
- (3,3): Green (Wine Glass)

Here's the code to create the scene using the specified physical attributes:

```python
sim = SIMULATOR_MODEL()
sim.create_pusher('3.0 3.0 0.05')
physical_parameters_for_object_id_tray = { 
    'sliding-friction': 0.1,
    'armature': 0.1,
    'stiffness': 0.0,
    'mass': 0.5,
    'damping': 20
}
sim.create_tray(object_physics = physical_parameters_for_object_id_tray)

# Physical parameters for bottle
physical_parameters_for_bottle = { 
    'sliding-friction': 0.21,
    'armature': 0.41,
    'stiffness': 0.55,
    'mass': 20.0,
    'damping': 7.0
}

# Physical parameters for martini glass
physical_parameters_for_martini_glass = { 
    'sliding-friction': 0.52,
    'armature': 0.63,
    'stiffness': 1.22,
    'mass': 10.0,
    'damping': 8.0
}

# Physical parameters for wine glass
physical_parameters_for_wine_glass = { 
    'sliding-friction': 0.72,
    'armature': 1.02,
    'stiffness': 0.94,
    'mass': 4.0,
    'damping': 7.9
}

sim.create_object(
    object_id=1,
    object_name='bottle',
    object_location=('row_1', 'column_1'),
    object_color='yellow',
    object_physics=physical_parameters_for_bottle,
)
sim.create_object(
    object_id=2,
    object_name='bottle',
    object_location=('row_1', 'column_2'),
    object_color='cyan',
    object_physics=physical_parameters_for_bottle,
)
sim.create_object(
    object_id=3,
    object_name='bottle',
    object_location=('row_1', 'column_3'),
    object_color='orange',
    object_physics=physical_parameters_for_bottle,
)
sim.create_object(
    object_id=4,
    object_name='martini_glass',
    object_location=('row_2', 'column_1'),
    object_color='brown',
    object_physics=physical_parameters_for_martini_glass,
)
sim.create_object(
    object_id=5,
    object_name='martini_glass',
    object_location=('row_2', 'column_2'),
    object_color='blue',
    object_physics=physical_parameters_for_martini_glass,
)
sim.create_object(
    object_id=6,
    object_name='martini_glass',
    object_location=('row_2', 'column_3'),
    object_color='pink',
    object_physics=physical_parameters_for_martini_glass,
)
sim.create_object(
    object_id=7,
    object_name='wine_glass',
    object_location=('row_3', 'column_1'),
    object_color='purple',
    object_physics=physical_parameters_for_wine_glass,
)
sim.create_object(
    object_id=8,
    object_name='wine_glass',
    object_location=('row_3', 'column_3'),
    object_color='green',
    object_physics=physical_parameters_for_wine_glass,
)

sim.create_scene()
sim_out=sim.run_simulation()
del sim
```

This code will recreate the scene as depicted in the provided images 
'task_image_top_view_1.png' and 'task_image_side_view_2.png' using the 
specific physical parameters given.

========

Certainly! Based on the feedback from the previous attempt and a careful review of 
the top-down view `task_image_top_view_1.png`, here is a refined version of the 
code to correctly match the object positions and colors.

```python
sim = SIMULATOR_MODEL()
sim.create_pusher('3.0 3.0 0.05')
physical_parameters_for_object_id_tray = { 
    'sliding-friction': 0.1,
    'armature': 0.1,
    'stiffness': 0.0,
    'mass': 0.5,
    'damping': 20
}
sim.create_tray(object_physics = physical_parameters_for_object_id_tray)

# Physical parameters for bottle
physical_parameters_for_bottle = { 
    'sliding-friction': 0.21,
    'armature': 0.41,
    'stiffness': 0.55,
    'mass': 20.0,
    'damping': 7.0
}

# Physical parameters for martini glass
physical_parameters_for_martini_glass = { 
    'sliding-friction': 0.52,
    'armature': 0.63,
    'stiffness': 1.22,
    'mass': 10.0,
    'damping': 8.0
}

# Physical parameters for wine glass
physical_parameters_for_wine_glass = { 
    'sliding-friction': 0.72,
    'armature': 1.02,
    'stiffness': 0.94,
    'mass': 4.0,
    'damping': 7.9
}

# Object placements based on correct interpretation
sim.create_object(
    object_id=1,
    object_name='bottle',
    object_location=('row_1', 'column_1'),
    object_color='yellow',
    object_physics=physical_parameters_for_bottle,
)
sim.create_object(
    object_id=2,
    object_name='bottle',
    object_location=('row_1', 'column_2'),
    object_color='cyan',
    object_physics=physical_parameters_for_bottle,
)
sim.create_object(
    object_id=3,
    object_name='bottle',
    object_location=('row_1', 'column_3'),
    object_color='orange',
    object_physics=physical_parameters_for_bottle,
)
sim.create_object(
    object_id=4,
    object_name='martini_glass',
    object_location=('row_2', 'column_1'),
    object_color='brown',
    object_physics=physical_parameters_for_martini_glass,
)
sim.create_object(
    object_id=5,
    object_name='wine_glass',
    object_location=('row_2', 'column_2'),
    object_color='blue',
    object_physics=physical_parameters_for_martini_glass,
)
sim.create_object(
    object_id=6,
    object_name='martini_glass',
    object_location=('row_2', 'column_3'),
    object_color='pink',
    object_physics=physical_parameters_for_martini_glass,
)
sim.create_object(
    object_id=7,
    object_name='wine_glass',
    object_location=('row_3', 'column_1'),
    object_color='purple',
    object_physics=physical_parameters_for_wine_glass,
)
sim.create_object(
    object_id=8,
    object_name='wine_glass',
    object_location=('row_3', 'column_2'),
    object_color='brown',
    object_physics=physical_parameters_for_wine_glass,
)
sim.create_object(
    object_id=9,
    object_name='wine_glass',
    object_location=('row_3', 'column_3'),
    object_color='green',
    object_physics=physical_parameters_for_wine_glass,
)

sim.create_scene()
sim_out=sim.run_simulation()
del sim
```

Steps taken:
- Identified that the issue with the object colors in previous Attempt #0.
  - 'cyan' should be 'blue'.
  - 'orange' should be 'purple'.
  - 'pink' should be 'brown'.
  - 'blue' should be 'cyan'.
- Corrected these placements and colors to match the top-down view 
`task_image_top_view_1.png`.

Please run this corrected code in your simulator to check if it matches 
the desired output.

========

Certainly! To reproduce the task images correctly, we need to ensure that each 
object is placed in the correct grid location and given the correct color. Let's 
carefully inspect the top-down image and the side-view image again:

From `task_image_top_view_1.png`:
1. Top-left (row 1, column 1): Yellow object
2. Top-middle (row 1, column 2): Cyan object
3. Top-right (row 1, column 3): Orange object
4. Middle-left (row 2, column 1): Brown object
5. Middle-middle (row 2, column 2): Blue object
6. Middle-right (row 2, column 3): Pink object
7. Bottom-left (row 3, column 1): Purple object
8. Bottom-middle (row 3, column 2): No object
9. Bottom-right (row 3, column 3): Green object

From `task_image_side_view_2.png`, identify the types of objects:
1. Bottles have a slender, tall shape.
2. Martini glasses have a distinctive cone shape.
3. Wine glasses have a bowl-like shape and are shorter than bottles.

Using these identifications and physical attributes provided, we will 
correct the mistake identified in the attempt. Below is the refined version 
of the Python code to correctly reproduce the top-down and side-view images 
of the scene.

```python
sim = SIMULATOR_MODEL()
sim.create_pusher('3.0 3.0 0.05')

physical_parameters_for_object_id_tray = { 
    'sliding-friction': 0.1,
    'armature': 0.1,
    'stiffness': 0.0,
    'mass': 0.5,
    'damping': 20
}
sim.create_tray(object_physics = physical_parameters_for_object_id_tray)

# Physical parameters for bottle
physical_parameters_for_bottle = { 
    'sliding-friction': 0.21,
    'armature': 0.41,
    'stiffness': 0.55,
    'mass': 20.0,
    'damping': 7.0
}

# Physical parameters for martini glass
physical_parameters_for_martini_glass = { 
    'sliding-friction': 0.52,
    'armature': 0.63,
    'stiffness': 1.22,
    'mass': 10.0,
    'damping': 8.0
}

# Physical parameters for wine glass
physical_parameters_for_wine_glass = { 
    'sliding-friction': 0.72,
    'armature': 1.02,
    'stiffness': 0.94,
    'mass': 4.0,
    'damping': 7.9
}

# Correct object placements and colors based on the images
sim.create_object(object_id=1, object_name='bottle',
                  object_location=('row_1', 'column_1'),
                  object_color='yellow',
                  object_physics=physical_parameters_for_bottle)
sim.create_object(object_id=2, object_name='bottle',
                  object_location=('row_1', 'column_2'),
                  object_color='cyan',
                  object_physics=physical_parameters_for_bottle)
sim.create_object(object_id=3, object_name='bottle',
                  object_location=('row_1', 'column_3'),
                  object_color='orange',
                  object_physics=physical_parameters_for_bottle)
sim.create_object(object_id=4, object_name='martini_glass',
                  object_location=('row_2', 'column_1'),
                  object_color='brown',
                  object_physics=physical_parameters_for_martini_glass)
sim.create_object(object_id=5, object_name='bottle',
                  object_location=('row_2', 'column_2'),
                  object_color='blue',
                  object_physics=physical_parameters_for_bottle)
sim.create_object(object_id=6, object_name='martini_glass',
                  object_location=('row_2', 'column_3'),
                  object_color='pink',
                  object_physics=physical_parameters_for_martini_glass)
sim.create_object(object_id=7, object_name='martini_glass',
                  object_location=('row_3', 'column_1'),
                  object_color='purple',
                  object_physics=physical_parameters_for_martini_glass)
sim.create_object(object_id=8, object_name='wine_glass',
                  object_location=('row_3', 'column_3'),
                  object_color='green',
                  object_physics=physical_parameters_for_wine_glass)

sim.create_scene()
sim_out=sim.run_simulation()
del sim
```

This code ensures that each object is assigned the correct position 
and color as per the task images. The corrected physical parameters 
will also contribute to a more accurate simulation output.

========

To correctly reproduce the scene shown in the "task_image_top_view_1.png" and 
"task_image_side_view_2.png" images, we need to ensure that the placement, 
object types, and colors are all accurately represented. From GPT Attempt #0, 
we've identified that there are issues with the colors of the objects placed 
in the scene.

Let's refine the code to align the objects' colors and positions accurately 
according to the provided images.

Here's the corrected code snippet:

```python
sim = SIMULATOR_MODEL()
sim.create_pusher('3.0 3.0 0.05')

physical_parameters_for_object_id_tray = { 
    'sliding-friction': 0.1,
    'armature': 0.1,
    'stiffness': 0.0,
    'mass': 0.5,
    'damping': 20
}
sim.create_tray(object_physics = physical_parameters_for_object_id_tray)

# Physical parameters for bottle
physical_parameters_for_bottle = { 
    'sliding-friction': 0.21,
    'armature': 0.41,
    'stiffness': 0.55,
    'mass': 20.0,
    'damping': 7.0
}

# Physical parameters for martini glass
physical_parameters_for_martini_glass = { 
    'sliding-friction': 0.52,
    'armature': 0.63,
    'stiffness': 1.22,
    'mass': 10.0,
    'damping': 8.0
}

# Physical parameters for wine glass
physical_parameters_for_wine_glass = { 
    'sliding-friction': 0.72,
    'armature': 1.02,
    'stiffness': 0.94,
    'mass': 4.0,
    'damping': 7.9
}

# Correct object placements and colors based on the images
sim.create_object(
    object_id=1,
    object_name='bottle',
    object_location=('row_1', 'column_1'),
    object_color='yellow',
    object_physics=physical_parameters_for_bottle,
)
sim.create_object(
    object_id=2,
    object_name='bottle',
    object_location=('row_1', 'column_2'),
    object_color='cyan',
    object_physics=physical_parameters_for_bottle,
)
sim.create_object(
    object_id=3,
    object_name='bottle',
    object_location=('row_1', 'column_3'),
    object_color='orange',
    object_physics=physical_parameters_for_bottle,
)
sim.create_object(
    object_id=4,
    object_name='martini_glass',
    object_location=('row_2', 'column_1'),
    object_color='brown',
    object_physics=physical_parameters_for_martini_glass,
)
sim.create_object(
    object_id=5,
    object_name='bottle',
    object_location=('row_2', 'column_2'),
    object_color='blue',
    object_physics=physical_parameters_for_bottle,
)
sim.create_object(
    object_id=6,
    object_name='wine_glass',
    object_location=('row_2', 'column_3'),
    object_color='pink',
    object_physics=physical_parameters_for_wine_glass,
)  # Corrected to wine glass
sim.create_object(
    object_id=7,
    object_name='martini_glass',
    object_location=('row_3', 'column_1'),
    object_color='purple',
    object_physics=physical_parameters_for_martini_glass,
)  # Corrected color position
sim.create_object(
    object_id=8,
    object_name='wine_glass',
    object_location=('row_3', 'column_3'),
    object_color='green',
    object_physics=physical_parameters_for_wine_glass,
)

sim.create_scene()
sim_out=sim.run_simulation()
del sim
```

Changes made:
1. For `object_id=6`, the type was changed to `wine_glass` instead of 
`martini_glass`.
2. The color `purple` was moved to position `(row_3, column_1)` for the 
`martini_glass` as per the task image.

Ensure the scene setup provided above now accurately reflects the task images. 
With these corrections, the positions and object types should match the top-down 
and side view images respectively.

========

To refine the Python code for generating the scene to exactly match 
the given images ('task_image_top_view_1.png' and 'task_image_side_view_2.png'), 
we need to correct the object colors and their placements on the grid. Below 
is the mistake analysis from GPT Attempt #0 and the necessary corrections:

1. **Object at `(row_1, column_2)`**: The given image shows this is yellow, but 
the existing code has it as cyan.
2. **Object at `(row_1, column_3)`**: The given image shows this is cyan, but 
the existing code has it as orange.
3. **Object at `(row_3, column_1)`**: The given image shows this is orange, but 
the existing code has it as purple.

We can use this information to adjust the code appropriately:

1. Place the cyan bottle at `(row_1, column_3)`.
2. Place the orange bottle at `(row_3, column_1)`.
3. Place the purple martini glass at `(row_2, column_2)`.

Let's write the corrected code accordingly:

```python
sim = SIMULATOR_MODEL()
sim.create_pusher('3.0 3.0 0.05')

physical_parameters_for_object_id_tray = { 
    'sliding-friction': 0.1,
    'armature': 0.1,
    'stiffness': 0.0,
    'mass': 0.5,
    'damping': 20
}
sim.create_tray(object_physics = physical_parameters_for_object_id_tray)

# Physical parameters for bottle
physical_parameters_for_bottle = { 
    'sliding-friction': 0.21,
    'armature': 0.41,
    'stiffness': 0.55,
    'mass': 20.0,
    'damping': 7.0
}

# Physical parameters for martini glass
physical_parameters_for_martini_glass = { 
    'sliding-friction': 0.52,
    'armature': 0.63,
    'stiffness': 1.22,
    'mass': 10.0,
    'damping': 8.0
}

# Physical parameters for wine glass
physical_parameters_for_wine_glass = { 
    'sliding-friction': 0.72,
    'armature': 1.02,
    'stiffness': 0.94,
    'mass': 4.0,
    'damping': 7.9
}

# Correct object placements and colors based on the images
sim.create_object(
    object_id=1,
    object_name='bottle',
    object_location=('row_1', 'column_1'),
    object_color='yellow',
    object_physics=physical_parameters_for_bottle,
)
sim.create_object(
    object_id=2,
    object_name='bottle',
    object_location=('row_1', 'column_3'),
    object_color='cyan',  # corrected color
    object_physics=physical_parameters_for_bottle,
)
sim.create_object(
    object_id=3,
    object_name='bottle',
    object_location=('row_3', 'column_1'),
    object_color='orange',  # corrected color
    object_physics=physical_parameters_for_bottle,
)
sim.create_object(
    object_id=4,
    object_name='martini_glass',
    object_location=('row_2', 'column_1'),
    object_color='brown',
    object_physics=physical_parameters_for_martini_glass,
)
sim.create_object(
    object_id=5,
    object_name='bottle',
    object_location=('row_2', 'column_2'),
    object_color='blue',
    object_physics=physical_parameters_for_bottle,
)
sim.create_object(
    object_id=6,
    object_name='wine_glass',
    object_location=('row_2', 'column_3'),
    object_color='pink',
    object_physics=physical_parameters_for_wine_glass,
)
sim.create_object(
    object_id=7,
    object_name='martini_glass',
    object_location=('row_1', 'column_2'),
    object_color='purple',  # corrected color
    object_physics=physical_parameters_for_martini_glass,
)
sim.create_object(
    object_id=8,
    object_name='wine_glass',
    object_location=('row_3', 'column_3'),
    object_color='green',
    object_physics=physical_parameters_for_wine_glass,
)

sim.create_scene()
sim_out=sim.run_simulation()
del sim
```

This code should now correctly reproduce the task images:

- **Object Placements**:
  - `(row_1, column_1)` -> Yellow bottle
  - `(row_1, column_2)` -> Purple martini glass
  - `(row_1, column_3)` -> Cyan bottle
  - `(row_2, column_1)` -> Brown martini glass
  - `(row_2, column_2)` -> Blue bottle
  - `(row_2, column_3)` -> Pink wine glass
  - `(row_3, column_1)` -> Orange bottle
  - `(row_3, column_3)` -> Green wine glass

With these corrected attributes and proper placements, running the simulation 
should yield the desired images.

========

Thank you for providing the various details and previous attempts. Let's analyze 
the array of object placements and the corresponding colors more carefully, 
given both views and the physical properties.

The given task_images (task_image_top_view_1.png and task_image_side_view_2.png) 
indicate a specific arrangement of colored objects. Let's identify and correct 
the colors and positions of the objects in the generated scene to match the task:

1. Yellow object: A bottle, located in row 1, column 1.
2. Cyan object: A bottle, located in row 1, column 3.
3. Orange object: A bottle, located in row 1, column 2.
4. Purple object: A martini glass, located in row 3, column 2.
5. Blue object: A bottle, located in row 2, column 2.
6. Pink object: A wine glass, located in row 3, column 1.
7. Green object: A wine glass, located in row 3, column 3.
8. Brown object: A martini glass, located in row 2, column 1.

Using the corrected layout information, here is the refined Python code to reproduce 
the task images correctly:

```python
sim = SIMULATOR_MODEL()
sim.create_pusher('3.0 3.0 0.05')

# Physical parameters for the tray
physical_parameters_for_object_id_tray = { 
    'sliding-friction': 0.1,
    'armature': 0.1,
    'stiffness': 0.0,
    'mass': 0.5,
    'damping': 20
}
sim.create_tray(object_physics = physical_parameters_for_object_id_tray)

# Physical parameters for bottle
physical_parameters_for_bottle = { 
    'sliding-friction': 0.21,
    'armature': 0.41,
    'stiffness': 0.55,
    'mass': 20.0,
    'damping': 7.0
}

# Physical parameters for martini glass
physical_parameters_for_martini_glass = { 
    'sliding-friction': 0.52,
    'armature': 0.63,
    'stiffness': 1.22,
    'mass': 10.0,
    'damping': 8.0
}

# Physical parameters for wine glass
physical_parameters_for_wine_glass = { 
    'sliding-friction': 0.72,
    'armature': 1.02,
    'stiffness': 0.94,
    'mass': 4.0,
    'damping': 7.9
}

# Correct object placements and colors based on the task images
sim.create_object(
    object_id=1,
    object_name='bottle',
    object_location=('row_1', 'column_1'),
    object_color='yellow',
    object_physics=physical_parameters_for_bottle,
)
sim.create_object(
    object_id=2,
    object_name='bottle',
    object_location=('row_1', 'column_3'),
    object_color='cyan',
    object_physics=physical_parameters_for_bottle,
)
sim.create_object(
    object_id=3,
    object_name='bottle',
    object_location=('row_1', 'column_2'),
    object_color='orange',
    object_physics=physical_parameters_for_bottle,
)
sim.create_object(
    object_id=4,
    object_name='martini_glass',
    object_location=('row_2', 'column_1'),
    object_color='brown',
    object_physics=physical_parameters_for_martini_glass,
)
sim.create_object(
    object_id=5,
    object_name='bottle',
    object_location=('row_2', 'column_2'),
    object_color='blue',
    object_physics=physical_parameters_for_bottle,
)
sim.create_object(
    object_id=6,
    object_name='wine_glass',
    object_location=('row_3', 'column_1'),
    object_color='pink',
    object_physics=physical_parameters_for_wine_glass,
)
sim.create_object(
    object_id=7,
    object_name='wine_glass',
    object_location=('row_3', 'column_3'),
    object_color='green',
    object_physics=physical_parameters_for_wine_glass,
)
sim.create_object(
    object_id=8,
    object_name='martini_glass',
    object_location=('row_3', 'column_2'),
    object_color='purple',
    object_physics=physical_parameters_for_martini_glass,
)

sim.create_scene()
sim_out=sim.run_simulation()
del sim
```

This code corrects the placement and color assignments for the given objects, 
ensuring they match what is observed in the task image. Run this code in your 
simulator to verify the outputs.
\end{lstlisting}